\documentclass{article} 
\usepackage{iclr2022_conference,times}

\usepackage{amsmath,amsfonts,bm}

\def\eqref#1{equation~\ref{#1}}

\def\1{\bm{1}}

\DeclareMathAlphabet{\mathsfit}{\encodingdefault}{\sfdefault}{m}{sl}
\SetMathAlphabet{\mathsfit}{bold}{\encodingdefault}{\sfdefault}{bx}{n}

\newcommand{\E}{\mathbb{E}}

\newcommand{\KL}{D_{\mathrm{KL}}}

\DeclareMathOperator*{\argmax}{arg\,max}
\DeclareMathOperator*{\argmin}{arg\,min}

\usepackage{hyperref}
\usepackage{url}

\usepackage{amsmath}
\usepackage{amsfonts}
\usepackage{amssymb}
\usepackage{subfig}

\usepackage{graphicx}		
\usepackage{wrapfig}		
\usepackage{algorithm}		
\usepackage{algorithmic}	
\usepackage{color}			
\usepackage{multicol}		
\usepackage{multirow} 		
\usepackage{makecell}
\usepackage{colortbl}
\usepackage{bbding}

\usepackage[figuresright]{rotating}

\usepackage{adjustbox}
\usepackage{booktabs}

\usepackage{tikz}
\usetikzlibrary{positioning, arrows.meta}

\title{DARA: Dynamics-Aware Reward Augmentation in Offline Reinforcement Learning}

\author{
	Jinxin Liu$^{123}$\thanks{Equal contribution.} \ \  \ \ 
	Hongyin Zhang$^{1*}$ \ \  \ 
	Donglin Wang$^{13}$\thanks{Corresponding author.} \\
	$^1$ Westlake University. \ \ $^2$ Zhejiang University. \\
	$^3$ Institute of Advanced Technology, Westlake Institute for Advanced Study. \\
	\{liujinxin, zhanghongyin, wangdonglin\}@westlake.edu.cn
}

\newcommand{\lbr}{\left [}
\newcommand{\rbr}{\right ]}
\newcommand{\lpa}{\left (}
\newcommand{\rpa}{\right )}

\newcommand{\bs}{\mathbf{s}}
\newcommand{\ba}{\mathbf{a}}
\newcommand{\B}{\mathcal{B}}
\newcommand{\G}{\mathcal{G}}
\newcommand{\D}{\mathcal{D}}

\newcommand{\DS}{{\textcolor{blue}{\mathcal{D'}}}}
\newcommand{\DT}{{\textcolor{red}{\mathcal{D}}}}
\newcommand{\TS}{{\textcolor{blue}{T'}}}
\newcommand{\TT}{{\textcolor{red}{T}}}
\newcommand{\MS}{{\textcolor{blue}{M'}}}
\newcommand{\MT}{{\textcolor{red}{M}}}
\newcommand{\pibs}{{{\textcolor{blue}{\pi_{b'}}}}}
\newcommand{\pibt}{{{\textcolor{red}{\pi_b}}}}
\newcommand{\HTS}{{\textcolor{blue}{\hat{T}'}}}
\newcommand{\HTT}{{\textcolor{red}{\hat{T}}}}
\newcommand{\HMS}{{\textcolor{blue}{\hat{M}'}}}
\newcommand{\HMT}{{\textcolor{red}{\hat{M}}}}

\newcommand{\SSS}{{\textcolor{blue}{S_{\textcolor{black}{\pi}}'}}}
\newcommand{\SST}{{\textcolor{red}{S}_\pi}}
\newcommand{\DR}{{\textcolor{blue}{\Delta r}}}

\newcommand{\phit}{{\textcolor{red}{\phi}}}
\newcommand{\phis}{{\textcolor{blue}{\phi'}}}
\newcommand{\thetat}{{\textcolor{red}{\theta}}}
\newcommand{\thetas}{{\textcolor{blue}{\theta'}}}

\newcommand{\TV}{D_\text{TV}}

\newcommand{\eg}{\emph{e.g.},\ }
\newcommand{\ie}{\emph{i.e.},\ }

\newtheorem{definition}{Definition}
\newtheorem{assumption}{Assumption}
\newtheorem{theorem}{Theorem}
\newtheorem{lemma}[theorem]{Lemma}

\definecolor{magiccolor}{RGB}{205, 232, 248}%

\usepackage{xcolor}
\usepackage{soul}

\iclrfinalcopy 
\begin{document}

\maketitle

\begin{abstract}
Offline reinforcement learning algorithms promise to be applicable in settings where a fixed dataset is available and no new experience can be acquired. 
However, such formulation is inevitably offline-data-hungry and, in practice, collecting a large offline dataset for one specific task over one specific environment is also costly and laborious. 
In this paper, we thus 1) formulate the offline dynamics adaptation by using (source) offline data collected from another dynamics to relax the requirement for the extensive (target) offline data, 2) characterize the dynamics shift problem in which prior offline methods do not scale well, and 3) derive a simple dynamics-aware reward augmentation (DARA) framework from both model-free and model-based offline settings. Specifically, DARA emphasizes learning from those source transition pairs that are adaptive for the target environment and mitigates the offline dynamics shift by characterizing state-action-next-state pairs instead of the typical state-action distribution sketched by prior offline RL methods. 
The experimental evaluation demonstrates that DARA, by augmenting rewards in the source offline dataset, can acquire an adaptive policy for the target environment and yet significantly reduce the requirement of target~offline~data. With only modest amounts of target offline data, our performance consistently outperforms the prior offline RL methods in both simulated and real-world tasks. 
\end{abstract}

\section{Introduction}
\begin{wrapfigure}{r}{0.305\textwidth}
	\vspace{-15pt}
	\begin{center}
		\includegraphics[scale=0.378]{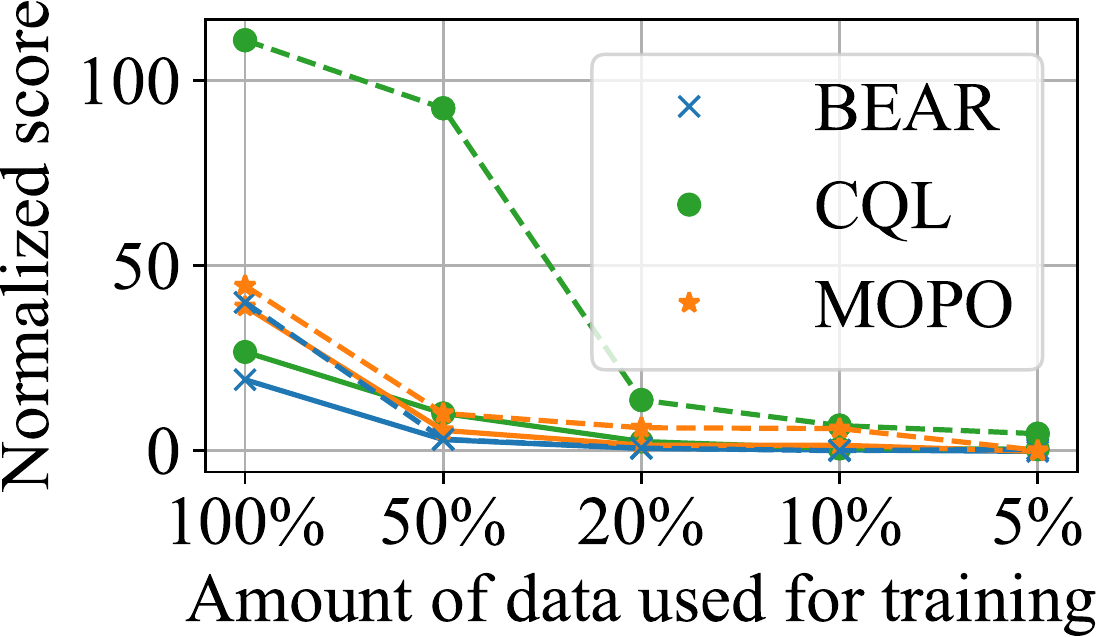}
	\end{center}
	\vspace{-11pt}
	\caption{Solid and dashed lines denote offline Medium-Replay and Medium-Expert data in D4RL (Walker2d){\tiny~}\emph{resp}.
	}
	\vspace{-15pt}
	\label{fig-introduction-data-reduce}
\end{wrapfigure}
Offline reinforcement learning (RL)~\citep{offline2020Levine, batch2012lange}, the task of learning from the previously collected dataset, holds the promise of acquiring policies without any costly active interaction required in the standard online RL paradigm. 
However, we note that although the active trail-and-error (online exploration) is eliminated, the performance of offline RL method heavily relies on the amount of offline data that is used for training. 
As shown in Figure~\ref{fig-introduction-data-reduce}, 
the performance deteriorates dramatically as the amount of offline data decreases. 
A natural question therefore arises: can we reduce the amount of the (target) offline data without significantly affecting the final performance for the target task?  

Bringing the idea from the transfer learning~\citep{transferlearning2020}, we assume that we have access to another (source) offline dataset, hoping that we can leverage this dataset to compensate for the performance degradation caused by the reduced (target) offline dataset. In the offline setting, previous work~\citep{keep2020siegel, actionable2021Chebotar} has characterized the reward (goal) difference between the source and target, relying on the "conflicting" or multi-goal offline dataset~\citep{d4rl2020Fu}, while we focus on the relatively unexplored transition dynamics difference between the source dataset and the target environment. 
Meanwhile, we believe that this dynamics shift is not arbitrary in reality: in healthcare treatment, offline data for a particular patient is often limited, whereas we can obtain diagnostic data from other patients with the same case (same reward/goal) and there often exist individual differences between patients (source dataset with different transition dynamics). Careful treatment with respect to the individual differences is thus a crucial requirement.  

Given source offline data, the main challenge is to cope with the transition dynamics difference, \ie strictly tracking the state-action supported by the source offline data can not guarantee that the same transition (state-action-next-state) can be achieved in the target environment. 
However, in the offline setting, such dynamics shift is not explicitly characterized by the previous offline RL methods, where they typically attribute the difficulty of learning from offline data to the state-action distribution shift~\citep{info2019Chen, breaking2018liu}. The corresponding algorithms~\citep{off2019Fujimoto, mapo2018Abdolmaleki, mopo2020Yu} that model the support of state-action distribution induced by the learned policy, will inevitably suffer from the transfer problem where dynamics shift happens. 

 


Our approach is motivated by the well established connection between reward modification and dynamics adaptation~\citep{one2020kumar, ifmaxent2019eysenbach, offd2018Eysenbach}, which indicates that, by modifying rewards, one can train a policy in one environment and make the learned policy to be suitable for another environment (with different dynamics). 
Thus, we propose to exploit the \emph{joint distribution of state-action-next-state}:  besides characterizing the state-action distribution shift as in prior offline RL algorithms, we additionally identify the dynamics (\emph{i.e.}, the conditional distribution of next-state given current state-action pair) shift and penalize the agent with a dynamics-aware reward modification. Intuitively, this reward modification aims to discourage the learning from these offline transitions that are likely in source but are unlikely in the target environment. Unlike the concurrent work~\citep{augumented2021Ball, offlinemeta2021Mitchell} paying attention to the offline domain generalization, we explicitly focus on the offline domain (dynamics) adaptation. 

Our principal contribution in this work is the characterization of the dynamics shift in offline RL and the derivation of dynamics-aware reward augmentation (DARA) framework built on prior model-free and model-based formulations. DARA is simple and general, can accommodate various offline RL methods, and can be implemented in just a few lines of code
on top of dataloader at training. 
In our offline dynamics adaptation setting, we also release a dataset, including the Gym-MuJoCo tasks (Walker2d, Hopper and HalfCheetah), with dynamics (mass, joint) shift compared to D4RL, and a 12-DoF quadruped robot in both simulator and real-world. 
With only modest amounts of target offline data, we show that DARA-based offline methods can acquire an adaptive policy for the target tasks and achieve better performance compared to baselines in both simulated and real-world~tasks.

\vspace{-2pt}
\section{Related Work}
\vspace{-2pt}

Offline RL describes the setting in which a learner has access to
only a fixed dataset of experience, while no interactive data collection is allowed during policy learning~\citep{offline2020Levine}. Prior work commonly \emph{assumes} that the offline experience is collected by some behavior policies on the same environment that the learned policy be deployed on. 
Thus, the main difficulty of such offline setting is the state-action distribution shift~\citep{off2019Fujimoto, breaking2018liu}. Algorithms address this issue by following the two main directions: the \emph{model-free} and \emph{model-based} offline RL.

\emph{Model-free} methods for such setting typically fall under three categories: 
1) Typical methods mitigate this problem by explicitly~\citep{off2019Fujimoto,stabilizing2019Kumar, behavior2019Wu} or implicitly~\citep{keep2020siegel,awr2019Peng,mapo2018Abdolmaleki} constraining the learned policy away from OOD state-action pairs. 
2) Conservative estimation based methods learn pessimistic value functions to prevent the overestimation~\citep{cql2020Kumar,cpq2021Xu}. 
3) Importance sampling based methods directly estimate the state-marginal importance ratio and obtain  an unbiased value estimation~\citep{gnedice2020zhang, reinforcement2020nachum, algaedice2019nachum}.

\emph{Model-based} methods typically eliminate the state-action distribution shift by incorporating a reward penalty, which relies on the uncertainty quantification of the learned dynamics~\citep{morel2020Kidambi, mopo2020Yu}. To remove this uncertainty estimation, \citet{combo2021Yu} learns conservative critic function by penalizing the values of the generated state-action pairs that are not in the offline{\tiny~}dataset.

These methods, however, define their objective based on the state-action distribution  shift, and ignore the potential dynamics shift between the fixed offline data and the target MDP. In contrast, we account for dynamics (state-action-next-state) shift and explicitly propose the dynamics aware~reward augmentation. 
A counterpart, close to our work, is off-dynamics RL~\citep{offd2018Eysenbach}, where they set up dynamics shift in the interactive environment while we focus on the offline setting.

\vspace{-2pt}
\section{Preliminaries}
\label{section-mdp-and-offlinerl}
\vspace{-2pt}

We study RL in the framework of Markov decision processes (MDPs) specified by the tuple ${M} := (\mathcal{S}, \mathcal{A}, r, T, \rho_0, \gamma)$, where $\mathcal{S}$ and $\mathcal{A}$ denote the state and action spaces, $r(\bs,\ba) \in [-R_{max}, R_{max}]$ is the reward function, $T(\bs'|\bs,\ba)$ is the transition dynamics, $\rho_0(\bs)$ is the initial state distribution, and $\gamma$ is the discount factor. The goal in RL is to optimize a policy $\pi(\ba|\bs)$ that maximizes the expected discounted return $\eta_M(\pi) :=\mathbb{E}_{\tau\sim p^\pi_M(\tau)} \left[ \sum_{t=0}^{\infty} \gamma^t r(\bs_t, \ba_t) \right]$, where  $\tau:=(\bs_0,\ba_0,\bs_1,\ba_1, ...)$. 
We also define Q-values $Q(\bs, \ba) :=  \E_{\tau\sim p^\pi_M(\tau)}\lbr \sum_{t=0}^{\infty}\gamma^t r(\bs_t, \ba_t) \vert \bs_0=\bs, \ba_0=\ba \rbr$,  V-values $V(\bs) := \E_{\ba\sim\pi(\ba|\bs)}\lbr Q(\bs, \ba) \rbr$, and the (unnormalized) state visitation distribution $d_M^\pi(\bs):=\sum_{t=0}^{\infty} \gamma^t P(\bs|\pi,$ $M,t)$, where  $P(\bs|\pi,M,t)$ denotes the probability of reaching state $\bs$ at time $t$ by running $\pi$ in $M$. 

In the \emph{offline RL} problem, we are provided with a static dataset $\D:=\{(\bs, \ba, r, \bs')\}$, which consists of transition tuples from trajectories collected by running one or more behavioral policies, denoted by $\pi_b$, on MDP ${M}$. 
With a slight abuse of notation, we write $\D=\{(\bs, \ba, r, \bs') \sim d_\D(\bs)\pi_b(\ba|\bs)r(\bs,\ba)T(\bs'|\bs,\ba)\}$, where the $d_\D(\bs)$ denotes state-marginal distribution in $\D$. 
In the offline setting, the goal is typically to learn the best possible policy using the fixed offline dataset.

\textbf{Model-free RL} algorithms based on dynamic programming typically perform policy iteration to find the optimal policy. 
Such methods iteratively conduct 
1) policy~improvement with $\G_M Q := \argmax_\pi \E_{\bs\sim d^{\pi}_M(\bs),\ba\sim\pi(\ba|\bs)}\lbr Q(\bs,\ba) \rbr$ and 
2) policy evaluation
by iterating the Bellman equation $Q(\bs,\ba) = \B_M^\pi Q(\bs,\ba) := r(\bs,\ba) + \gamma \E_{\bs'\sim T(\bs'|\bs,\ba), \ba'\sim \pi(\ba'|\bs')}\lbr Q(\bs',\ba') \rbr $~over~$d^{\pi}_M(\bs)\pi(\ba|\bs)$. 
Given off-policy $\D$, we resort to 1) improvement with $\G_\D Q := \argmax_\pi \E_{\bs\sim d_\D(\bs),\ba\sim\pi(\ba|\bs)}\lbr Q(\bs,\ba) \rbr$ and 2) evaluation by iterating~$Q(\bs,\ba) = {\B}_\mathcal{D}^\pi Q(\bs,\ba) := r(\bs,\ba) + \gamma \E_{\bs'\sim T_\D(\bs'|\bs,\ba),\ba'\sim\pi(\ba'|\bs')}\lbr  Q(\bs',\ba') \rbr$ over all $(\bs,\ba)$ in $\D$. Specifically, given any initial $Q^0$, it iterates\footnote{For parametric Q-function, we often perform $Q^{k+1} \leftarrow \argmin_Q \E_{(\bs,\ba)\sim\D}[ ( {\B}_\D^{\pi^{k+1}} Q^k(\bs,\ba) - Q(\bs,\ba) )^2 ]$.} 
\begin{equation}\label{eq-pi-pe}
\text{Policy improvement: \ } \pi^{k+1} = \G_D Q^k, \quad \quad  
\text{Policy evaluation: \ } Q^{k+1} = {\B}_\D^{\pi^{k+1}} Q^k. 
\end{equation}
\textbf{Model-free offline RL} based on the above iteration suffers from the  state-action distribution shift, \ie policy evaluation ${\B}_\D^{\pi^k} Q^{k-1}$ may encounter unfamiliar state action regime that is not covered by the fixed offline dataset $\D$, causing erroneous estimation of $Q^k$. Policy improvement $\G_\D Q^k $ further exaggerates such error, biasing policy $\pi^{k+1}$ towards out-of-distribution (OOD) actions with erroneously high Q-values. To address this distribution shift, prior works 1) explicitly constrain policy to be close to the behavior policy~\citep{off2019Fujimoto, stabilizing2019Kumar, behavior2019Wu, emaq2021Ghasemipour}, introducing penalty ${\alpha D(\pi(\ba|\bs), \pi_b(\ba|\bs))}$ into $\G_\D$ or ${\B}_\D^\pi $ in Equation~\ref{eq-pi-pe}: 
\begin{equation}
\label{eq-model-free-D}
\begin{split}
\G_\D Q &= \argmax_\pi \E_{\bs\sim d_\D(\bs), \ba\sim \pi(\ba|\bs)} \lbr Q(\bs,\ba) - {\alpha D(\pi(\ba|\bs), \pi_b(\ba|\bs))} \rbr, \\
{\B}_\D^\pi Q(\bs,\ba) &= r(\bs, \ba) + \gamma \E_{\bs'\sim T_\D(\bs'|\bs,\ba), \ba'\sim\pi(\ba'|\bs')}\lbr Q(\bs',\ba') - {\alpha D(\pi(\ba'|\bs'), \pi_b(\ba'|\bs'))} \rbr,
\end{split}
\end{equation}
where $D$ is a divergence function between distributions over actions (\eg MMD or KL divergence), 
or 2) train pessimistic value functions~\citep{cql2020Kumar, combo2021Yu, cpq2021Xu}, penalizing Q-values at states in the offline dataset $\D$ for actions generated by the current policy $\pi$:
\begin{equation}
\label{eq-cql}
Q = \argmin_Q \  \E_{\bs\sim d_\D(\bs), \ba\sim\pi(\ba|\bs)}\lbr Q(\bs,\ba) \rbr, \quad \text{s.t.} \ \  Q = {\B}_\D^{\pi} Q.
\end{equation} 
\textbf{Model-based{\tiny~}RL} algorithms iteratively 1) model the transition dynamics $T(\bs'|\bs,\ba)$, using the data collected in  $M$: $\max_{\hat{T}} \E_{\bs,\ba,\bs' \sim d_M^\pi(\bs)\pi(\ba|\bs)T(\bs'|\bs,\ba)}[ \log \hat{T}(\bs'|\bs,\ba) ]$, and 2) infer a policy $\pi$ from the modeled $\hat{M} = (\mathcal{S}, \mathcal{A}, r, \hat{T}, \rho_0, \gamma)$, where we assume that $r$ and $\rho_0$ are known,  maximizing $\eta_{\hat{M}}(\pi)$ with a planner or the Dyna-style algorithms~\citep{mdp1990Sutton}. In this paper, we focus on the latter. 


\textbf{Model-based offline RL} algorithms similarly suffer from OOD state-action~\citep{morel2020Kidambi, behavioral2021Cang} if we directly apply policy iteration over $\hat{T}:=\max_{\hat{T}} \E_{\bs,\ba,\bs' \sim \D}[ \log \hat{T}(\bs'|\bs,\ba) ]$. 
Like the conservative estimation approach described in Equation~\ref{eq-cql}, recent conservative model-based offline RL methods provide the policy with a penalty for visiting states under the estimated $\hat{T}$ where $\hat{T}$ is likely to be incorrect. Taking $u(\bs,\ba)$ as the oracle uncertainty~\citep{mopo2020Yu} that provides a consistent estimate of the accuracy of model $\hat{T}$ at $(\bs, \ba)$, we can modify the reward function to obtain a conservative MDP: $\hat{M}_c = (\mathcal{S}, \mathcal{A}, r-\alpha u, \hat{T}, \rho_0, \gamma)$, then learn a policy $\pi$ by maximizing~$\eta_{\hat{M}_c}(\pi)$. 

\vspace{-2pt}
\section{Problem Formulation}
\vspace{-2pt}

In standard offline RL problem, the static offline dataset $\DT$ consists of samples~$\{(\bs,\ba,r,\bs')\sim d_\DT (\bs)\pibt(\ba|\bs)r(\bs,\ba)\TT(\bs'|\bs,\ba)\}$. 
Although offline RL methods learn policy for the target MDP $\MT := (\mathcal{S}, \mathcal{A}, r, \TT, \rho_0, \gamma)$ without (costly) online data, as we shown in Figure~\ref{fig-introduction-data-reduce}, it requires a fair amount of (target) offline data $\DT$ collected on $\MT$. 
Suppose we have another (source) offline dataset $\DS$, consisting of samples $\{(\bs,\ba,r,\bs')\sim d_\DS (\bs)\pibs(\ba|\bs)r(\bs,\ba)\TS(\bs'|\bs,\ba)\}$ collected by the behavior policy $\pibs$ on MDP $\MS := (\mathcal{S}, \mathcal{A}, r, \TS, \rho_0, \gamma)$, then we hope the transfer of knowledge between offline dataset $\{\DS \cup \DT\}$ can reduce the data requirements on $\DT$ for learning policy for the target~$\MT$. 



\vspace{-2.5pt}
\subsection{Dynamics Shift in Offline RL}
\vspace{-2.5pt}

Although offline RL methods in Section~\ref{section-mdp-and-offlinerl} have incorporated the state-action distribution constrained backups (policy constraints or conservative estimation), they also fail to learn an adaptive policy for the target MDP $\MT$ with the mixed datasets $\{\DS \cup \DT\}$, as we show in Figure~\ref{appendix:figure-introduction-plus10s} (Appendix). We attribute this failure to the dynamics shift  (Definition~\ref{def-dynamics-shifts}) between $\DS$ and $\MT$ in this adaptation setting.

\vspace{-5pt}
\begin{definition} 
	\label{def-emperical-mdp}
	(Empirical MDP) An empirical MDP estimated from $\D$ is $\hat{M} := (\mathcal{S}, \mathcal{A}, r, \hat{T}, \rho_0, \gamma)$ where $\hat{T} = \max_{\hat{T}} \E_{\bs,\ba,\bs' \sim \D}[ \log \hat{T}(\bs'|\bs,\ba) ]$ and $\hat{T}(\bs'|\bs,\ba) = 0$ for all $(\bs,\ba,\bs')$ not in dataset $\D$. 
\end{definition}
\vspace{-5pt}

\begin{definition} 
	\label{def-dynamics-shifts}
	(Dynamics shift)
	Let $\hat{M} := (\mathcal{S}, \mathcal{A}, r, \hat{T}, \rho_0, \gamma)$ be the empirical MDP estimated from $\D$. To evaluate a policy $\pi$ for $M := (\mathcal{S}, \mathcal{A}, r, T, \rho_0, \gamma)$ with offline dataset $\D$, 
	we say that the dynamics shift (between $\D$ and $M$) in offline RL happens if there exists at least one transition pair $(\bs,\ba,\bs') \in \{ (\bs,\ba,\bs'):d_{\hat{M}}^\pi(\bs)\pi(\ba|\bs)\hat{T}(\bs'|\bs,\ba)>0 \}  $ such that $\hat{T}(\bs'|\bs,\ba) \neq T(\bs'|\bs,\ba)$. 
\end{definition} 

In practice, for a stochastic $M$ and any finite offline data $\D$ collected in $M$, there always exists the dynamics shift. 
The main concern is that finite samples are always not sufficient to exactly model stochastic dynamics. Following~\citet{off2019Fujimoto}, we thus assume both MDPs $\MT$ and $\MS$ are deterministic, which means the empirical $\HMT$ and $\HMS$ are both also deterministic. More importantly, such assumption enables us to explicitly characterize the dynamics shift under finite  offline samples.


\begin{lemma}\label{lemma-target-no-dynamics-shift}
Under deterministic transition dynamics, there is no dynamics shift between $\DT$ and $\MT$. 
\end{lemma}

For offline RL tasks, prior methods generally apply ${\B}_\D^\pi Q$ along with the state-action distribution correction (Equations~\ref{eq-model-free-D} and~\ref{eq-cql}), which overlooks the potential dynamics shift between the (source) offline dataset and the target MDP (\eg $\DS \to \MT$). As a result, these methods do not scale well to the setting in which dynamics shift happens, 
\eg learning an adaptive policy for $\MT$ with~(source)~$\DS$.


\vspace{-2pt}
\subsection{Dynamics Shift in Model-free and Model-based Offline Formulations}
\vspace{-2pt}

From the \textbf{model-free} (policy iteration) view, an exact policy evaluation on $\MT$ is characterized by iterating $Q(\bs,\ba) = \B_\MT^\pi Q(\bs,\ba)$ for all $(\bs,\ba)$ such that $d_\MT^\pi(\bs)\pi(\ba|\bs)>0$. Thus, to formalize the policy evaluation with offline $\DT$ or $\DS$ (for an adaptive $\pi$ on target $\MT$), 
we require that Bellman operator $\B_\DT^\pi Q(\bs,\ba)$ or $\B_\DS^\pi Q(\bs,\ba)$ approximates the oracle $\B_\MT^\pi Q(\bs,\ba)$ for all $(\bs,\ba)$ in $\SST$ or $\SSS$, where $\SST$ and $\SSS$ denote the sets $\{(\bs,\ba):d_\DT(\bs)\pi(\ba|\bs)>0\}$ and $\{(\bs,\ba) : d_\DS(\bs)\pi(\ba|\bs)>0\}$~respectively. 

1) To evaluate a policy $\pi$ for $\MT$ with $\DT$ (\ie calling the Bellman operator $\B_\DT^\pi$), notable model-free offline method BCQ~\citep{off2019Fujimoto} translates the requirement of $\B_\DT^\pi =\B_\MT^\pi$ into the requirement of $\HTT(\bs'|\bs,\ba) = \TT(\bs'|\bs,\ba)$. 
Note that under deterministic environments, we have the property that for all $(\bs,\ba,\bs')$ in offline data $\DT$, 
$\HTT(\bs'|\bs,\ba) = \TT(\bs'|\bs,\ba)$ (Lemma~\ref{lemma-target-no-dynamics-shift}). 
As a result, such property permits BCQ to evaluate a policy $\pi$ by calling $\B_\DT^\pi$, replacing the oracle $\B_\MT^\pi$, meanwhile constraining $\SST$ to be a subset of the support of $d_\DT(\bs)\pibt(\ba|\bs)$. This means a policy $\pi$ which only traverses transitions contained in (target) offline data $\DT$, can be evaluated on $\MT$ without error. 

2) To evaluate a policy $\pi$ for $\MT$ with $\DS$ (\ie calling the Bellman operator $\B_\DS^\pi$), we have lemma~\ref{theorem-bellman-shifts}:
\begin{lemma} 
	\label{theorem-bellman-shifts}
	Dynamics shift produces that $\B_\DS^\pi Q(\bs, \ba) \neq \B_\MT^\pi Q(\bs, \ba)$ for some $(\bs,\ba)$ in $\SSS$.
\end{lemma} 

With the offline data $\DS$, lemma~\ref{theorem-bellman-shifts} suggests that the above requirement $\B_\DS^\pi = \B_\MT^\pi$ becomes infeasible, which limits the practical applicability of prior offline RL methods under the dynamics shift. 
To be specific, 
characterizing an adaptive policy for target MDP $\MT$ with $\DS$ moves beyond the reach of the off-policy evaluation based on iterating $Q = \B_\DS^\pi Q$ (Equations~\ref{eq-model-free-D} and~\ref{eq-cql}). Such iteration may cause the evaluated $Q$ (or learned policy $\pi$) overfits to $\HTS$ and struggle to adapt to the target $\TT$. To overcome the dynamics shift, we would like to resort an additional compensation $\Delta_{\HTS,\TT}$ such that 
\begin{equation}\label{eq-motivation-model-free}
\B_\DS^\pi Q(\bs, \ba) + \Delta_{\HTS,\TT}(\bs,\ba) = \B_\MT^\pi Q(\bs, \ba)
\end{equation}
for all $(\bs,\ba)$ in $\SSS$. Thus, we can apply $\B_\DS^\pi Q+ \Delta_{\HTS,\TT}$ to act as a substitute for the oracle $\B_\MT^\pi Q$. 

From the \textbf{model-based} view, the oracle $\eta_{\MT}(\pi)$ (calling the Bellman operator $\B_\MT^\pi$ on the target $\MT$) and the viable $\eta_{\HMS}(\pi)$ (calling $\B_\HMS^\pi $ on the estimated $\HMS$ from source $\DS$) have the following{\tiny~}lemma. 
\begin{lemma}
	\label{lemma-model-based-error}
	Let $\B_M^\pi V(\bs) = \E_{\ba\sim\pi(\ba|\bs)}\lbr r(\bs,\ba) + \gamma \E_{\bs'\sim T(\bs'|\bs,\ba)}\lbr V(\bs') \rbr \rbr$. 
	For any $\pi$, we have:
	\begin{equation}
	\eta_{\HMS}(\pi) = \eta_{\MT}(\pi) +  \E_{\bs\sim d_\HMS^\pi(\bs)} \lbr \B_\HMS^\pi V_\MT(\bs) - \B_\MT^\pi V_\MT(\bs) \rbr. \nonumber 
	\end{equation}
\end{lemma}
Lemma~\ref{lemma-model-based-error} states that if we maximize $\eta_{\HMS}(\pi)$ subject to $\vert \E_{\bs\sim d_\HMS^\pi(\bs)} [ \B_\HMS^\pi V_\MT(\bs) - \B_\MT^\pi V_\MT(\bs) ] \vert \leq \epsilon$, $\eta_{\MT}(\pi)$ will be improved. 
If $\mathcal{F}$ is a set of functions ${f: \mathcal{S} \to \mathbb{R}}$ that contains $V_\MT$, then we have
\begin{equation}\label{eq-motivation-model-based-upper}
\begin{split}
\left\vert \E_{\bs\sim d_\HMS^\pi(\bs)} \lbr \B_\HMS^\pi V_\MT(\bs) - \B_\MT^\pi V_\MT(\bs) \rbr \right\vert \leq \gamma \E_{\bs,\ba\sim d_\HMS^\pi(\bs)\pi(\ba|\bs)} \lbr  d_{\mathcal{F}}(\HTS(\bs'|\bs,\ba), \TT(\bs'|\bs,\ba))  \rbr, 
\end{split}
\end{equation}
where $d_{\mathcal{F}}(\HTS(\bs'|\bs,\ba), \TT(\bs'|\bs,\ba)) = \sup_{f \in \mathcal{F}} \vert  \E_{\bs'\sim \HTS(\bs'|\bs,\ba)}\lbr f(\bs') \rbr - \E_{\bs'\sim \TT(\bs'|\bs,\ba)}\lbr f(\bs') \rbr \vert $, which is the integral probability metric (IPM). 
Note that if we directly follow the \emph{admissible error} assumption in MOPO~\citep{mopo2020Yu} \ie  assuming $d_{\mathcal{F}}(\HTS(\bs'|\bs,\ba), \TT(\bs'|\bs,\ba)) \leq u(\bs,\ba)$ for all $(\bs,\ba)$, this would be too restrictive: given that $\HTS$ is estimated from the source offline samples collected under $\TS$, not the target $\TT$, thus such error would not decrease as  the  source data increases. Further, we~find
\begin{equation}\label{eq-motivation-model-based}
d_{\mathcal{F}}(\HTS(\bs'|\bs,\ba), \TT(\bs'|\bs,\ba)) \leq d_{\mathcal{F}}(\HTS(\bs'|\bs,\ba), \HTT(\bs'|\bs,\ba)) + d_{\mathcal{F}}(\HTT(\bs'|\bs,\ba), \TT(\bs'|\bs,\ba)).
\end{equation}
Thus, we can bound the $d_{\mathcal{F}}(\HTS, \TT)$ term with the admissible error assumption over $d_{\mathcal{F}}(\HTT, \TT)$, as in MOPO, and the auxiliary constraints $d_{\mathcal{F}}(\HTS, \HTT)$. See next section for the detailed implementation. 

 
\textbf{In summary}, we show that both prior offline model-free and model-based formulations  suffer from the dynamics shift, which also suggests us to learn a  modification ($\Delta$ or $d_{\mathcal{F}}$) to eliminate this~shift.


\vspace{-1pt}
\section{Dynamics-Aware Reward Augmentation}
\vspace{-1pt}

In this section, we propose the dynamics-aware reward augmentation (DARA), a simple data augmentation procedure based on prior (model-free and model-based) offline RL methods. 
We first provide an overview of our offline reward augmentation motivated by the compensation $\Delta_{\HTS,\TT}$ in Equation~\ref{eq-motivation-model-free} and the auxiliary constraints $d_{\mathcal{F}}(\HTS, \HTT)$ in Equation~\ref{eq-motivation-model-based}, and then describe its theoretical derivation in both model-free and model-based formulations. 
With the (reduced) target offline data $\DT$ and the source offline data $\DS$, 
we summarize the overall DARA framework in~Algorithm~\ref{alg-dara}. 

\begin{algorithm}
	\small
	\caption{~Framework for Dynamics-Aware Reward Augmentation (DARA)} 
	\label{alg-dara}
	\hspace{1pt} \textbf{Require:} Target offline data $\DT$  (reduced) and source offline data~$\DS$ 
	\begin{algorithmic} [1]
		\STATE {Learn classifiers ($q_{\text{sas}}$ and $q_{\text{sa}}$) that distinguish source data $\DS$ from target data $\DT$.}  (See~Appendix~\ref{appendix:learning-classifiers}) 
		\STATE Set dynamics-aware $\DR(\bs_t, \ba_t, \bs_{t+1}) = \log \frac{q_{\text{sas}}(\text{\textcolor{blue}{source}}|\bs_t,\ba_t,\bs_{t+1})}{q_{\text{sas}}(\text{\textcolor{red}{target}}|\bs_t,\ba_t,\bs_{t+1})} - \log \frac{q_{\text{sa}}(\text{\textcolor{blue}{source}}|\bs_t,\ba_t)}{q_{\text{sa}}(\text{\textcolor{red}{target}}|\bs_t,\ba_t)}$.\\
		\STATE Modify rewards for all $(\bs_t, \ba_t, r_t, \bs_{t+1})$ in $\DS$: $r_t \leftarrow r_t - \eta \DR$.
		\STATE Learn policy with $\{\DT \cup \DS\}$ using prior model-free or model-based offline RL algorithms. 
	\end{algorithmic}
\end{algorithm}

\vspace{-4pt}
\subsection{Dynamics-Aware Reward Augmentation in Model-free Formulation}
\vspace{-2pt}

Motivated by the well established connection of RL and probabilistic inference~\citep{controlasinference2018Levine}, we first cast the model-free RL problem as that of inference in a particular probabilistic model. Specifically, we introduce the binary random variable $\mathcal{O}$ that denotes whether the trajectory $\tau:=(\bs_0,\ba_0,\bs_1,...)$ is optimal ($\mathcal{O}=1$) or not ($\mathcal{O}=0$). The likelihood of a trajectory can then be modeled as $p(\mathcal{O}=1|\tau)=\exp\lpa \sum_t r_t/\eta\rpa$, where $r_t := r(\bs_t,\ba_t)$ and $\eta>0$ is a temperature parameter. 

\textbf{(Reward Augmentation with Explicit Policy/Value Constraints)} 
We now introduce a variational distribution $p_\HMS^{\pi}(\tau) = p(\bs_0)\prod_{t=1}\HTS(\bs_{t+1}|\bs_t,\ba_t){\pi}(\ba_t|\bs_t)$ to approximate the posterior distribution $p_\MT^\pi(\tau|\mathcal{O}=1)$, which leads to the evidence lower bound of $\log p_\MT^\pi(\mathcal{O}=1) $: 
\begin{align}\label{eq-rewardrel-model-free}
\log p_\MT^\pi(\mathcal{O}=1) 
&=  \log \E_{\tau\sim p_\MT^\pi(\tau)} \lbr p(\mathcal{O}=1|\tau) \rbr 
\geq \E_{\tau\sim p_\HMS^{\pi}(\tau)}\lbr \log p(\mathcal{O}=1|\tau) + \log \frac{p_\MT^\pi(\tau)}{p_\HMS^\pi(\tau)} \rbr \nonumber \\
&= \E_{\tau\sim p_\HMS^{\pi}(\tau)}\lbr \sum_{t} \lpa r_t/\eta - \log \frac{\HTS(\bs_{t+1}|\bs_t,\ba_t)}{\TT(\bs_{t+1}|\bs_t,\ba_t)} \rpa \rbr.
\end{align}
Since we are interested in infinite horizon problems, we introduce the discount factor $\gamma$ and {take the limit of steps in each rollout,} \ie $H\to\infty$. Thus, the RL problem on the MDP $\MT$, cast as the inference problem $\argmax_{\pi} \log p_\MT^\pi(\mathcal{O}=1) $, can be stated as a maximum of the lower bound $\E_{\tau\sim p_\HMS^{\pi}(\tau)}\lbr \sum_{t=0}^{\infty} \gamma^t\lpa r_t - \eta \log \frac{\HTS(\bs_{t+1}|\bs_t,\ba_t)}{\TT(\bs_{t+1}|\bs_t,\ba_t)} \rpa \rbr$. This is equivalent to an RL problem on $\HMS$ with the augmented reward $r \leftarrow r(\bs,\ba) - \eta \log  \frac{\HTS(\bs'|\bs,\ba)}{\TT(\bs'|\bs,\ba)} $. 
Intuitively, the $-\eta\log \frac{\HTS(\bs'|\bs,\ba)}{\TT(\bs'|\bs,\ba)}$ term 
discourages transitions (state-action-next-state) in $\DS$ that have low transition probability in the target $\MT$. 
In the model-free offline setting, we can add the explicit policy or Q-value constraints (Equations-\ref{eq-model-free-D}~and~\ref{eq-cql}) to mitigate the OOD state-actions. 
Thus, such formulation allows the oracle $\B_\MT^\pi$ to be~re-expressed by $\B_\DS^\pi$ and the modification $\log \frac{\HTS}{\TT}$, which makes the motivation in Equation~\ref{eq-motivation-model-free}~practical.

\textbf{(Reward Augmentation with Implicit Policy Constraints)} 
If we introduce the variational distribution $p_\HMS^{\pi'}(\tau) := p(\bs_0)\prod_{t=1}\HTS(\bs_{t+1}|\bs_t,\ba_t){\pi'}(\ba_t|\bs_t)$, we can recover the weighted-regression-style~\citep{crr2020wang, awr2019Peng, mapo2018Abdolmaleki, reps2010peters} objective by maximizing 
$ \mathcal{J}(\pi',\pi):= \E_{\tau\sim p_\HMS^{\pi'}(\tau)}\lbr \sum_{t=0}^{\infty} \gamma^t\lpa r_t - \eta \log \frac{\HTS(\bs_{t+1}|\bs_t,\ba_t)}{\TT(\bs_{t+1}|\bs_t,\ba_t)}  - \eta \log \frac{\pi'(\ba_t|\bs_t)}{\pi(\ba_t|\bs_t)} \rpa \rbr$ (lower bound of $\log p_\MT^\pi(\mathcal{O}=1)$). 
Following the Expectation Maximization (EM) algorithm, we can maximize $\mathcal{J(\pi',\pi)}$ by iteratively (E-step) improving $\mathcal{J}(\pi', \cdot)$ w.r.t. $\pi'$ and (M-step) updating $\pi$ w.r.t.~$\pi'$.

\emph{(E-step)} We define $\tilde{Q}(\bs,\ba,\bs') = \E_{\tau\sim p_\HMS^{\pi'}(\tau)}\lbr \sum_{t} \gamma^t  \log \frac{\HTS(\bs'|\bs,\ba)}{\TT(\bs'|\bs,\ba)} \vert \bs_0=\bs,\ba_0=\ba,\bs_1=\bs' \rbr $. Then, given offline data $\DS$, we can rewrite $\mathcal{J}(\pi', \cdot)$ as a constrained objective~\citep{mapo2018Abdolmaleki}: 
\begin{equation}
\max_{\pi'} \E_{d_\DS(\bs)\pi'(\ba|\bs)\HTS(\bs'|\bs,\ba) } \lbr Q(\bs,\ba)-\eta\tilde{Q}(\bs,\ba,\bs')\rbr, \text{ \ \ s.t. } \E_{\bs\sim d_\DS(\bs)} \lbr \KL \lpa \pi'(\ba|\bs) \Vert \pi(\ba|\bs) \rpa \rbr\leq \epsilon. \nonumber 
\end{equation}


When considering a fixed $\pi$, the above optimization over $\pi'$ can be solved analytically~\citep{kl2020Vieillard, kl2019Geist, awr2019Peng}. The optimal $\pi'_{*}$ is then given by $\pi'_{*}(\ba|\bs) \varpropto \pi(\ba|\bs)\exp\lpa Q(\bs,\ba) \rpa \exp ( -\eta\tilde{Q}(\bs,\ba,\HTS(\bs'|\bs,\ba)) )$. As the policy evaluation in Equation~\ref{eq-pi-pe} (Footnote-2), we  estimate $Q(\bs,\ba)$  and $\tilde{Q}(\bs,\ba,\bs')$ by minimizing the Bellman error with offline samples in~$\DS$. 

\emph{(M-step)} Then, we can project $\pi'_{*}$ onto the manifold of the parameterized $\pi$:
\begin{align}\label{eq-awr-qsas}
&\ \quad \argmin_\pi \quad \E_{\bs\sim d_\DS(\bs)} \lbr \KL \lpa \pi'_{*}(\ba|\bs) \Vert \pi(\ba|\bs) 
\rpa \rbr \nonumber \\
&= \argmax_\pi \quad \E_{\bs,\ba,\bs'\sim\DS} \lbr \log \pi(\ba|\bs) \exp\lpa Q(\bs,\ba) \rpa \exp\lpa - \eta \tilde{Q}(\bs,\ba,\bs') \rpa \rbr.
\end{align}
From the regression view, prior work MPO{\tiny~}\citep{mapo2018Abdolmaleki} infers actions with Q-value weighted regression, progressive approach compared to behavior cloning; however, such paradigm lacks the ability to capture transition dynamics. We explicitly introduce the $\exp (- \eta \tilde{Q}(\bs,\ba,\bs') )$ term, which as we show in experiments, is a crucial component for eliminating the  dynamics shift. 

\emph{Implementation:} In practice, we adopt offline samples in $\DT$ to approximate the true dynamics $\TT$ of $\MT$, and introduce a pair of binary classifiers, $q_{\text{sas}}(\cdot|\bs,\ba,\bs')$ and $q_{\text{sa}}(\cdot|\bs,\ba)$, to replace~$\log \frac{\HTS(\bs'|\bs,\ba)}{\TT(\bs'|\bs,\ba)}$ 
as in~\citet{offd2018Eysenbach}: $\log \frac{\HTS(\bs'|\bs,\ba)}{\TT(\bs'|\bs,\ba)} = \log \frac{q_{\text{sas}}(\text{\textcolor{blue}{source}}|\bs,\ba,\bs')}{q_{\text{sas}}(\text{\textcolor{red}{target}}|\bs,\ba,\bs')} - \log \frac{q_{\text{sa}}(\text{\textcolor{blue}{source}}|\bs,\ba)}{q_{\text{sa}}(\text{\textcolor{red}{target}}|\bs,\ba)}$. (See Appendix-\ref{appendix:learning-classifiers} for details). Although the amount of data $\DT$ sampled from the target $\MT$ is reduced in our~problem setup, we experimentally find that such classifiers are sufficient to achieve good~performance.  

\vspace{-2pt}
\subsection{Dynamics-Aware Reward Augmentation in Model-based Formulation}
\vspace{-2pt}

Following Equation~\ref{eq-motivation-model-based}, we then characterize the dynamics shift compensation term as in the above model-free analysis in the model-based offline formulation. We will find that across different derivations, our reward augmentation $\DR$ has always maintained the functional consistency and simplicity. 

Following MOPO, we assume $\mathcal{F} = \{f: \Vert f \Vert_\infty \leq 1\}$, then we have 
$
d_{\mathcal{F}}(\HTS(\bs'|\bs,\ba), \HTT(\bs'|\bs,\ba)) =  \TV(\HTS(\bs'|\bs,\ba), \HTT(\bs'|\bs,\ba)) 
\leq ({\KL(\HTS(\bs'|\bs,\ba), \HTT(\bs'|\bs,\ba))/2})^{\frac{1}{2}}
$, 
where $\TV$ is the total variance distance. 
Then we introduce the admissible error $\textcolor{red}{u}(\bs,\ba)$ such that $d_{\mathcal{F}}(\HTT(\bs'|\bs,\ba), \TT(\bs'|\bs,\ba))\leq \textcolor{red}{u}(\bs,\ba)$ for all $(\bs,\ba)$, 
and $\eta$ and $\delta$ such that $(\KL(\HTS,\HTT)/2)^{\frac{1}{2}} \leq \eta \KL(\HTS,\HTT) + \delta$. 
Following Lemma~\ref{lemma-model-based-error}, we thus can maximize the following lower bound with the samples in $\HMS$ ($\lambda := \frac{\gamma R_{max}}{1-\gamma} $): 
\vspace{-1pt}  
\begin{equation}\label{eq-model-based-formulation-lower-bound}
\eta_\MT(\pi) \geq \E_{\bs,\ba,\bs'\sim d_\HMS^\pi(\bs)\pi(\ba|\bs)\HTS(\bs'|\bs,\ba)}\lbr r(\bs,\ba) - \eta\lambda  \log \frac{\HTS(\bs'|\bs,\ba)}{\HTT(\bs'|\bs,\ba)} - \lambda \textcolor{red}{u}(\bs,\ba) - \lambda \delta \rbr. 
\end{equation}
\emph{Implementation:} We model the dynamics $\HTS$ and $\HTT$ with an ensemble of 2*N parameterized Gaussian distributions: $\mathcal{N}^i_\HTS(\mu_{\thetas}(\bs,\ba), \Sigma_{\phis}(\bs,\ba))$ and $\mathcal{N}^i_\HTT(\mu_\thetat(\bs,\ba), \Sigma_\phit(\bs,\ba))$, where $i\in[1,N]$. We approximate  $\textcolor{red}{u}$ with the maximum standard deviation of the learned models in the ensemble: $\textcolor{red}{u}(\bs,\ba) = \max_{i=1}^N\Vert \Sigma_\phit(\bs,\ba) \Vert_{\text{F}}$, omit the training-independent $\delta$, and treat $\lambda$ as a hyperparameter as in MOPO. For the $\log \frac{\HTS}{\HTT}$ term, we resort to the above classifiers ($q_{\text{sas}}$ and $q_{\text{sa}}$) in  model-free setting. 
(See~Appendix-\ref{appendix:comparsion-between-calssifiers-dynamics}  for comparison between using classifiers and estimated-dynamics~ratio.) 



\vspace{-1pt}
\section{Experiments}
\vspace{-1pt}

We present empirical demonstrations of our dynamics-aware reward augmentation (DARA) in a variety of settings. We start with two simple control experiments that illustrate the significance of DARA under the domain (dynamics) adaptation setting. Then we incorporate DARA into state-of-the-art (model-free and model-based) offline RL methods and evaluate the performance on the D4RL tasks. Finally, we compare our framework to several cross-domain-based baselines on simulated and real-world tasks. Note that for the dynamics adaptation, we also release a (source) dataset as a complement to D4RL, along with the quadruped robot dataset in simulator (source) and real (target). 

\vspace{-2pt}
\subsection{How does DARA handle the dynamics shift in offline setting?}
\begin{figure*}[h!]
	\vspace{-3pt}
	\centering
	\begin{minipage}[t]{0.490\linewidth}
		\centering
		\includegraphics[scale=0.415]{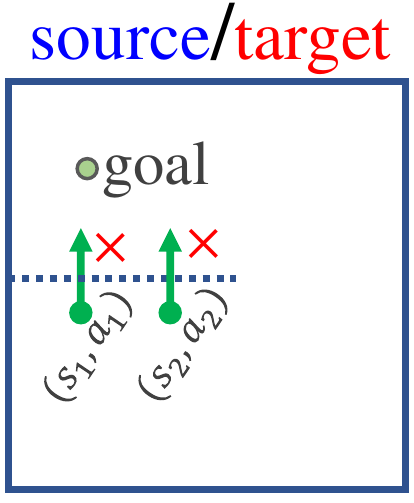}
		\includegraphics[scale=0.360]{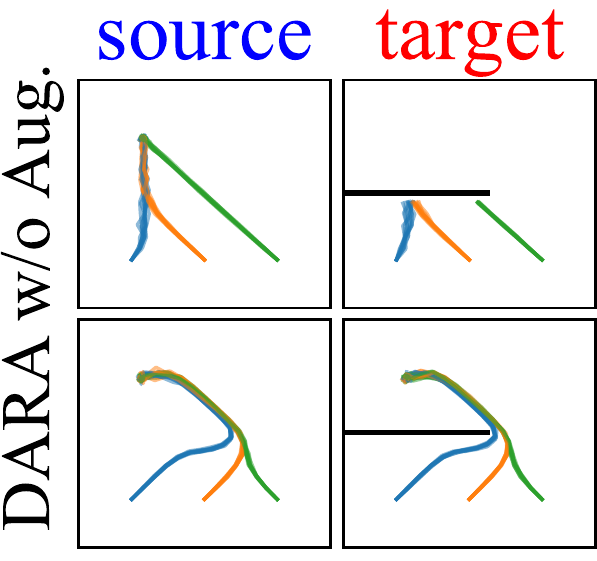} \ 
		\includegraphics[scale=0.365]{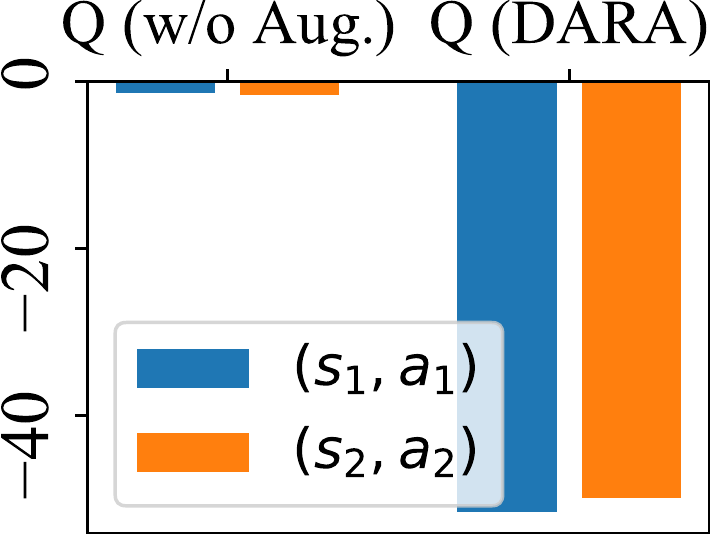}
		\vspace{-7pt}
		\caption{External dynamics shift: \emph{(left)} source and target MDPs (target contains an obstacle represented with the dashed line); \emph{(middle)} top plots (\emph{w/o Aug.}) depict the trajectories that are generated by the learned policy with vanilla MPO; \emph{(middle)} bottom plots (\emph{DARA}) depict the trajectories that are generated by the learned policy with DARA-based MPO; \emph{(right)} learned Q-values on the state-action pairs in {left subfigure}.}
		\label{fig-exp1-map}
	\end{minipage}
	\ \ \ \ 
	\begin{minipage}[t]{0.470\linewidth}
		\centering
		\includegraphics[scale=0.678]{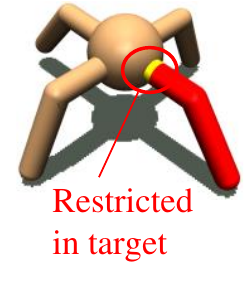}
		\includegraphics[scale=0.3875]{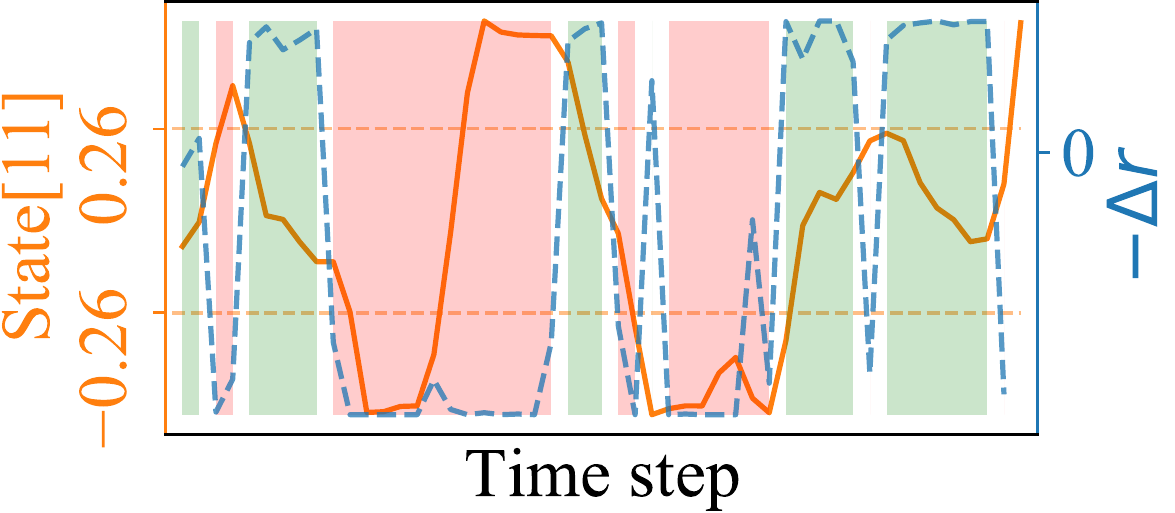}
		\vspace{-7pt}
		\caption{ Internal dynamics shift: \emph{(left)} source and target MDPs (range of the right-back-leg of the ant (state[11]) is limited: $[-0.52, 0.52]$ in source MDP $\to$ $[-0.26, 0.26]$ in target MDP); \emph{(right)} the solid (orange) line denotes the state of the right-back-leg over one trajectory collected in source, dashed (blue) line denotes the learned reward modification $-\DR$ over the trajectory, and green  and red slices denote transition pairs where $-\DR\geq$ and $-\DR < 0$, \emph{resp}.
		}
		\label{fig-exp1-ant}
	\end{minipage}
	\vspace{-7pt}
\end{figure*}
Here we characterize both external and internal dynamics shifts: In Map tasks (Figure~\ref{fig-exp1-map} left), the source dataset  $\DS$ is collected in a 2D map and the target $\DT$ is collected in the same environment but with an obstacle (the dashed line); In Ant tasks (Figure~\ref{fig-exp1-ant} left), the source dataset  $\DS$ is collected using the Mujoco Ant and the target $\DT$ is collected with the same Ant but one joint of which is~restricted.


Using MPO, as an example of offline RL method, we train a policy on dataset $\{\DS \cup \DT\}$ and deploy the acquired policy in both source and target MDPs. As shown in Figure~\ref{fig-exp1-map} (middle-top, \emph{w/o Aug.}), such training paradigm does not produce an adaptive policy for the target. By modifying rewards in source $\DS$, we show that applying the same training paradigm on the reward augmented data exhibits a positive transfer ability in Figure~\ref{fig-exp1-map} (middle-bottom, \emph{DARA}).  
In Figure~\ref{fig-exp1-map} (right), we show that our {DARA} produces low Q-values on the obstructive state-action pairs (in left) compared to the vanilla MPO, which thus prevents the Q-value weighted-regression on these unproductive state-action pairs. 

More generally, we illustrate how DARA can handle the dynamics adaptation from the reward modification view. 
In Figure~\ref{fig-exp1-ant} (right), the learned reward modification $-\DR$ (dashed {blue} line) clearly produces a penalty (red slices) on these state-action pairs (in source) that produce infeasible next-state transitions in the target MDP.
If we directly apply prior offline RL methods, these transitions that are beyond reach in target and yet are high valued, would yield a negative transfer. 
Thus, we can think of DARA as finding out these transitions that exhibit dynamics shifts and enabling dynamics adaptation with reward modifications, \eg penalizing transitions covered by red slices~($-\DR<0$).

\begin{table}[t]
	\centering
	\caption{Normalized scores for the (target) D4RL tasks, where our results are averaged {over{\tiny ~}5{\tiny ~}seeds.} 
		The arrows in each four-tuple indicate whether the current performance has improved (\textcolor{red}{$\uparrow$}) or not  (\textcolor{blue}{$\downarrow$}) compared to the previous value. 
		If   \emph{1T+10S DARA} achieves comparable (less than 10\%  degradation) or better performance compared to baseline \emph{10T}, we highlight our scores in bold (in each four-tuple). 
	}
	\begin{adjustbox}{max width=\textwidth}
		\begin{tabular}{clrrrrrrrrrrrr}
			\toprule 
			\multicolumn{2}{c}{\multirow{2}[1]{*}{\textcolor{black}{Body Mass Shift}}} & \multicolumn{1}{c}{\multirow{2}[1]{*}{10T}} & \multicolumn{1}{c}{\multirow{2}[1]{*}{1T}} & \multicolumn{1}{l}{1T+10S} & \multicolumn{1}{l}{1T+10S} & \multicolumn{1}{c}{\multirow{2}[1]{*}{10T}} & \multicolumn{1}{c}{\multirow{2}[1]{*}{1T}} & \multicolumn{1}{l}{1T+10S} & \multicolumn{1}{l}{1T+10S} & \multicolumn{1}{c}{\multirow{2}[1]{*}{10T}} & \multicolumn{1}{c}{\multirow{2}[1]{*}{1T}} & \multicolumn{1}{l}{1T+10S} & \multicolumn{1}{l}{1T+10S} \\
			\multicolumn{2}{c}{} &       &       & \multicolumn{1}{l}{w/o Aug.} & \multicolumn{1}{l}{DARA} &       &       & \multicolumn{1}{l}{w/o Aug.} & \multicolumn{1}{l}{DARA} &       &       & \multicolumn{1}{l}{w/o Aug.} & \multicolumn{1}{l}{DARA} \\
			\cmidrule{1-14}    \multirow{5}[4]{*}{\rotatebox{90}{\colorbox{magiccolor}{\quad Hopper \quad}}} &       & \multicolumn{4}{c}{BEAR}      & \multicolumn{4}{c}{BRAC-p}    & \multicolumn{4}{c}{AWR} \\
			\cmidrule(r){3-6}  \cmidrule(r){7-10} \cmidrule(r){11-14}           & Random & 11.4  & 1.0 \textcolor{blue}{$\downarrow$} & 4.6 \textcolor{red}{$\uparrow$} & 8.4 \textcolor{red}{$\uparrow$} & 11.0    & 10.9 \textcolor{blue}{$\downarrow$} & 9.6 \textcolor{blue}{$\downarrow$} & \textbf{11.0} \textcolor{red}{$\uparrow$} & 10.2  & 10.3 \textcolor{red}{$\uparrow$} & 3.4 \textcolor{blue}{$\downarrow$} & 4.5 \textcolor{red}{$\uparrow$} \\
			& Medium & 52.1  & 0.8 \textcolor{blue}{$\downarrow$} & 0.9 \textcolor{red}{$\uparrow$} & 1.6 \textcolor{red}{$\uparrow$} & 32.7  & 29.0 \textcolor{blue}{$\downarrow$} & 29.2 \textcolor{red}{$\uparrow$} & \textbf{32.9} \textcolor{red}{$\uparrow$} & 35.9  & 30.9 \textcolor{blue}{$\downarrow$} & 20.8 \textcolor{blue}{$\downarrow$} & 28.9 \textcolor{red}{$\uparrow$} \\
			& Medium-R & 33.7  & 1.3 \textcolor{blue}{$\downarrow$} & 18.2 \textcolor{red}{$\uparrow$} & \textbf{34.1} \textcolor{red}{$\uparrow$} & 0.6   & 5.4 \textcolor{red}{$\uparrow$} & 20.1 \textcolor{red}{$\uparrow$} & \textbf{30.8} \textcolor{red}{$\uparrow$} & 28.4  & 8.8 \textcolor{blue}{$\downarrow$} & 4.1 \textcolor{blue}{$\downarrow$} & 4.2 \textcolor{red}{$\uparrow$} \\
			& Medium-E & 96.3  & 0.8 \textcolor{blue}{$\downarrow$} & 0.6 \textcolor{blue}{$\downarrow$} & 1.2 \textcolor{red}{$\uparrow$} & 1.9   & 34.5 \textcolor{red}{$\uparrow$} & 32.3 \textcolor{blue}{$\downarrow$} & \textbf{34.7} \textcolor{red}{$\uparrow$} & 27.1  & 27.0 \textcolor{blue}{$\downarrow$} & 26.8 \textcolor{blue}{$\downarrow$} & \textbf{26.6} \textcolor{blue}{$\downarrow$} \\
			\cmidrule{1-14}    \multirow{5}[4]{*}{\rotatebox{90}{\colorbox{magiccolor}{\quad Hopper \quad}}} &       & \multicolumn{4}{c}{BCQ}       & \multicolumn{4}{c}{CQL}       & \multicolumn{4}{c}{MOPO} \\
			\cmidrule(r){3-6}  \cmidrule(r){7-10} \cmidrule(r){11-14}           & Random & 10.6  & 10.6 \textcolor{blue}{$\downarrow$} & 8.3 \textcolor{blue}{$\downarrow$} & \textbf{9.7} \textcolor{red}{$\uparrow$} & 10.8  & 10.6 \textcolor{blue}{$\downarrow$} & 10.2 \textcolor{blue}{$\downarrow$} & \textbf{10.4} \textcolor{red}{$\uparrow$} & 11.7  & 4.8 \textcolor{blue}{$\downarrow$} & 2.0 \textcolor{blue}{$\downarrow$} & 2.1 \textcolor{red}{$\uparrow$} \\
			& Medium & 54.5  & 37.1 \textcolor{blue}{$\downarrow$} & 25.7 \textcolor{blue}{$\downarrow$} & 38.4 \textcolor{red}{$\uparrow$} & 58.0    & 43.0 \textcolor{blue}{$\downarrow$} & 44.9 \textcolor{red}{$\uparrow$} & \textbf{59.3} \textcolor{red}{$\uparrow$} & 28.0    & 4.1 \textcolor{blue}{$\downarrow$} & 5.0 \textcolor{red}{$\uparrow$} & 10.7 \textcolor{red}{$\uparrow$} \\
			& Medium-R & 33.1  & 9.3 \textcolor{blue}{$\downarrow$} & 28.7 \textcolor{red}{$\uparrow$} & \textbf{32.8} \textcolor{red}{$\uparrow$} & 48.6  & 9.6 \textcolor{blue}{$\downarrow$} & 1.4 \textcolor{blue}{$\downarrow$} & 3.7 \textcolor{red}{$\uparrow$} & 67.5  & 1.0 \textcolor{blue}{$\downarrow$} & 5.5 \textcolor{red}{$\uparrow$} & 8.4 \textcolor{red}{$\uparrow$} \\
			& Medium-E & 110.9 & 58 \textcolor{blue}{$\downarrow$} & 75.4 \textcolor{red}{$\uparrow$} & 84.2 \textcolor{red}{$\uparrow$} & 98.7  & 59.7 \textcolor{blue}{$\downarrow$} & 53.6 \textcolor{blue}{$\downarrow$} & \textbf{99.7} \textcolor{red}{$\uparrow$} & 23.7  & 1.6 \textcolor{blue}{$\downarrow$} & 4.8 \textcolor{red}{$\uparrow$} & 5.8 \textcolor{red}{$\uparrow$} \\
			\cmidrule{1-14}    \multirow{5}[4]{*}{\rotatebox{90}{\colorbox{magiccolor}{Walker2d \quad}}} &       & \multicolumn{4}{c}{BEAR}      & \multicolumn{4}{c}{BRAC-p}    & \multicolumn{4}{c}{AWR} \\
			\cmidrule(r){3-6}  \cmidrule(r){7-10} \cmidrule(r){11-14}           & Random & 7.3   & 1.5 \textcolor{blue}{$\downarrow$} & 3.1 \textcolor{red}{$\uparrow$} & 3.2 \textcolor{red}{$\uparrow$} & -0.2  & 0.0 \textcolor{red}{$\uparrow$} & 1.3 \textcolor{red}{$\uparrow$} & \textbf{3.2} \textcolor{red}{$\uparrow$} & 1.5   & 1.3 \textcolor{blue}{$\downarrow$} & 2.0 \textcolor{red}{$\uparrow$} & \textbf{2.4} \textcolor{red}{$\uparrow$} \\
			& Medium & 59.1  & -0.5 \textcolor{blue}{$\downarrow$} & 0.6 \textcolor{red}{$\uparrow$} & 0.3 \textcolor{blue}{$\downarrow$} & 77.5  & 6.4 \textcolor{blue}{$\downarrow$} & 70.0 \textcolor{red}{$\uparrow$} & \textbf{78.0} \textcolor{red}{$\uparrow$} & 17.4  & 14.8 \textcolor{blue}{$\downarrow$} & 17.1 \textcolor{red}{$\uparrow$} & \textbf{17.2} \textcolor{red}{$\uparrow$} \\
			& Medium-R & 19.2  & 0.7 \textcolor{blue}{$\downarrow$} & 6.5 \textcolor{red}{$\uparrow$} & 7.3 \textcolor{red}{$\uparrow$} & -0.3  & 8.5 \textcolor{red}{$\uparrow$} & 9.9 \textcolor{red}{$\uparrow$} & \textbf{18.6} \textcolor{red}{$\uparrow$} & 15.5  & 7.4 \textcolor{blue}{$\downarrow$} & 1.6 \textcolor{blue}{$\downarrow$} & 1.5 \textcolor{blue}{$\downarrow$} \\
			& Medium-E & 40.1  & -0.1 \textcolor{blue}{$\downarrow$} & 1.5 \textcolor{red}{$\uparrow$} & 2.3 \textcolor{red}{$\uparrow$} & 76.9  & 20.6 \textcolor{blue}{$\downarrow$} & 64.1 \textcolor{red}{$\uparrow$} & \textbf{77.5} \textcolor{red}{$\uparrow$} & 53.8  & 35.5 \textcolor{blue}{$\downarrow$} & 52.5 \textcolor{red}{$\uparrow$} & \textbf{53.3} \textcolor{red}{$\uparrow$} \\
			\cmidrule{1-14}    \multirow{5}[4]{*}{\rotatebox{90}{\colorbox{magiccolor}{Walker2d \quad}}} &       & \multicolumn{4}{c}{BCQ}       & \multicolumn{4}{c}{CQL}       & \multicolumn{4}{c}{MOPO} \\
			\cmidrule(r){3-6}  \cmidrule(r){7-10} \cmidrule(r){11-14}           & Random & 4.9   & 1.8 \textcolor{blue}{$\downarrow$} & 4.5 \textcolor{red}{$\uparrow$} & \textbf{4.8} \textcolor{red}{$\uparrow$} & 7.0     & 1.7 \textcolor{blue}{$\downarrow$} & 3.2 \textcolor{red}{$\uparrow$} & 3.4 \textcolor{red}{$\uparrow$} & 13.6  & -0.2 \textcolor{blue}{$\downarrow$} & -0.1 \textcolor{red}{$\uparrow$} & -0.1 \textcolor{blue}{$\downarrow$} \\
			& Medium & 53.1  & 32.8 \textcolor{blue}{$\downarrow$} & 50.9 \textcolor{red}{$\uparrow$} & \textbf{52.3} \textcolor{red}{$\uparrow$} & 79.2  & 42.9 \textcolor{blue}{$\downarrow$} & 80.0 \textcolor{red}{$\uparrow$} & \textbf{81.7 }\textcolor{red}{$\uparrow$} & 17.8  & 7.0 \textcolor{blue}{$\downarrow$} & 5.7 \textcolor{blue}{$\downarrow$} & 11.0 \textcolor{red}{$\uparrow$} \\
			& Medium-R & 15.0    & 6.9 \textcolor{blue}{$\downarrow$} & 14.9 \textcolor{red}{$\uparrow$} & \textbf{15.1} \textcolor{red}{$\uparrow$} & 26.7  & 4.6 \textcolor{blue}{$\downarrow$} & 0.8 \textcolor{blue}{$\downarrow$} & 2.0 \textcolor{red}{$\uparrow$} & 39.0    & 5.1 \textcolor{blue}{$\downarrow$} & 3.1 \textcolor{blue}{$\downarrow$} & 14.2 \textcolor{red}{$\uparrow$} \\
			& Medium-E & 57.5  & 32.5 \textcolor{blue}{$\downarrow$} & 55.2 \textcolor{red}{$\uparrow$} & \textbf{57.2} \textcolor{red}{$\uparrow$} & 111.0   & 49.5 \textcolor{blue}{$\downarrow$} & 63.5 \textcolor{red}{$\uparrow$} & 93.3 \textcolor{red}{$\uparrow$} & 44.6  & 5.3 \textcolor{blue}{$\downarrow$} & 5.5 \textcolor{red}{$\uparrow$} & 17.2 \textcolor{red}{$\uparrow$} \\
			\bottomrule
		\end{tabular}
		\label{tab:exp-1}%
	\end{adjustbox}
	\vspace{-5pt}
\end{table}

\vspace{-3pt}
\subsection{Can DARA enable an adaptive policy with reduced offline data in target?}
\vspace{-3pt}
\label{sec:enable}

To characterize the offline dynamics shift, 
we consider the Hopper, Walker2d and Halfcheetah from the Gym-MuJoCo environment, using offline samples from D4RL as our target offline dataset. 
For the source dataset, we change the body mass of agents or add joint noise to the motion, and, similar to D4RL, collect the Random, Medium, Medium-R and Medium-E offline datasets for the three environments. 
Based on various offline RL algorithms (BEAR, BRAC-p, BCQ,  CQL, AWR, MOPO), we perform the following comparisons: 1) employing the $100\%$ of D4RL data (\emph{10T}), 2) employing only $10\%$ of the D4RL data (\emph{1T}), 3) employing $10\%$ of the D4RL data and $100\%$ of our collected source offline data (\emph{1T+10S w/o Aug.}), and 4) employing $10\%$ of the D4RL data and $100\%$ of our collected source offline data along with our reward augmentation (\emph{1T+10S DARA}). 
Due to page limit, here we focus on the dynamics shift concerning the body mass on Walker2d and Hopper. We refer the reader to appendix for more experimental details, tasks, and more baselines (BC, COMBO). 


As shown in Table~\ref{tab:exp-1}, in most of the tasks, the performance degrades substantially when we decrease the amount of target offline data, \ie \emph{10T} $\to$ \emph{1T}. 
Training with additional ten times source offline data (\emph{1T+10S w/o Aug.}) also does not 
bring substantial improvement (compensating for the reduced data in target), which even degrades the performance in some tasks. 
We believe that such degradation (compared to \emph{10T}) is caused by the lack of target offline data as well as the dynamics shift (induced by the source data). Incorporating our reward augmentation, we observe that compared to \emph{1T} and \emph{1T+10S w/o Aug.} that both use $10\%$ of the target offline data, our \emph{1T+10S DARA} significantly improves the performance across a majority of tasks. Moreover, DARA can  achieve~comparable or better performance compared to baseline \emph{10T} that training with ten times as much target~offline~data. 


\vspace{-2pt}
\subsection{Can DARA perform better than cross-domain  baselines?}
\vspace{-2pt}

\begin{table}[t]
	\centering
	\caption{Normalized scores in (target) D4RL tasks, where "Tune" denotes baseline "fine-tune". We observe that with same amount ($10\%$) of target offline data, DARA greatly outperforms~baselines. 
	}
	\vspace{-3pt}
	\begin{adjustbox}{max width=\textwidth}
		\begin{tabular}{clrrrrrrrrrrrr}
			\toprule 
			\multicolumn{2}{l}{\textcolor{black}{Body Mass Shift}} & Tune  & DARA  & Tune  & DARA  & Tune  & DARA  & Tune  & DARA  & Tune  & DARA  & $\pi_p\hat{T}$  & $\hat{T}\pi_p$ \\
			\midrule
			\multirow{5}[1]{*}{\rotatebox{90}{\colorbox{magiccolor}{\quad Hopper \quad}}} &       & \multicolumn{2}{c}{BEAR} & \multicolumn{2}{c}{BRAC-p} & \multicolumn{2}{c}{BCQ} & \multicolumn{2}{c}{CQL} & \multicolumn{2}{c}{MOPO} & \multicolumn{2}{c}{MABE} \\
			\cmidrule(r){3-4}  \cmidrule(r){5-6} \cmidrule(r){7-8} \cmidrule(r){9-10} \cmidrule(r){11-12}  \cmidrule(r){13-14} 
			& Random & 0.8   & 8.4 \textcolor{red}{$\uparrow$}   & 6.0     & 11.0 \textcolor{red}{$\uparrow$}  & 8.8   & 9.7 \textcolor{red}{$\uparrow$}   & \textbf{31.6}  & 10.4  \textcolor{blue}{$\downarrow$}   & 0.7   & 2.1 \textcolor{red}{$\uparrow$} & 10.6  & 9.0 \\
			& Medium & 0.8   & 1.6 \textcolor{red}{$\uparrow$}  & 22.7  & 32.9 \textcolor{red}{$\uparrow$}    & 31.7  & 38.4 \textcolor{red}{$\uparrow$} & 44.5  & \textbf{59.3} \textcolor{red}{$\uparrow$} & 0.7   & 10.7 \textcolor{red}{$\uparrow$} & 48.8  & 23.1  \\
			& Medium-R & 0.7   & \textbf{34.1} \textcolor{red}{$\uparrow$} & 14.7  & 30.8 \textcolor{red}{$\uparrow$} & 27.5  & 32.8 \textcolor{red}{$\uparrow$} & 1.3   & 3.7 \textcolor{red}{$\uparrow$}  & 0.6   & 8.4 \textcolor{red}{$\uparrow$} & 17.1  & 20.4 \\
			& Medium-E & 0.9   & 1.2 \textcolor{red}{$\uparrow$}  & 19.2  & 34.7 \textcolor{red}{$\uparrow$} & 85.9  & 84.2  \textcolor{blue}{$\downarrow$} & 47.6  & \textbf{99.7} \textcolor{red}{$\uparrow$} & 2.2   & 5.8 \textcolor{red}{$\uparrow$} & 28.1  & 38.9 \\
			\midrule
			\multirow{5}[1]{*}{\rotatebox{90}{\colorbox{magiccolor}{ Walker2d \quad}}} &       & \multicolumn{2}{c}{BEAR} & \multicolumn{2}{c}{BRAC-p} & \multicolumn{2}{c}{BCQ} & \multicolumn{2}{c}{CQL} & \multicolumn{2}{c}{MOPO} & \multicolumn{2}{c}{MABE} \\
			\cmidrule(r){3-4}  \cmidrule(r){5-6} \cmidrule(r){7-8} \cmidrule(r){9-10} \cmidrule(r){11-12}  \cmidrule(r){13-14} 
			& Random & \textbf{6.6}   & 3.2  \textcolor{blue}{$\downarrow$} & 3.9   & 3.2  \textcolor{blue}{$\downarrow$} & 4.7   & 4.8 \textcolor{red}{$\uparrow$} & 1.1   & 3.4 \textcolor{red}{$\uparrow$}  & 0.1   & -0.1  \textcolor{blue}{$\downarrow$} & 6.0     & -0.2 \\
			& Medium & 0.3   & 0.3  \textcolor{blue}{$\downarrow$}  & 76.0    & 78.0 \textcolor{red}{$\uparrow$} & 28.4  & 52.3 \textcolor{red}{$\uparrow$} & 72.3  & \textbf{81.7} \textcolor{red}{$\uparrow$} & -0.2  & 11.0 \textcolor{red}{$\uparrow$} & 30.1  & 56.7 \\
			& Medium-R & 1.2   & 7.3 \textcolor{red}{$\uparrow$}  & 10.0    & \textbf{18.6} \textcolor{red}{$\uparrow$} & 10.4  & 15.1 \textcolor{red}{$\uparrow$} & 1.8   & 2.0  \textcolor{red}{$\uparrow$} & 0.0     & 14.2 \textcolor{red}{$\uparrow$}   & 13.3  & 12.5 \\
			& Medium-E & 2.4   & 2.3 \textcolor{blue}{$\downarrow$} & 74.5  & 77.5 \textcolor{red}{$\uparrow$}   & 22.7  & 57.2 \textcolor{red}{$\uparrow$} & 68.6  & \textbf{93.3} \textcolor{red}{$\uparrow$} & 7.3   & 17.2  \textcolor{red}{$\uparrow$}  & 43.7  & 82.7 \\
			\bottomrule
		\end{tabular}%
	\end{adjustbox}
	\label{tab:exp-2}%
	\vspace{-8pt}
\end{table}

In Section~\ref{sec:enable}, \emph{1T+10S w/o Aug.} does not explicitly learn policy for the target dynamics, thus one proposal (\emph{1T+10S fine-tune}) for adapting the target dynamics is fine-tuning the model that learned with source offline data, using the (reduced) target offline data. Moreover, we also compare DARA with the recently proposed MABE~\citep{behavioral2021Cang}, which is suitable well for our cross-dynamics setting by introducing behavioral priors $\pi_p$ in the model-based offline setting. Thus, we implement two baselines, 1) \emph{1T+10S MABE $\pi_p\hat{T}$} and 2) \emph{1T+10S MABE $\hat{T}\pi_p$}, which denote 1) learning $\pi_p$ with target domain data and $\hat{T}$ with source domain data, and 2) learning $\pi_p$ with source domain data and $\hat{T}$ with target domain data, respectively. 
We show the results for the Walker (with body mass shift) in Table~\ref{tab:exp-2}, and more experiments in Appendix~\ref{appendix:comparsion-with-cross-domain}.  Our results show that DARA achieves significantly better performance than the naïve fine-tune-based approaches in a majority of tasks (67  "\textcolor{red}{$\uparrow$}" vs. 13 "\textcolor{blue}{$\downarrow$}", including results in appendix). On twelve out of the sixteen tasks (including results in appendix), DARA-based methods outperform the MABE-based methods. We attribute MABE's failure to the difficulty of the reduced target offline data, which limits the generalization of the learned $\pi_p$ or $\hat{T}$ under such data. However, such reduced data ($10\%$ of target) is sufficient to modify rewards in the source offline data, which thus encourages better performance for our DARA.

\begin{wraptable}{r}{0.333\textwidth}
	\vspace{-13.5pt} 
	\centering
	\caption{Average distance  covered in an episode in real  robot.} %
	\vspace{-7pt} 
	\begin{adjustbox}{max width=0.333\textwidth}
	\begin{tabular}{lcc}
		\toprule
		(BCQ) & {w/o Aug.} & {DARA} \\ 
		\midrule
		Medium & 0.85  & 1.35 \textcolor{red}{$\uparrow$} \\
		Medium-E & 1.15   & 1.41 \textcolor{red}{$\uparrow$} \\
		Medium--R-E  & 1.27    & 1.55 \textcolor{red}{$\uparrow$} \\
		\bottomrule
	\end{tabular}%
	\end{adjustbox}
	\label{tab:exp-3-real-dog}%
	\vspace{-11pt}
\end{wraptable}
\emph{\textbf{For real-world tasks}}, we also test DARA in a new offline dataset on the quadruped robot (see appendix for details). Note that we can not access  the privileged information (\eg coordinate) in real robot, thus the target offline data (collected in real-world) does not contain rewards. 
This means that prior fine-tune-based and MABE-based methods become unavailable. However, our reward augmentation frees us from the requisite of rewards in target domain. We can freely perform offline training only using the augmented source offline data as long as the learned $\DR$ is sufficient. For comparison, we also employ a baseline (\emph{w/o Aug.}): directly deploying the learned policy with source data into the (target) real-world. 
We present the results (deployed in real with obstructive stairs) in Table~\ref{tab:exp-3-real-dog} and videos in supplementary material.~We can observe that training with our reward augmentation, the performance can be substantially improved. 
Due to page limit, we refer readers to Appendix~\ref{appendix:sec:additional-results-on-dog} for more experimental results~and~discussion. 




\vspace{-4pt}
\section{Conclusion}
\vspace{-5pt}
In this paper, we formulate the dynamics shift in offline RL. 
Based on prior model-based and model-free offline algorithms, we propose the dynamics-award reward augmentation (DARA) framework that characterizes constraints over state-action-next-state distributions. Empirically we demonstrate DARA can eliminate the dynamics shift and outperform baselines in simulated and real-world tasks. 

In Appendix~\ref{appendix:regularization-view}, we characterize our dynamics-aware reward augmentation from the density regularization view, which shows that it is straightforward to derive the reward modification built on prior regularized max-return objective \eg AlgaeDICE~\citep{algaedice2019nachum}. We list some related~works in Table~\ref{table-rl-regularization}, where the majority of the existing work focuses on regularizing state-action distribution, while dynamics shift receives relatively little attention. Thus, we hope to shift the focus of the community towards analyzing how dynamics shift affects RL and how to eliminate the effect.


%

\subsubsection*{Reproducibility Statement}
Our experimental evaluation is conducted with publicly available D4RL~\citep{d4rl2020Fu} and NeoRL~\citep{qin2021neorl}. In Appendix~\ref{appendix:environments-and-dataset} and~\ref{appendix:sec:training-a1-dog}, we provide the environmental details and training setup for our real-world sim2real tasks. In supplementary material, we upload our source code and the collected offline dataset for the the quadruped robot. 

\subsubsection*{Acknowledgments}
We thank Zifeng Zhuang, Yachen Kang and Qiangxing Tian for helpful feedback and discussions. This work is supported by NSFC General Program (62176215).

\bibliography{iclr2022_conference}
\bibliographystyle{iclr2022_conference}

\newpage
\appendix
\section{Appendix}

\subsection{Derivation}

\subsubsection{Proof of Lemma 3}

Let $\B_M^\pi V(\bs) = \E_{\ba\sim\pi(\ba|\bs)}\lbr r(\bs,\ba) + \gamma \E_{\bs'\sim T(\bs'|\bs,\ba)}\lbr V(\bs') \rbr \rbr$ and $r(\bs) = \E_{\ba\sim\pi(\ba|\bs)}\lbr r(\bs,\ba) \rbr$. Then, we have 
\begin{align}
\eta_{\HMS}(\pi) - \eta_{\MT}(\pi) 
&=  \E_{\bs_0\sim\rho_0(\bs)}\lbr V_\HMS(\bs_0) - V_\MT(\bs_0) \rbr \nonumber \\
&= \sum_{t=0}^{\infty} \gamma^t \E_{\bs_t\sim P(\bs_t|\pi,\HMS,t)}\E_{\ba_t\sim \pi(\ba_t|\bs_t)} \lbr r(\bs_t,\ba_t) \rbr - \E_{\bs_0\sim\rho_0(\bs)}\lbr V_\MT(\bs_0) \rbr \nonumber \\
&= \sum_{t=0}^{\infty} \gamma^t \E_{\bs_t\sim P(\bs_t|\pi,\HMS,t)}\lbr r(\bs_t) + V_\MT(\bs_t) - V_\MT(\bs_t)  \rbr - \E_{\bs_0\sim\rho_0(\bs)}\lbr V_\MT(\bs_0) \rbr \nonumber \\
&= \sum_{t=0}^{\infty} \gamma^t \E_{\begin{subarray}{l}\bs_t\sim P(\bs_t|\pi,\HMS,t)\\ \bs_{t+1}\sim P(\bs_{t+1}|\pi,\HMS,t+1)
	\end{subarray}}\lbr r(\bs_t) + \gamma V_\MT(\bs_{t+1}) - V_\MT(\bs_t)  \rbr \nonumber \\
&= \sum_{t=0}^{\infty} \gamma^t \E_{\begin{subarray}{l}\bs_t\sim P(\bs_t|\pi,\HMS,t)\\ \bs_{t+1}\sim P(\bs_{t+1}|\pi,\HMS,t+1)
	\end{subarray}}\lbr r(\bs_t) + \gamma V_\MT(\bs_{t+1}) - \lpa r(\bs_t) + \gamma \E_{\ba\sim\pi(\ba|\bs_t),\bs'\sim\TT(\bs_t,\ba)}\lbr V_\MT(\bs')\rbr \rpa  \rbr \nonumber \\
&= \sum_{t=0}^{\infty} \gamma^t \E_{\bs_t\sim P(\bs_t|\pi,\HMS,t)}\lbr \B_\HMS^\pi V_\MT(\bs_t) -  \B_\MT^\pi V_\MT(\bs_t)  \rbr \nonumber \\
&= \E_{\bs\sim d_\HMS^\pi(\bs)} \lbr \B_\HMS^\pi V_\MT(\bs) - \B_\MT^\pi V_\MT(\bs) \rbr. \nonumber
\end{align}

\subsubsection{Model-based formulation}
\label{appendix:model-based-formualation}

Here we provide detailed derivation of the lower bound in Equation~\ref{eq-model-based-formulation-lower-bound} in the main text. 

\begin{assumption}\label{appendix:assumption:scale-c}
Assume a scale $c$ and a function class $\mathcal{F}$ such that $V_\MT \in c \mathcal{F}$. 
\end{assumption}

Following MOPO~\citep{mopo2020Yu}, we set $\mathcal{F} = \{f: \Vert f \Vert_\infty \leq 1\}$. In Section Preliminaries, we have that the reward function is bounded: $r(\bs,\ba) \in [-R_{max}, R_{max}]$. Thus, we have $\Vert V_\MT \Vert_{\infty} \leq \sum_{t=0}^{\infty} \gamma^t R_{max} = \frac{R_{max}}{1-\gamma}$ and hence the scale $c = \frac{R_{max}}{1-\gamma} $. 

As a direct corollary of Assumption~\ref{appendix:assumption:scale-c} and Equation~\ref{eq-motivation-model-based-upper}, we have 
\begin{equation} 
\begin{split}
\left\vert \E_{\bs\sim d_\HMS^\pi(\bs)} \lbr \B_\HMS^\pi V_\MT(\bs) - \B_\MT^\pi V_\MT(\bs) \rbr \right\vert \leq \gamma c \cdot \E_{\bs,\ba\sim d_\HMS^\pi(\bs)\pi(\ba|\bs)} \lbr  d_{\mathcal{F}}(\HTS(\bs'|\bs,\ba), \TT(\bs'|\bs,\ba))  \rbr. \label{appendix:model-based-derivation-upper-exp} 
\end{split}
\end{equation}

Further, we find
\begin{align}
d_{\mathcal{F}}(\HTS(\bs'|\bs,\ba), \TT(\bs'|\bs,\ba)) 
&\leq d_{\mathcal{F}}(\HTS(\bs'|\bs,\ba), \HTT(\bs'|\bs,\ba)) + d_{\mathcal{F}}(\HTT(\bs'|\bs,\ba), \TT(\bs'|\bs,\ba))  \label{appendix:model-based-derivation-upper-ff}
\end{align}

For the first term $d_{\mathcal{F}}(\HTS(\bs'|\bs,\ba), \HTT(\bs'|\bs,\ba))$ in Equation~\ref{appendix:model-based-derivation-upper-kl},  through Pinsker’s inequality, we have
\begin{align}
d_{\mathcal{F}}(\HTS(\bs'|\bs,\ba), \HTT(\bs'|\bs,\ba)) =  \TV(\HTS(\bs'|\bs,\ba), \HTT(\bs'|\bs,\ba)) 
\leq \sqrt{\frac{1}{2}\KL(\HTS(\bs'|\bs,\ba), \HTT(\bs'|\bs,\ba))} \label{appendix:model-based-derivation-upper-kl} 
\end{align}
To keep consistent with the DARA-based method-free offline methods, we introduce scale  $\eta$ and bias $\delta$ to eliminate the square root in Equation~\ref{appendix:model-based-derivation-upper-kl}. To be specific, we assume\footnote{In implementation, we clip the maximum deviation of $\log \frac{\HTS(\bs'|\bs,\ba)}{\HTT(\bs'|\bs,\ba)}$ for each $(\bs,\ba,\bs')$, which thus makes $\KL(\HTS(\bs'|\bs,\ba), \HTT(\bs'|\bs,\ba))$ bounded.} scale $\eta$ and bias $\delta$ such that $\sqrt{\frac{1}{2}\KL(\HTS, \HTT)} \leq \eta \KL(\HTS, \HTT) + \delta$. Thus, we obtain
\begin{align}
d_{\mathcal{F}}(\HTS(\bs'|\bs,\ba), \HTT(\bs'|\bs,\ba)) =  \TV(\HTS(\bs'|\bs,\ba), \HTT(\bs'|\bs,\ba)) 
\leq \eta \KL(\HTS(\bs'|\bs,\ba), \HTT(\bs'|\bs,\ba)) + \delta  \label{appendix:model-based-derivation-upper-klnosqrt} 
\end{align}

For the second term $d_{\mathcal{F}}(\HTT(\bs'|\bs,\ba), \TT(\bs'|\bs,\ba))$ in Equation~\ref{appendix:model-based-derivation-upper-ff}, we assume that we have access to an oracle uncertainty qualification module that provides an upper bound on the error of the estimated empirical MDP $\HMT := \{ \mathcal{S}, \mathcal{A}, r, \HTT, \rho_0, \gamma \}$. 
\begin{assumption} 
Let $\mathcal{F}$ be the function class in Assumption~\ref{appendix:assumption:scale-c}. We say $\textcolor{red}{u}:\mathcal{S}\times\mathcal{A} \to \mathbb{R}$ is an admissible error estimator for $\HTT$ if $d_{\mathcal{F}}(\HTT(\bs'|\bs,\ba), \TT(\bs'|\bs,\ba))\leq \textcolor{red}{u}(\bs,\ba)$ for all $(\bs,\ba)$. 
\end{assumption}
Thus, we  have 
\begin{align}
\E_{\bs,\ba\sim d_\HMS^\pi(\bs)\pi(\ba|\bs)} \lbr  d_{\mathcal{F}}(\HTT(\bs'|\bs,\ba), \TT(\bs'|\bs,\ba))  \rbr 
&\leq  \E_{\bs,\ba\sim d_\HMS^\pi(\bs)\pi(\ba|\bs)} \lbr \textcolor{red}{u}(\bs,\ba) \rbr  \label{appendix:model-based-derivation-upper-f}
\end{align}

Bring Inequations~\ref{appendix:model-based-derivation-upper-exp}, \ref{appendix:model-based-derivation-upper-ff}, \ref{appendix:model-based-derivation-upper-klnosqrt}, and \ref{appendix:model-based-derivation-upper-f} into Lemma~\ref{lemma-model-based-error}, we thus have 
\begin{equation} 
\eta_\MT(\pi) \geq \E_{\bs,\ba,\bs'\sim d_\HMS^\pi(\bs)\pi(\ba|\bs)\HTS(\bs'|\bs,\ba)}\lbr r(\bs,\ba) - \eta \gamma c  \log \frac{\HTS(\bs'|\bs,\ba)}{\HTT(\bs'|\bs,\ba)} -  \gamma c \textcolor{red}{u}(\bs,\ba) -  \gamma c \delta \rbr. 
\end{equation}


\subsubsection{Learning Classifiers}
\label{appendix:learning-classifiers}

Applying Bayes' rule, we have 
\begin{align}
\HTS(\bs'|\ba,\bs) &:= p(\bs'|\bs,\ba,\text{\textcolor{blue}{source}}) = \frac{p(\text{\textcolor{blue}{source}}|\bs,\ba,\bs')p(\bs,\ba,\bs')}{p(\text{\textcolor{blue}{source}}|\bs,\ba)p(\bs,\ba)}, \nonumber \\
\HTT(\bs'|\ba,\bs) &:= p(\bs'|\bs,\ba,\text{\textcolor{red}{target}}) = \frac{p(\text{\textcolor{red}{target}}|\bs,\ba,\bs')p(\bs,\ba,\bs')}{p(\text{\textcolor{red}{target}}|\bs,\ba)p(\bs,\ba)}. \nonumber 
\end{align}
Then we parameterize $p(\cdot|\bs,\ba,\bs')$ and $p(\cdot|\bs,\ba)$ with the two classifiers $q_{\text{sas}}$ and $q_{\text{sa}}$ respectively. 
Using the standard cross-entropy loss, we learn $q_{\text{sas}}$ and $q_{\text{sa}}$ with the following optimization objective: 
\begin{align}
\max &\ \quad  \E_{(\bs,\ba,\bs')\sim\DS} \lbr \log q_{\text{sas}}(\text{\textcolor{blue}{source}}|\bs,\ba,\bs') \rbr + \E_{(\bs,\ba,\bs')\sim\DT} \lbr \log q_{\text{sas}}(\text{\textcolor{red}{target}}|\bs,\ba,\bs') \rbr, \nonumber \\
\max &\ \quad \E_{(\bs,\ba)\sim\DS} \lbr \log q_{\text{sa}}(\text{\textcolor{blue}{source}}|\bs,\ba) \rbr + \E_{(\bs,\ba)\sim\DT} \lbr \log q_{\text{sa}}(\text{\textcolor{red}{target}}|\bs,\ba) \rbr. \nonumber
\end{align}
With the trained $q_{\text{sas}}$ and $q_{\text{sa}}$, we have
\begin{align}
\log \frac{\HTS(\bs'|\bs,\ba)}{\HTT(\bs'|\bs,\ba)} = \log \frac{q_{\text{sas}}(\text{\textcolor{blue}{source}}|\bs,\ba,\bs')}{q_{\text{sas}}(\text{\textcolor{red}{target}}|\bs,\ba,\bs')} - \log \frac{q_{\text{sa}}(\text{\textcolor{blue}{source}}|\bs,\ba)}{q_{\text{sa}}(\text{\textcolor{red}{target}}|\bs,\ba)}.
\end{align}
In our implementation, we also clip the above reward modification between $-10$ and $10$.


\subsection{Regularization View of Dynamics-Aware Reward Augmentation}
\label{appendix:regularization-view}
Here we shortly characterize our dynamics-aware reward augmentation from the density regularization. Note the standard max-return objective $\eta_\MT(\pi)$ in RL can be written exclusively in terms of the on-policy distribution $d_\MT^\pi(\bs)\pi(\ba|\bs)$. To introduce an off-policy distribution $d_\DT(\bs)\pibt(\ba|\bs)$ in the objective, prior works often incorporate a regularization (penalty): $ D ( d_\MT^\pi(\bs)\pi(\ba|\bs) \Vert d_\DT(\bs)\pibt(\ba|\bs) )$, as in Equations~\ref{eq-model-free-D} and~\ref{eq-cql}. However, facing dynamics shift, such regularization should take into account the transition dynamics, which is penalizing $D ( d_\MT^\pi(\bs)\pi(\ba|\bs)\TT(\bs'|\bs,\ba) \Vert d_\DS(\bs)\pibs(\ba|\bs)\HTS(\bs'|\bs,\ba) )$. 
From this view, it is also straightforward to derive the reward modification built on prior regularized off-policy max-return objective \eg the off-policy approach AlgaeDICE~\citep{algaedice2019nachum}. 

In Table~\ref{table-rl-regularization}, we provide some related works with respect to the (state-action pair) $d_\DT(\bs)\pibt(\ba|\bs)$ regularization and the (state-action-next-state pair) $d_\DS(\bs)\pibs(\ba|\bs)\HTS(\bs'|\bs,\ba)$ regularization. 
We can find that the majority of the existing work focuses on regularizing state-action distribution, while dynamics shift receives relatively little attention. Thus, we hope to shift the focus of the community towards analyzing how the dynamics shift affects RL and how to eliminate the effect.

\begin{table}[H]
	\caption{Some related works with \textbf{explicit} (state-action $p(\bs,\ba)$ or state-action-next-state $p(\bs,\ba,\bs')$) \textbf{regularization}. More papers with respect to unsupervised RL, inverse RL (imitation learning), meta RL, multi-agent RL, and hierarchical RL are not included.} 
	\label{table-rl-regularization}
	\begin{center}
		\begin{tabular}{p{0.375\textwidth}p{0.40\textwidth}||p{0.33\textwidth}}
			\multicolumn{2}{c||}{\bf reg. with $d_\DT(\bs)\pibt(\ba|\bs)$}		&\multicolumn{1}{c}{\bf reg. with $d_\DS(\bs)\pibs(\ba|\bs)\HTS(\bs'|\bs,\ba)$}
			\\ \hline  \hline 
			\emph{Online:} & &  \\
			\textcolor{red}{see summarization in \citet{kl2019Geist} and \citet{kl2020Vieillard}. }& & \citet{offd2018Eysenbach} (DARC) \citet{liu2021unsupervised} (DARS); \\ 
			 &  & \citet{DAmbrl2021Haghgoo} \\
			& & \\
			\emph{Offline (off-policy evaluation):} & & \\
			\citet{off2019Fujimoto} (BCQ); & \citet{stabilizing2019Kumar} (BEAR); & \\
			\citet{behavior2019Wu} (BRAC-p); & \citet{mapo2018Abdolmaleki} (MPO); & \\
			\citet{awr2019Peng} (AWR); & \citet{accelerating2020nair} (AWAC); & \\
			\citet{crr2020wang} (CRR); & \citet{keep2020siegel}; & \\ 
			\citet{bail2019chen}; (BAIL) & \citet{cql2020Kumar} (CQL); & \\
			\citet{cpq2021Xu} (CPQ); & \citet{offfisher2021Kostrikov} (Fisher-BRC); & \\
			\citet{breaking2018liu}; & \citet{dualdice2019nachum} (DualDICE); & \\ 
			\citet{algaedice2019nachum} (AlgaeDICE); & \citet{gnedice2020zhang} (GenDICE); & \\
			\citet{off2020yang}; & \citet{reinforcement2020nachum}; & \\
			\citet{minimax2020Jiang}; & \citet{minimax2020Uehara}; & \\
			\citet{mopo2020Yu} (MOPO); & \citet{morel2020Kidambi} (MOReL); & \\ 
			\citet{combo2021Yu} (COMBO); & \citet{behavioral2021Cang} (MABE); & \\

		\end{tabular}
	\end{center}
\end{table}

\subsection{More Experiments}

\subsubsection{Training with $\{\DS \cup \DT \}$}
\label{appendix:trainingwithdsdt}
\begin{figure*}[h!]
	\centering
		\includegraphics[scale=0.235]{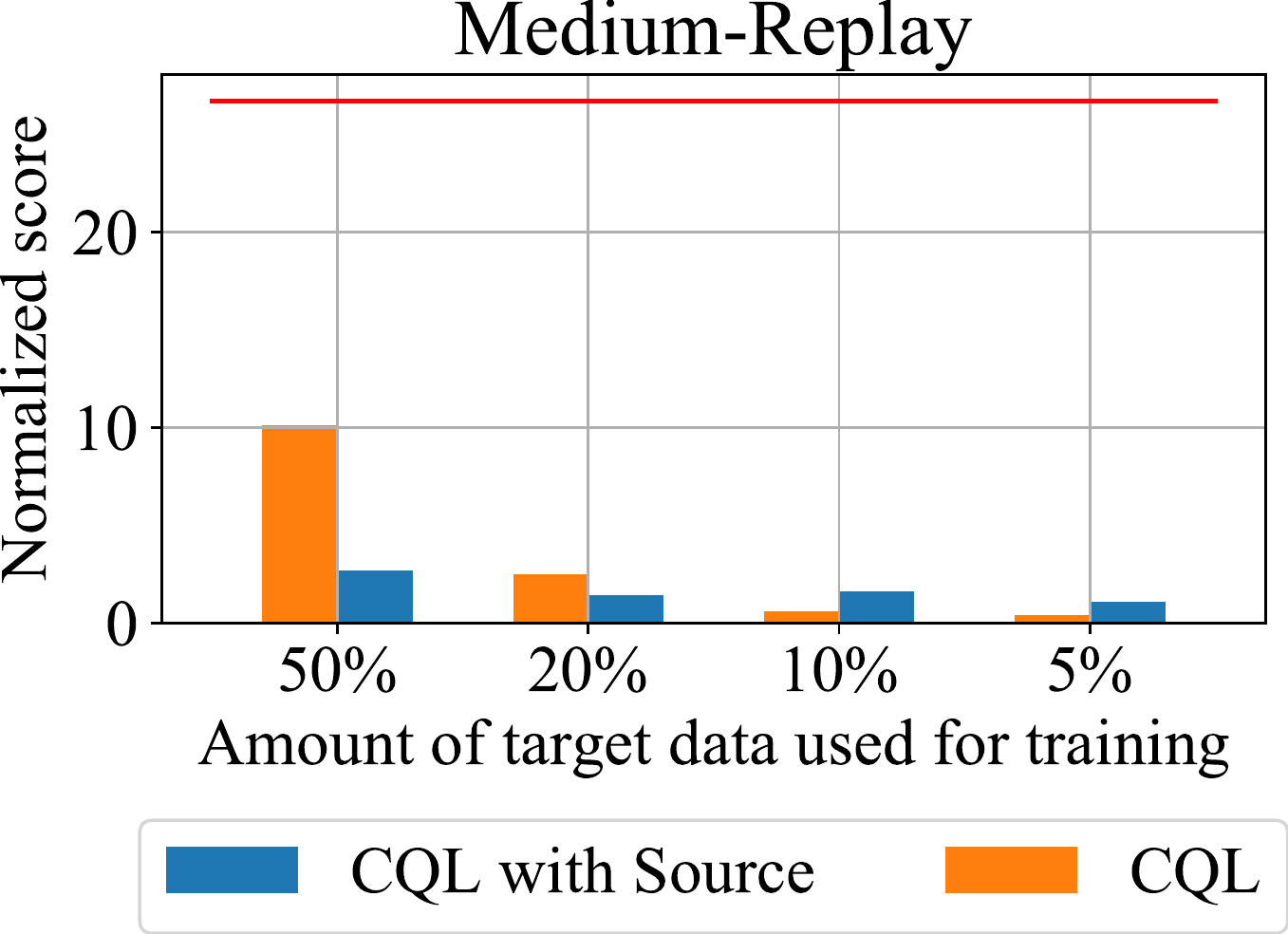}
		\includegraphics[scale=0.235]{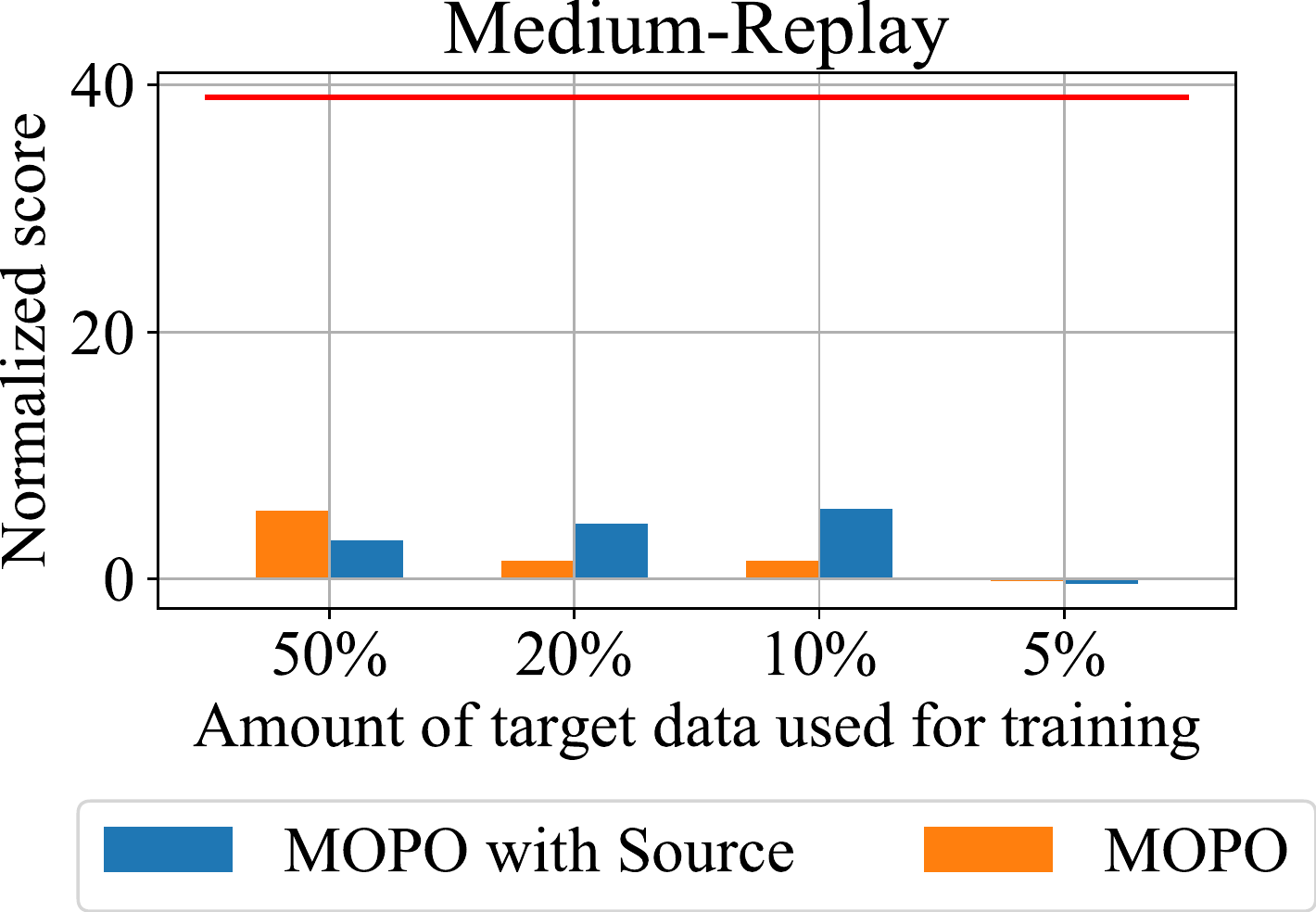}
		\includegraphics[scale=0.235]{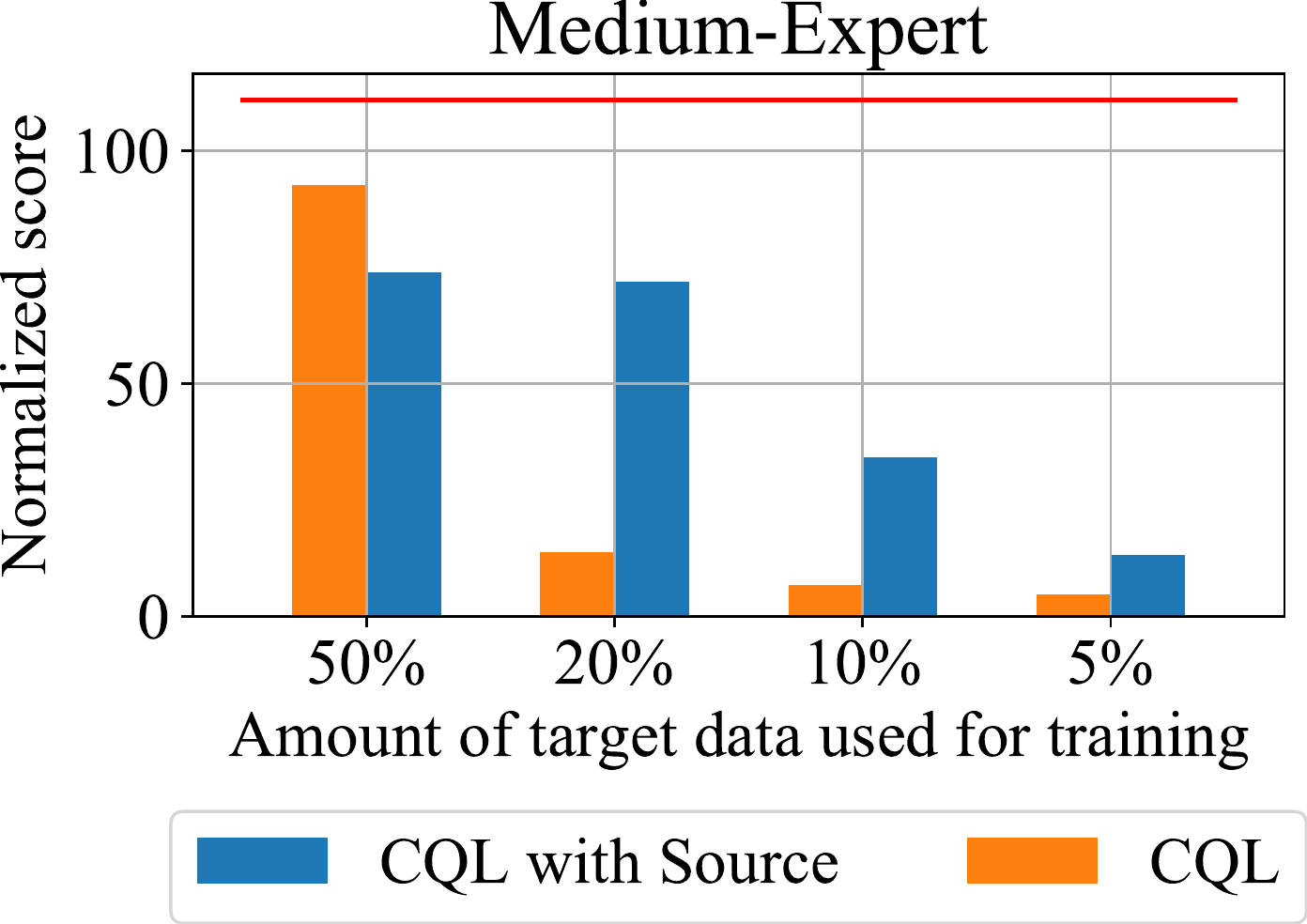}
		\includegraphics[scale=0.235]{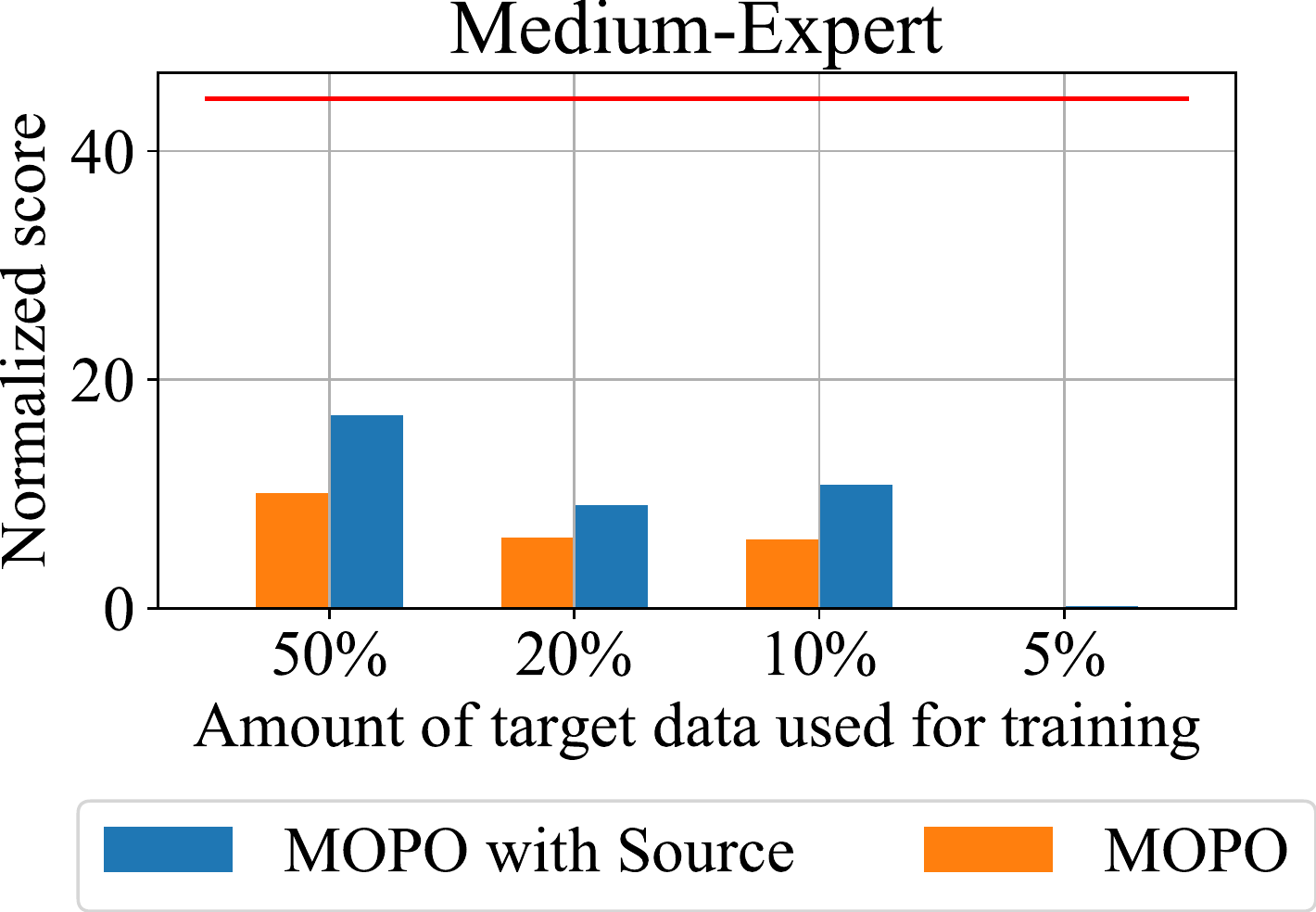}
		\caption{Final performance on the D4RL (Walker2d) task: The orange bars denote the final performance with different amount ($50\% \DT$, $20\% \DT$, $10\% \DT$, $5\% \DT$) of target offline data; The blue bars denote the final performance of mixing $100\%$ of source offline data $\DS$ and different amount of target data $x\% \DT$ ($x \in [50,20,10,5]$), \ie training with $\{100\%\DS \cup  x\% \DT\}$; The red lines denote  the final performance of training with $100\%$ of target offline data $\DT$. 
			We can observe that 1) the performance deteriorates dramatically as the amount of (target) offline data decreases ($100\%\DT \text{ (red line)} \to 50\%\DT  \text{ (orange bar)} \to 20\%\DT \text{ (orange bar)} \to 10\%\DT \text{ (orange bar)} \to 5\%\DT \text{ (orange bar)}$), 2) after training with the additional $100\%$ of source offline data, $\{100\%\DS \cup  x\% \DT\}$, the final performance is improved in some tasks, but most of the improvement is a pittance compared to the original performance degradation (compared to that training with the $100\%$ of target offline data, \ie the red lines), 
			and 3) what is worse is that adding source offline data $\DS$ even leads performance degradation in some tasks, \eg CQL with $50\%\DT$ and $20\%\DT$ in Medium-Random. 
		}
	\label{appendix:figure-introduction-plus10s}
\end{figure*}

As we show in Figure~\ref{fig-introduction-data-reduce} in Section Introduction, the performance of prior offline RL methods deteriorates dramatically as the amount of (target) offline data $\DT$ decreases. In Figure~\ref{appendix:figure-introduction-plus10s}, we show that directly training with the mixed dataset $\{ \DS \cup \DT \}$ will not compensate for the deteriorated performance caused by the reduced target offline data, and  training with such additional source offline data can even lead the performance degradation in some tasks. 

\subsubsection{Comparison between learning classifiers and learning dynamics (for the reward modification)} 
\label{appendix:comparsion-between-calssifiers-dynamics}

\begin{table}[H]
	\centering
	\caption{Normalized scores for the Hopper tasks with the body mass (dynamics) shift. Rat. and Cla. denote estimating the reward modification with the estimated-dynamics ratio and learned classifiers (Appendix~\ref{appendix:learning-classifiers}), respectively.}
	\begin{adjustbox}{max width=\textwidth}
	\begin{tabular}{llcccccccccccccccccc}
		\toprule
		\multicolumn{2}{c}{\multirow{1}[1]{*}{\textcolor{black}{Body Mass Shift}}}  & \multicolumn{3}{c}{BEAR} & \multicolumn{3}{c}{BRAC-p} & \multicolumn{3}{c}{AWR} & \multicolumn{3}{c}{BCQ} & \multicolumn{3}{c}{CQL} & \multicolumn{3}{c}{MOPO} \\
		\midrule
		\multirow{5}[0]{*}{\rotatebox{90}{\colorbox{magiccolor}{\quad Hopper \quad}}} &       & Rat. &       & Cla.  & Cla. &       & Cla.  & Rat. &       & Cla.  & Rat. &       & Cla.  & Rat. &       & Cla.  & Rat. &       & Cla. \\
		\cmidrule(r){3-5}  \cmidrule(r){6-8} \cmidrule(r){9-11} \cmidrule(r){12-14}  \cmidrule(r){15-17} \cmidrule(r){18-20}
		& Random & 9.9   & $>$ & 8.4   & 11.2  & $>$ & 11.0  & 3.7   & $<$ & 4.5   & 8.5   & $<$ & 9.7   & 11.8  & $>$ & 10.4  & 1.8   & $<$ & 2.1  \\
		& Medium & 0.8   & $<$ & 1.6   & 31.7  & $<$ & 32.9  & 18.0  & $<$ & 28.9  & 33.2  & $<$ & 38.4  & 45.9  & $<$ & 59.3  & 3.1   & $<$ & 10.7  \\
		& Medium-R & 28.4  & $<$ & 34.1  & 36.5  & $>$ & 30.8  & 2.5   & $<$ & 4.2   & 33.9  & $>$ & 32.8  & 2.0   & $<$ & 3.7   & 3.8   & $<$ & 8.4  \\
		& Medium-E & 0.8   & $<$ & 1.2   & 50.9  & $>$ & 34.7  & 45.8  & $<$ & 26.6  & 68.4  & $<$ & 84.2  & 107.3  & $>$ & 99.7  & 5.7   & $<$ & 5.8  \\
		\bottomrule 
	\end{tabular}%
	\label{appendix:tab:comparsion-between-calssifiers-dynamics}%
\end{adjustbox}
\end{table}%

In Table~\ref{appendix:tab:comparsion-between-calssifiers-dynamics}, we show the comparison between learning classifiers and learning dynamics (for our reward modification) in the Hopper tasks. We can observe that the two schemes for estimating the reward modification have similar performance. Thus, for simplicity and following~\cite{offd2018Eysenbach}, we adopt the classifiers to modify rewards in the source offline data in our experiments.

\subsubsection{More Examples with respect to the reward augmentation}

\begin{figure}[H]
	\centering
	\includegraphics[scale=0.369]{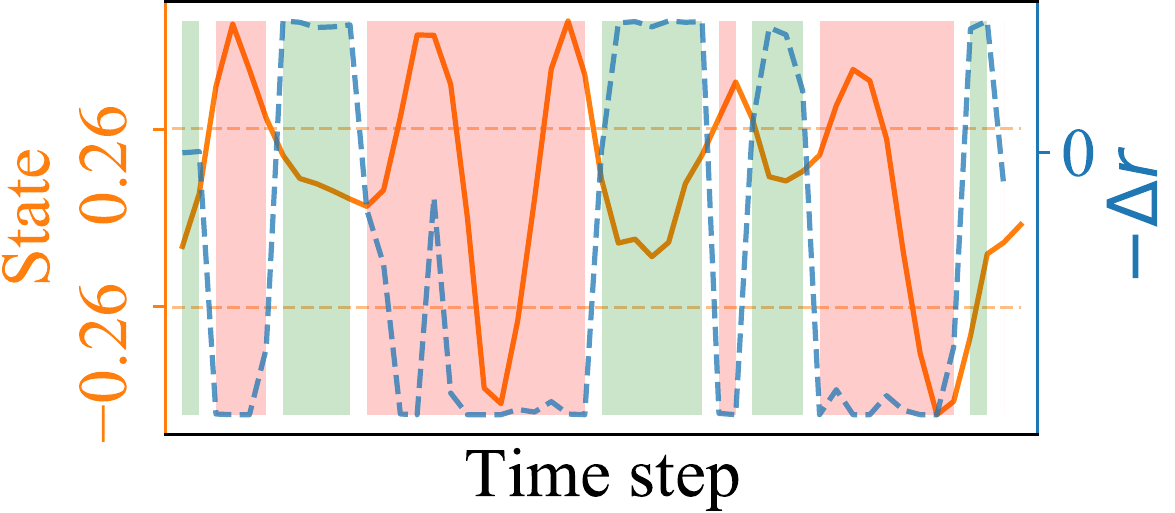} \ \ \   
	\includegraphics[scale=0.369]{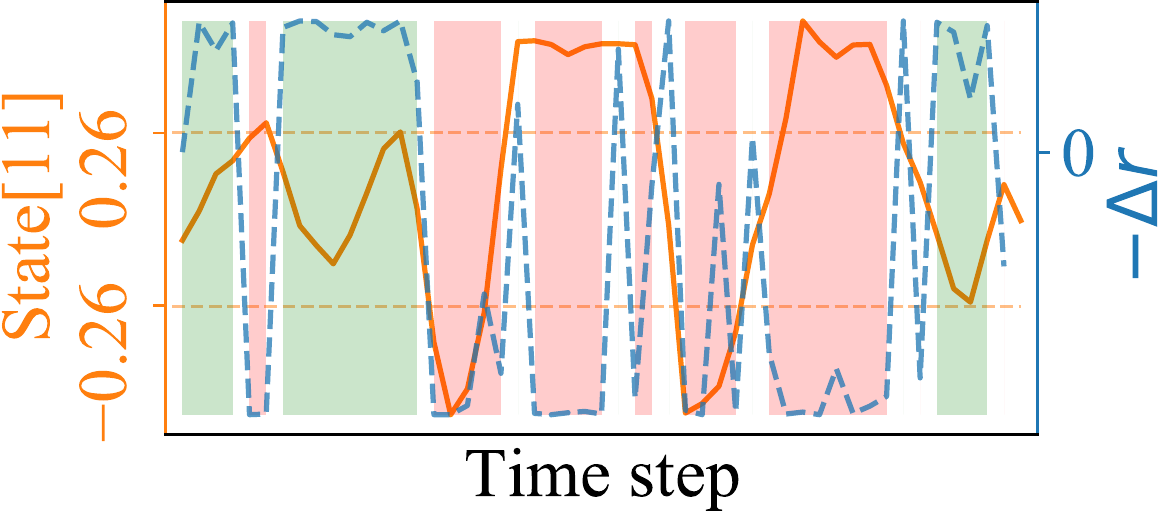} \  \  \ 
	\includegraphics[scale=0.369]{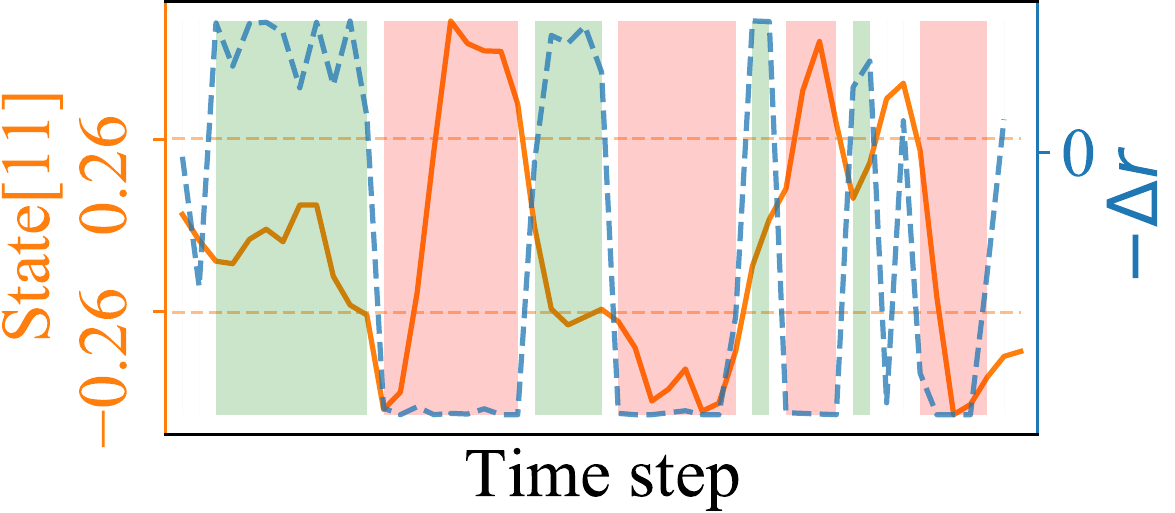}
	\centering
	\caption{We can observe that our reward augmentation 1) encourages ($\textcolor{blue}{-\Delta r} > 0$, \ie the green slice parts) these transitions ($-0.26\leq \text{next-state}[11] \leq 0.26$) that have the same dynamics with the target environment, and 2) discourages ($\textcolor{blue}{-\Delta r} < 0$, \ie the red slice parts) these transitions that have different (unreachable) dynamics ($ \text{next-state}[11] \leq -0.26$ or $ \text{next-state}[11] \geq 0.26$) in the target.}
	\label{appendix:exp1-ant-more-exmaples}
\end{figure}
In Figure~\ref{appendix:exp1-ant-more-exmaples},  we provide more examples with respect to the reward augmentation in the Ant task in Figure~\ref{fig-exp1-ant} (left).

\subsubsection{Comparison between \emph{10T}, \emph{1T}, \emph{1T+10S w/o Aug.}, and \emph{1T+10S DARA}} 
\label{appendix:comparsion-between-t-t-woaug-dara}

Based on various offline RL algorithms (BEAR~\citep{stabilizing2019Kumar},  BRAC-p~\citep{behavior2019Wu}, BCQ~\citep{off2019Fujimoto},  CQL~\citep{cql2020Kumar}, AWR~\citep{awr2019Peng}, MOPO~\citep{mopo2020Yu}, BC (behavior cloning), COMBO~\citep{combo2021Yu}), we provide the additional results in Tables~\ref{appendix:hopper41}, \ref{appendix:hopper46}, \ref{appendix:walker41}, \ref{appendix:walker46}, and \ref{appendix:halfcheetah46}.

\begin{table}[H]
	\centering
	\caption{Normalized scores for the Hopper tasks with the body mass (dynamics) shift. (The comparison results for BEAR, BRAC-p, AWR, CQL, and MOPO are provided in the main text.)}
	\begin{adjustbox}{max width=\textwidth}
		\begin{tabular}{clrrrrrrrrrrrr}
			\toprule 
			\multicolumn{2}{c}{\multirow{2}[1]{*}{\textcolor{black}{Body Mass Shift}}} & \multicolumn{1}{c}{\multirow{2}[1]{*}{10T}} & \multicolumn{1}{c}{\multirow{2}[1]{*}{1T}} & \multicolumn{1}{l}{1T+10S} & \multicolumn{1}{l}{1T+10S} & \multicolumn{1}{c}{\multirow{2}[1]{*}{10T}} & \multicolumn{1}{c}{\multirow{2}[1]{*}{1T}} & \multicolumn{1}{l}{1T+10S} & \multicolumn{1}{l}{1T+10S} & \multicolumn{1}{c}{\multirow{2}[1]{*}{10T}} & \multicolumn{1}{c}{\multirow{2}[1]{*}{1T}} & \multicolumn{1}{l}{1T+10S} & \multicolumn{1}{l}{1T+10S} \\
			\multicolumn{2}{c}{} &       &       & \multicolumn{1}{l}{w/o Aug.} & \multicolumn{1}{l}{DARA} &       &       & \multicolumn{1}{l}{w/o Aug.} & \multicolumn{1}{l}{DARA} &       &       & \multicolumn{1}{l}{w/o Aug.} & \multicolumn{1}{l}{DARA} \\
			\cmidrule{1-14}    \multirow{5}[3]{*}{\rotatebox{90}{\colorbox{magiccolor}{\quad Hopper \quad}}} &       & \multicolumn{4}{c}{BC}        & \multicolumn{4}{c}{COMBO}     & \multicolumn{4}{r}{} \\
			\cmidrule(r){3-6}  \cmidrule(r){7-10}           & Random & 9.8   & 9.8  \textcolor{red}{$\uparrow$}  & 6.9 \textcolor{blue}{$\downarrow$}   & 10.1 \textcolor{red}{$\uparrow$} & 17.9  & 0.7   \textcolor{blue}{$\downarrow$} & 5.4 \textcolor{red}{$\uparrow$}   & 4.6   \textcolor{blue}{$\downarrow$} &       &       &       &  \\
			& Medium & 29.0  & 27.9  \textcolor{blue}{$\downarrow$} & 17.6 \textcolor{blue}{$\downarrow$} & 25.0  \textcolor{red}{$\uparrow$} & 94.9  & 1.8   \textcolor{blue}{$\downarrow$} & 33.7 \textcolor{red}{$\uparrow$} & 45.7  \textcolor{red}{$\uparrow$} &       &       &       &  \\
			& Medium-R & 11.8  & 7.8  \textcolor{blue}{$\downarrow$}  & 7.7  \textcolor{blue}{$\downarrow$}  & {11.6} \textcolor{red}{$\uparrow$} & 73.1  & 13.1  \textcolor{blue}{$\downarrow$} & 11.0 \textcolor{blue}{$\downarrow$}  & 27.9 \textcolor{red}{$\uparrow$} &       &       &       &  \\
			& Medium-E & 111.9 & 21.5 \textcolor{blue}{$\downarrow$}  & 20.8  \textcolor{blue}{$\downarrow$} & 35.7 \textcolor{red}{$\uparrow$} & 111.1 & 0.8  \textcolor{blue}{$\downarrow$}  & 14.9 \textcolor{red}{$\uparrow$} & {108.1} \textcolor{red}{$\uparrow$} &       &       &       &  \\
			\bottomrule
		\end{tabular}%
		\label{appendix:hopper41}%
	\end{adjustbox}
\end{table}

\begin{table}[H]
	\centering
	\caption{Normalized scores for the Hopper tasks with the joint noise (dynamics) shift.}
	\begin{adjustbox}{max width=\textwidth}
		\begin{tabular}{clrrrrrrrrrrrr}
			\toprule 
			\multicolumn{2}{c}{\multirow{2}[1]{*}{\textcolor{black}{Joint Noise Shift}}} & \multicolumn{1}{c}{\multirow{2}[1]{*}{10T}} & \multicolumn{1}{c}{\multirow{2}[1]{*}{1T}} & \multicolumn{1}{l}{1T+10S} & \multicolumn{1}{l}{1T+10S} & \multicolumn{1}{c}{\multirow{2}[1]{*}{10T}} & \multicolumn{1}{c}{\multirow{2}[1]{*}{1T}} & \multicolumn{1}{l}{1T+10S} & \multicolumn{1}{l}{1T+10S} & \multicolumn{1}{c}{\multirow{2}[1]{*}{10T}} & \multicolumn{1}{c}{\multirow{2}[1]{*}{1T}} & \multicolumn{1}{l}{1T+10S} & \multicolumn{1}{l}{1T+10S} \\
			\multicolumn{2}{c}{} &       &       & \multicolumn{1}{l}{w/o Aug.} & \multicolumn{1}{l}{DARA} &       &       & \multicolumn{1}{l}{w/o Aug.} & \multicolumn{1}{l}{DARA} &       &       & \multicolumn{1}{l}{w/o Aug.} & \multicolumn{1}{l}{DARA} \\
			\cmidrule{1-14}    \multirow{5}[4]{*}{\rotatebox{90}{\colorbox{magiccolor}{\quad Hopper \quad}}} &       & \multicolumn{4}{c}{BEAR}      & \multicolumn{4}{c}{BRAC-p}    & \multicolumn{4}{c}{AWR} \\
			\cmidrule(r){3-6}  \cmidrule(r){7-10} \cmidrule(r){11-14}          & Random & 11.4  & 0.6 \textcolor{blue}{$\downarrow$} & 7.4 \textcolor{red}{$\uparrow$} & 4.2 \textcolor{blue}{$\downarrow$} & 11.0    & 10.8 \textcolor{blue}{$\downarrow$} & 10.0 \textcolor{blue}{$\downarrow$} & 10.8 \textcolor{red}{$\uparrow$} & 10.2  & 10.1 \textcolor{blue}{$\downarrow$} & 3.6 \textcolor{blue}{$\downarrow$} & 4.0 \textcolor{red}{$\uparrow$} \\
			& Medium & 52.1  & 0.8 \textcolor{blue}{$\downarrow$} & 2.0 \textcolor{red}{$\uparrow$} & 2.0 \textcolor{blue}{$\downarrow$} & 32.7  & 26.6 \textcolor{blue}{$\downarrow$} & 27.6 \textcolor{red}{$\uparrow$} & 37.6 \textcolor{red}{$\uparrow$} & 35.9  & 30.3 \textcolor{blue}{$\downarrow$} & 38.8 \textcolor{red}{$\uparrow$} & 41.3 \textcolor{red}{$\uparrow$} \\
			& Medium-R & 33.7  & 2.7 \textcolor{blue}{$\downarrow$} & 3.6 \textcolor{red}{$\uparrow$} & 9.9 \textcolor{red}{$\uparrow$} & 0.6   & 13.4 \textcolor{red}{$\uparrow$} & 89.9 \textcolor{red}{$\uparrow$} & 101.4 \textcolor{red}{$\uparrow$} & 28.4  & 12.4 \textcolor{blue}{$\downarrow$} & 6.7 \textcolor{blue}{$\downarrow$} & 7.2 \textcolor{red}{$\uparrow$} \\
			& Medium-E & 96.3  & 0.8 \textcolor{blue}{$\downarrow$} & 0.8 \textcolor{blue}{$\downarrow$} & 1.4 \textcolor{red}{$\uparrow$} & 1.9   & 19.8 \textcolor{red}{$\uparrow$} & 57.6 \textcolor{red}{$\uparrow$} & 87.8 \textcolor{red}{$\uparrow$} & 27.1  & 25.5 \textcolor{blue}{$\downarrow$} & 27.0 \textcolor{red}{$\uparrow$} & 27.0 \textcolor{blue}{$\downarrow$} \\
			\cmidrule{1-14}    \multirow{5}[4]{*}{\rotatebox{90}{\colorbox{magiccolor}{\quad Hopper \quad}}} &       & \multicolumn{4}{c}{BCQ}       & \multicolumn{4}{c}{CQL}       & \multicolumn{4}{c}{MOPO} \\
			\cmidrule(r){3-6}  \cmidrule(r){7-10} \cmidrule(r){11-14}          & Random & 10.6  & 10.5 \textcolor{blue}{$\downarrow$} & 7.0 \textcolor{blue}{$\downarrow$} & 9.6 \textcolor{red}{$\uparrow$} & 10.8  & 10.4 \textcolor{blue}{$\downarrow$} & 10.4 \textcolor{blue}{$\downarrow$} & 10.8 \textcolor{red}{$\uparrow$} & 11.7  & 1.5 \textcolor{blue}{$\downarrow$} & 1.3 \textcolor{blue}{$\downarrow$} & 2.9 \textcolor{red}{$\uparrow$} \\
			& Medium & 54.5  & 45.8 \textcolor{blue}{$\downarrow$} & 49.0 \textcolor{red}{$\uparrow$} & 54.4 \textcolor{red}{$\uparrow$} & 58.0    & 46.2 \textcolor{blue}{$\downarrow$} & 58.0 \textcolor{red}{$\uparrow$} & 58.0 \textcolor{blue}{$\downarrow$} & 28.0    & 2.7 \textcolor{blue}{$\downarrow$} & 9.2 \textcolor{red}{$\uparrow$} & 17.3 \textcolor{red}{$\uparrow$} \\
			& Medium-R & 33.1  & 13.0 \textcolor{blue}{$\downarrow$} & 23.8 \textcolor{red}{$\uparrow$} & 32.0 \textcolor{red}{$\uparrow$} & 48.6  & 13.6 \textcolor{blue}{$\downarrow$} & 2.6 \textcolor{blue}{$\downarrow$} & 3.6 \textcolor{red}{$\uparrow$} & 67.5  & 0.8 \textcolor{blue}{$\downarrow$} & 2.3 \textcolor{red}{$\uparrow$} & 6.4 \textcolor{red}{$\uparrow$} \\
			& Medium-E & 110.9 & 44.6 \textcolor{blue}{$\downarrow$} & 96 \textcolor{red}{$\uparrow$} & 109 \textcolor{red}{$\uparrow$} & 98.7  & 50.7 \textcolor{blue}{$\downarrow$} & 73.4 \textcolor{red}{$\uparrow$} & 108.9 \textcolor{red}{$\uparrow$} & 23.7  & 1 \textcolor{blue}{$\downarrow$} & 6.1 \textcolor{red}{$\uparrow$} & 7.5 \textcolor{red}{$\uparrow$} \\
			\cmidrule{1-14}    \multirow{5}[3]{*}{\rotatebox{90}{\colorbox{magiccolor}{\quad Hopper \quad}}} &       & \multicolumn{4}{c}{BC}        & \multicolumn{4}{c}{COMBO}     & \multicolumn{4}{r}{} \\
			\cmidrule(r){3-6}  \cmidrule(r){7-10}           & Random & 9.8   & 9.8  \textcolor{red}{$\uparrow$}  & 7.5 \textcolor{blue}{$\downarrow$}  & 9.1  \textcolor{red}{$\uparrow$}  & 17.9  & 0.7 \textcolor{blue}{$\downarrow$}  & 1.8  \textcolor{red}{$\uparrow$}  & 4.9  \textcolor{red}{$\uparrow$}  &       &       &       &  \\
			& Medium & 29.0  & 27.9 \textcolor{blue}{$\downarrow$} & 29.0  \textcolor{red}{$\uparrow$} & 29.0  \textcolor{red}{$\uparrow$} & 94.9  & 1.8  \textcolor{blue}{$\downarrow$} & 0.7 \textcolor{blue}{$\downarrow$}  & 9.6   \textcolor{red}{$\uparrow$} &       &       &       &  \\
			& Medium-R & 11.8  & 7.8 \textcolor{blue}{$\downarrow$}  & 8.5  \textcolor{red}{$\uparrow$}  & 11.3  \textcolor{red}{$\uparrow$} & 73.1  & 13.1 \textcolor{blue}{$\downarrow$} & 4.0 \textcolor{blue}{$\downarrow$}    & 9.6  \textcolor{red}{$\uparrow$}  &       &       &       &  \\
			& Medium-E & 111.9 & 21.5 \textcolor{blue}{$\downarrow$} & 53.5  \textcolor{red}{$\uparrow$} & 77.9 \textcolor{red}{$\uparrow$}  & 111.1 & 0.8 \textcolor{blue}{$\downarrow$}  & 34.0   \textcolor{red}{$\uparrow$}  & 45.9  \textcolor{red}{$\uparrow$} &       &       &       &  \\
			\bottomrule
		\end{tabular}%
		\label{appendix:hopper46}%
	\end{adjustbox}
\end{table}

\begin{table}[H]
	\centering
	\caption{Normalized scores for the Walker2d tasks with the body mass (dynamics) shift. (The comparison  results for  BEAR, BRAC-p, AWR, CQL, and MOPO are provided in the main text.)}
	\begin{adjustbox}{max width=\textwidth}
		\begin{tabular}{clrrrrrrrrrrrr}
			\toprule
			\multicolumn{2}{c}{\multirow{2}[1]{*}{\textcolor{black}{Body Mass Shift}}} & \multicolumn{1}{c}{\multirow{2}[1]{*}{10T}} & \multicolumn{1}{c}{\multirow{2}[1]{*}{1T}} & \multicolumn{1}{l}{1T+10S} & \multicolumn{1}{l}{1T+10S} & \multicolumn{1}{c}{\multirow{2}[1]{*}{10T}} & \multicolumn{1}{c}{\multirow{2}[1]{*}{1T}} & \multicolumn{1}{l}{1T+10S} & \multicolumn{1}{l}{1T+10S} & \multicolumn{1}{c}{\multirow{2}[1]{*}{10T}} & \multicolumn{1}{c}{\multirow{2}[1]{*}{1T}} & \multicolumn{1}{l}{1T+10S} & \multicolumn{1}{l}{1T+10S} \\
			\multicolumn{2}{c}{} &       &       & \multicolumn{1}{l}{w/o Aug.} & \multicolumn{1}{l}{DARA} &       &       & \multicolumn{1}{l}{w/o Aug.} & \multicolumn{1}{l}{DARA} &       &       & \multicolumn{1}{l}{w/o Aug.} & \multicolumn{1}{l}{DARA} \\
			\cmidrule{1-14}    \multirow{5}[3]{*}{\rotatebox{90}{\colorbox{magiccolor}{Walker2d \quad}}} &       & \multicolumn{4}{c}{BC}        & \multicolumn{4}{c}{COMBO}     & \multicolumn{4}{r}{} \\
			\cmidrule(r){3-6}  \cmidrule(r){7-10}            & Random & 1.6   & 0.1 \textcolor{blue}{$\downarrow$}  & 1.7  \textcolor{red}{$\uparrow$}  & {2.7}  \textcolor{red}{$\uparrow$}  & 7.0   & 1.8 \textcolor{blue}{$\downarrow$}  & 2.0  \textcolor{red}{$\uparrow$}  & 3.5  \textcolor{red}{$\uparrow$}  &       &       &       &  \\
			& Medium & 6.6   & 5.5  \textcolor{blue}{$\downarrow$}  & 3.8  \textcolor{blue}{$\downarrow$} & {6.6}  \textcolor{red}{$\uparrow$}  & 75.5  & -1.0 \textcolor{blue}{$\downarrow$} & 23.9  \textcolor{red}{$\uparrow$} & 36.6  \textcolor{red}{$\uparrow$} &       &       &       &  \\
			& Medium-R & 11.3  & 6.6 \textcolor{blue}{$\downarrow$}  & 8.1  \textcolor{red}{$\uparrow$}  & 11.0  \textcolor{red}{$\uparrow$} & 56.0  & 0.1 \textcolor{blue}{$\downarrow$}  & 11.4  \textcolor{red}{$\uparrow$} & 22.6  \textcolor{red}{$\uparrow$} &       &       &       &  \\
			& Medium-E & 6.4   & 3.1  \textcolor{blue}{$\downarrow$} & 6.0  \textcolor{red}{$\uparrow$}  & 6.2  \textcolor{red}{$\uparrow$}  & 96.1  & -0.9 \textcolor{blue}{$\downarrow$} & -0.1  \textcolor{red}{$\uparrow$} & -0.1  \textcolor{red}{$\uparrow$} &       &       &       &  \\
			\bottomrule
		\end{tabular}%
		\label{appendix:walker41}%
	\end{adjustbox}
\end{table}

\begin{table}[H]
	\centering
	\caption{Normalized scores for the Walker2d tasks with the joint noise (dynamics) shift.}
	\begin{adjustbox}{max width=\textwidth}
		\begin{tabular}{clrrrrrrrrrrrr}
			\toprule
			\multicolumn{2}{c}{\multirow{2}[1]{*}{\textcolor{black}{Joint Noise Shift}}} & \multicolumn{1}{c}{\multirow{2}[1]{*}{10T}} & \multicolumn{1}{c}{\multirow{2}[1]{*}{1T}} & \multicolumn{1}{l}{1T+10S} & \multicolumn{1}{l}{1T+10S} & \multicolumn{1}{c}{\multirow{2}[1]{*}{10T}} & \multicolumn{1}{c}{\multirow{2}[1]{*}{1T}} & \multicolumn{1}{l}{1T+10S} & \multicolumn{1}{l}{1T+10S} & \multicolumn{1}{c}{\multirow{2}[1]{*}{10T}} & \multicolumn{1}{c}{\multirow{2}[1]{*}{1T}} & \multicolumn{1}{l}{1T+10S} & \multicolumn{1}{l}{1T+10S} \\
			\multicolumn{2}{c}{} &       &       & \multicolumn{1}{l}{w/o Aug.} & \multicolumn{1}{l}{DARA} &       &       & \multicolumn{1}{l}{w/o Aug.} & \multicolumn{1}{l}{DARA} &       &       & \multicolumn{1}{l}{w/o Aug.} & \multicolumn{1}{l}{DARA} \\
			\cmidrule{1-14}    \multirow{5}[4]{*}{\rotatebox{90}{\colorbox{magiccolor}{Walker2d \quad}}} &       & \multicolumn{4}{c}{BEAR}      & \multicolumn{4}{c}{BRAC-p}    & \multicolumn{4}{c}{AWR} \\
			\cmidrule(r){3-6}  \cmidrule(r){7-10} \cmidrule(r){11-14}          & Random & 7.3   & 2.2 \textcolor{blue}{$\downarrow$} & 0.6 \textcolor{blue}{$\downarrow$} & 2.6 \textcolor{red}{$\uparrow$} & -0.2  & 2.8 \textcolor{red}{$\uparrow$} & 3.3 \textcolor{red}{$\uparrow$} & 8.8 \textcolor{red}{$\uparrow$} & 1.5   & 0.9 \textcolor{blue}{$\downarrow$} & 1.5 \textcolor{red}{$\uparrow$} & 1.5 \textcolor{blue}{$\downarrow$} \\
			& Medium & 59.1  & -0.4 \textcolor{blue}{$\downarrow$} & 0.6 \textcolor{red}{$\uparrow$} & 0.1 \textcolor{blue}{$\downarrow$} & 77.5  & 28.8 \textcolor{blue}{$\downarrow$} & 55.2 \textcolor{red}{$\uparrow$} & 72.9 \textcolor{red}{$\uparrow$} & 17.4  & 12.2 \textcolor{blue}{$\downarrow$} & 17.2 \textcolor{red}{$\uparrow$} & 17.2 \textcolor{blue}{$\downarrow$} \\
			& Medium-R & 19.2  & 0.4 \textcolor{blue}{$\downarrow$} & 4 \textcolor{red}{$\uparrow$} & 10.4 \textcolor{red}{$\uparrow$} & -0.3  & 6.3 \textcolor{red}{$\uparrow$} & 32.1 \textcolor{red}{$\uparrow$} & 34.8 \textcolor{red}{$\uparrow$} & 15.5  & 6 \textcolor{blue}{$\downarrow$} & 1.4 \textcolor{blue}{$\downarrow$} & 2.1 \textcolor{red}{$\uparrow$} \\
			& Medium-E & 40.1  & -0.2 \textcolor{blue}{$\downarrow$} & 0.8 \textcolor{red}{$\uparrow$} & 0.6 \textcolor{blue}{$\downarrow$} & 76.9  & 21.8 \textcolor{blue}{$\downarrow$} & 62.3 \textcolor{red}{$\uparrow$} & 74.3 \textcolor{red}{$\uparrow$} & 53.8  & 40.4 \textcolor{blue}{$\downarrow$} & 53 \textcolor{red}{$\uparrow$} & 53.6 \textcolor{red}{$\uparrow$} \\
			\cmidrule(r){1-14}     \multirow{5}[4]{*}{\rotatebox{90}{\colorbox{magiccolor}{Walker2d \quad}}} &       & \multicolumn{4}{c}{BCQ}       & \multicolumn{4}{c}{CQL}       & \multicolumn{4}{c}{MOPO} \\
			\cmidrule(r){3-6}  \cmidrule(r){7-10} \cmidrule(r){11-14}          & Random & 4.9   & 3.7 \textcolor{blue}{$\downarrow$} & 3.4 \textcolor{blue}{$\downarrow$} & 5.2 \textcolor{red}{$\uparrow$} & 7     & 0.5 \textcolor{blue}{$\downarrow$} & 2.7 \textcolor{red}{$\uparrow$} & 6.4 \textcolor{red}{$\uparrow$} & 13.6  & -0.3 \textcolor{blue}{$\downarrow$} & -0.2 \textcolor{red}{$\uparrow$} & -0.2 \textcolor{blue}{$\downarrow$} \\
			& Medium & 53.1  & 43 \textcolor{blue}{$\downarrow$} & 44.9 \textcolor{red}{$\uparrow$} & 52.7 \textcolor{red}{$\uparrow$} & 79.2  & 43.9 \textcolor{blue}{$\downarrow$} & 73.2 \textcolor{red}{$\uparrow$} & 81.2 \textcolor{red}{$\uparrow$} & 17.8  & 5.8 \textcolor{blue}{$\downarrow$} & 7.8 \textcolor{red}{$\uparrow$} & 12.2 \textcolor{red}{$\uparrow$} \\
			& Medium-R & 15    & 5.7 \textcolor{blue}{$\downarrow$} & 9.8 \textcolor{red}{$\uparrow$} & 14.6 \textcolor{red}{$\uparrow$} & 26.7  & 1.8 \textcolor{blue}{$\downarrow$} & 1.4 \textcolor{blue}{$\downarrow$} & 1.8 \textcolor{red}{$\uparrow$} & 39    & 0.8 \textcolor{blue}{$\downarrow$} & 9.3 \textcolor{red}{$\uparrow$} & 16.4 \textcolor{red}{$\uparrow$} \\
			& Medium-E & 57.5  & 44.5 \textcolor{blue}{$\downarrow$} & 40.6 \textcolor{blue}{$\downarrow$} & 57.2 \textcolor{red}{$\uparrow$} & 111   & 46.8 \textcolor{blue}{$\downarrow$} & 109.9 \textcolor{red}{$\uparrow$} & 116.5 \textcolor{red}{$\uparrow$} & 44.6  & 2.9 \textcolor{blue}{$\downarrow$} & 15.2 \textcolor{red}{$\uparrow$} & 26.3 \textcolor{red}{$\uparrow$} \\
			\cmidrule{1-14}    \multirow{5}[3]{*}{\rotatebox{90}{\colorbox{magiccolor}{Walker2d \quad}}} &       & \multicolumn{4}{c}{BC}        & \multicolumn{4}{c}{COMBO}     & \multicolumn{4}{r}{} \\
			\cmidrule(r){3-6}  \cmidrule(r){7-10}           & Random & 1.6   & 0.1  \textcolor{blue}{$\downarrow$}  & 0.9  \textcolor{red}{$\uparrow$}  & 1.6  \textcolor{red}{$\uparrow$}  & 7.0   & 1.8  \textcolor{blue}{$\downarrow$}  &  0.1 \textcolor{blue}{$\downarrow$}     &   1.5  \textcolor{red}{$\uparrow$}    &       &       &       &  \\
			& Medium & 6.6   & 5.5  \textcolor{blue}{$\downarrow$}  & 6.4  \textcolor{red}{$\uparrow$}  & 6.5  \textcolor{red}{$\uparrow$}  & 75.5  & -1.0  \textcolor{blue}{$\downarrow$} &   0.4 \textcolor{red}{$\uparrow$}    & 0.7 \textcolor{red}{$\uparrow$}      &       &       &       &  \\
			& Medium-R & 11.3  & 6.6  \textcolor{blue}{$\downarrow$}  & 4.6  \textcolor{blue}{$\downarrow$}  & 10.4  \textcolor{red}{$\uparrow$} & 56.0  & 0.1  \textcolor{blue}{$\downarrow$}  &  5.6 \textcolor{red}{$\uparrow$}     & 7.4 \textcolor{red}{$\uparrow$}      &       &       &       &  \\
			& Medium-E & 6.4   & 3.1  \textcolor{blue}{$\downarrow$}  & 6.2  \textcolor{red}{$\uparrow$}  & 6.4  \textcolor{red}{$\uparrow$}  & 96.1  & -0.9 \textcolor{blue}{$\downarrow$}  & 0.8 \textcolor{red}{$\uparrow$} & -0.1 \textcolor{blue}{$\downarrow$} &       &       &       &  \\
			\bottomrule
		\end{tabular}%
		\label{appendix:walker46}
	\end{adjustbox}
\end{table}

\begin{table}[H]
	\centering
	\caption{Normalized scores for the Halfcheetah tasks with the joint noise (dynamics) shift.}
	\begin{adjustbox}{max width=\textwidth}
		\begin{tabular}{clrrrrrrrrrrrr}
			\toprule
			\multicolumn{2}{c}{\multirow{2}[1]{*}{\textcolor{black}{Joint Noise Shift}}} & \multicolumn{1}{c}{\multirow{2}[1]{*}{10T}} & \multicolumn{1}{c}{\multirow{2}[1]{*}{1T}} & \multicolumn{1}{l}{1T+10S} & \multicolumn{1}{l}{1T+10S} & \multicolumn{1}{c}{\multirow{2}[1]{*}{10T}} & \multicolumn{1}{c}{\multirow{2}[1]{*}{1T}} & \multicolumn{1}{l}{1T+10S} & \multicolumn{1}{l}{1T+10S} & \multicolumn{1}{c}{\multirow{2}[1]{*}{10T}} & \multicolumn{1}{c}{\multirow{2}[1]{*}{1T}} & \multicolumn{1}{l}{1T+10S} & \multicolumn{1}{l}{1T+10S} \\
			\multicolumn{2}{c}{} &       &       & \multicolumn{1}{l}{w/o Aug.} & \multicolumn{1}{l}{DARA} &       &       & \multicolumn{1}{l}{w/o Aug.} & \multicolumn{1}{l}{DARA} &       &       & \multicolumn{1}{l}{w/o Aug.} & \multicolumn{1}{l}{DARA} \\
			\cmidrule{1-14}    \multirow{5}[4]{*}{\rotatebox{90}{\colorbox{magiccolor}{Halfcheetah \ }}} &       & \multicolumn{4}{c}{BEAR}      & \multicolumn{4}{c}{BRAC-p}    & \multicolumn{4}{c}{AWR} \\
			\cmidrule(r){3-6}  \cmidrule(r){7-10} \cmidrule(r){11-14}          & Random & 25.1  & 17.8  \textcolor{blue}{$\downarrow$} & 25.0  \textcolor{red}{$\uparrow$}  & 25.1 \textcolor{red}{$\uparrow$} & 24.1  & 10.0   \textcolor{blue}{$\downarrow$}  & 25.0  \textcolor{red}{$\uparrow$}  & 26.7 \textcolor{red}{$\uparrow$} & 2.5   & 2.7 \textcolor{red}{$\uparrow$}  & 3.1 \textcolor{red}{$\uparrow$}  & 48.9 \textcolor{red}{$\uparrow$} \\
			& Medium & 41.7  & -0.2  \textcolor{blue}{$\downarrow$} & 0.8  \textcolor{red}{$\uparrow$} & 1.5 \textcolor{red}{$\uparrow$}  & 43.8  & 43.0   \textcolor{blue}{$\downarrow$}  & 52.4 \textcolor{red}{$\uparrow$} & 53.0  \textcolor{red}{$\uparrow$}  & 37.4  & 38.2 \textcolor{red}{$\uparrow$} & 48.7 \textcolor{red}{$\uparrow$} & 37.4  \textcolor{blue}{$\downarrow$} \\
			& Medium-R & 38.6  & 9.3  \textcolor{blue}{$\downarrow$}  & -0.6  \textcolor{blue}{$\downarrow$} & -0.5 \textcolor{red}{$\uparrow$} & 45.4  & 2.5  \textcolor{blue}{$\downarrow$}  & -2.3  \textcolor{blue}{$\downarrow$} & 45.3 \textcolor{red}{$\uparrow$} & 40.3  & 2.6   \textcolor{blue}{$\downarrow$} & 2.3   \textcolor{blue}{$\downarrow$} & 2.3 \textcolor{blue}{$\downarrow$}\ \\
			& Medium-E & 53.4  & -1.2  \textcolor{blue}{$\downarrow$} & 1.0  \textcolor{red}{$\uparrow$}   & -1.4  \textcolor{blue}{$\downarrow$} & 44.2  & 6.9  \textcolor{blue}{$\downarrow$} & 0.9  \textcolor{blue}{$\downarrow$}  & 45.3 \textcolor{red}{$\uparrow$} & 52.7  & 32.2  \textcolor{blue}{$\downarrow$} & 80.6 \textcolor{red}{$\uparrow$} & 79.2  \textcolor{blue}{$\downarrow$}\\
			\cmidrule(r){1-14}     \multirow{5}[4]{*}{\rotatebox{90}{\colorbox{magiccolor}{Halfcheetah \ }}} &       & \multicolumn{4}{c}{BCQ}       & \multicolumn{4}{c}{CQL}       & \multicolumn{4}{c}{MOPO} \\
			\cmidrule(r){3-6}  \cmidrule(r){7-10} \cmidrule(r){11-14}          & Random & 2.2   & 2.3 \textcolor{red}{$\uparrow$}  & 2.2   \textcolor{blue}{$\downarrow$} & 2.3  \textcolor{red}{$\uparrow$} & 35.4  & -2.3  \textcolor{blue}{$\downarrow$} & -2.4  \textcolor{blue}{$\downarrow$} & 10.4 \textcolor{red}{$\uparrow$} & 35.4  & 2.3  \textcolor{blue}{$\downarrow$}  & 1.2  \textcolor{blue}{$\downarrow$}  & 1.1  \textcolor{blue}{$\downarrow$} \\
			& Medium & 40.7  & 37.6  \textcolor{blue}{$\downarrow$} & 40.0  \textcolor{red}{$\uparrow$}  & 48.6 \textcolor{red}{$\uparrow$} & 44.4  & 35.4  \textcolor{blue}{$\downarrow$} & 40.7 \textcolor{red}{$\uparrow$} & 52.6 \textcolor{red}{$\uparrow$} & 42.3  & 3.2  \textcolor{blue}{$\downarrow$}  & 3.5  \textcolor{blue}{$\downarrow$}  & 5.3 \textcolor{red}{$\uparrow$}\\
			& Medium-R & 38.2  & 1.1  \textcolor{blue}{$\downarrow$}  & 39.4 \textcolor{red}{$\uparrow$} & 41.3 \textcolor{red}{$\uparrow$} & 46.2  & 0.6  \textcolor{blue}{$\downarrow$}  & 2.0  \textcolor{red}{$\uparrow$}   & 1.9   \textcolor{blue}{$\downarrow$} & 53.1  & -0.1  \textcolor{blue}{$\downarrow$} & 2.6 \textcolor{red}{$\uparrow$}  & 4.2 \textcolor{red}{$\uparrow$} \\
			& Medium-E & 64.7  & 37.3  \textcolor{blue}{$\downarrow$} & 55.3 \textcolor{red}{$\uparrow$} & 76.9 \textcolor{red}{$\uparrow$} & 62.4  & -3.3  \textcolor{blue}{$\downarrow$} & 7.7 \textcolor{red}{$\uparrow$}  & 1.7  \textcolor{blue}{$\downarrow$}  & 63.3  & 4.2  \textcolor{blue}{$\downarrow$} & 1.5   \textcolor{blue}{$\downarrow$} & 7.2 \textcolor{red}{$\uparrow$}\\
			\cmidrule{1-14}    \multirow{5}[3]{*}{\rotatebox{90}{\colorbox{magiccolor}{Halfcheetah \ }}} &       & \multicolumn{4}{c}{BC}        & \multicolumn{4}{c}{COMBO}     & \multicolumn{4}{r}{} \\
			\cmidrule(r){3-6}  \cmidrule(r){7-10}           &  Random & 2.1   & 2.0  \textcolor{blue}{$\downarrow$}    & 2.2 \textcolor{red}{$\uparrow$}  & 2.2 \textcolor{red}{$\uparrow$}\    & 38.8  & 24.0   \textcolor{blue}{$\downarrow$}  & 18.7  \textcolor{blue}{$\downarrow$} & 20.3 \textcolor{red}{$\uparrow$} &       &       &       &  \\
			& Medium & 36.1  & 36.5 \textcolor{red}{$\uparrow$} & 49.4 \textcolor{red}{$\uparrow$} & 49.8 \textcolor{red}{$\uparrow$} & 54.2  & 15.7  \textcolor{blue}{$\downarrow$} & 14.9  \textcolor{blue}{$\downarrow$} & 15.9 \textcolor{red}{$\uparrow$} &       &       &       &  \\
			& Medium-R & 38.4  & 36.5  \textcolor{blue}{$\downarrow$} & 24.6  \textcolor{blue}{$\downarrow$} & 15.7  \textcolor{blue}{$\downarrow$} & 55.1  & -2.6  \textcolor{blue}{$\downarrow$} & -2.4 \textcolor{red}{$\uparrow$} & 4.8 \textcolor{red}{$\uparrow$}  &       &       &       &  \\
			& Medium-E & 35.8  & 36.3 \textcolor{red}{$\uparrow$} & 49.0  \textcolor{red}{$\uparrow$}  & 49.3 \textcolor{red}{$\uparrow$} & 90.0    & 4.4  \textcolor{blue}{$\downarrow$}  & 6.5  \textcolor{red}{$\uparrow$} & 11.1 \textcolor{red}{$\uparrow$} &       &       &       &  \\
			\bottomrule
		\end{tabular}%
		\label{appendix:halfcheetah46}
	\end{adjustbox}
\end{table}

\subsubsection{Comparison with the cross-domain based baselines} 
\label{appendix:comparsion-with-cross-domain}

In Tables~\ref{appendix:hopper46finetune} and~\ref{appendix:walker46finetune}, we provide the comparison between our DARA-based methods, fine-tune based methods, and MABE-based methods in Hopper and Walker2d tasks, over the dynamics shift concerning the joint noise of motion. 
We can observe that in a majority of tasks, our DARA-based methods outperforms the fine-tune-based method (67  "\textcolor{red}{$\uparrow$}" vs. 13 "\textcolor{blue}{$\downarrow$}", including the results in the main text). Moreover, our DARA can achieve comparable or better performance compared to MABE-based baselines on eleven out of sixteen tasks (including the results in the main text).

\begin{table}[H]
	\centering
	\caption{Normalized scores in the (target) D4RL Hopper tasks with the joint noise shift., where "Tune" denotes baseline "fine-tune".}
	\begin{adjustbox}{max width=\textwidth}
		\begin{tabular}{clrrrrrrrrrrrr}
			\toprule 
			\multicolumn{2}{l}{\textcolor{black}{Joint Noise Shift}} & Tune  & DARA  & Tune  & DARA  & Tune  & DARA  & Tune  & DARA  & Tune  & DARA  & $\pi_p\hat{T}$  & $\hat{T}\pi_p$ \\
			\midrule
			\multirow{5}[1]{*}{\rotatebox{90}{\colorbox{magiccolor}{\quad Hopper \quad}}} &       & \multicolumn{2}{c}{BEAR} & \multicolumn{2}{c}{BRAC-p} & \multicolumn{2}{c}{BCQ} & \multicolumn{2}{c}{CQL} & \multicolumn{2}{c}{MOPO} & \multicolumn{2}{c}{MABE} \\
			\cmidrule(r){3-4}  \cmidrule(r){5-6} \cmidrule(r){7-8} \cmidrule(r){9-10} \cmidrule(r){11-12}  \cmidrule(r){13-14} 
			& Random & 0.8   & 4.2  \textcolor{red}{$\uparrow$} & 6.4   & 10.8 \textcolor{red}{$\uparrow$} & 8.1   & 9.6 \textcolor{red}{$\uparrow$}  & \textbf{32.2}  & 10.8  \textcolor{blue}{$\downarrow$} & 0.6   & 2.9  \textcolor{red}{$\uparrow$} & 10.8  & 8.1 \\
			& Medium & 1.9   & 2.0 \textcolor{red}{$\uparrow$}  & 44.9  & 37.6  \textcolor{blue}{$\downarrow$} & 47.7  & 54.4 \textcolor{red}{$\uparrow$} & 52.5  & 58.0 \textcolor{red}{$\uparrow$} & 0.8   & 17.3 \textcolor{red}{$\uparrow$} & \textbf{63.5}  & 57.7 \\
			& Medium-R & 0.7   & 9.9 \textcolor{red}{$\uparrow$}  & 32.4  & \textbf{101.4} \textcolor{red}{$\uparrow$} & 29.6  & 32.0 \textcolor{red}{$\uparrow$} & 1.3   & 3.6 \textcolor{red}{$\uparrow$}  & 1.8   & 6.4  \textcolor{red}{$\uparrow$} & 21.5  & 35.4 \\
			& Medium-E & 0.8   & 1.4 \textcolor{red}{$\uparrow$}   & 98.2  & 87.8 \textcolor{blue}{$\downarrow$} & 90.5  & \textbf{109.0} \textcolor{red}{$\uparrow$} & 107.3 & 108.9 \textcolor{red}{$\uparrow$} & 4.9   & 7.5 \textcolor{red}{$\uparrow$}  & 15.5  & 104.8 \\
			\bottomrule
		\end{tabular}%
	\end{adjustbox}
	\label{appendix:hopper46finetune}%
\end{table}

\begin{table}[H]
	\centering
	\caption{Normalized scores in the (target) D4RL Walker2d tasks with the joint noise shift., where "Tune" denotes baseline "fine-tune".}
	\begin{adjustbox}{max width=\textwidth}
		\begin{tabular}{clrrrrrrrrrrrr}
			\toprule 
			\multicolumn{2}{l}{\textcolor{black}{Joint Noise Shift}} & Tune  & DARA  & Tune  & DARA  & Tune  & DARA  & Tune  & DARA  & Tune  & DARA  & $\pi_p\hat{T}$  & $\hat{T}\pi_p$ \\
			\midrule
			\multirow{5}[1]{*}{\rotatebox{90}{\colorbox{magiccolor}{ Walker2d \quad}}} &       & \multicolumn{2}{c}{BEAR} & \multicolumn{2}{c}{BRAC-p} & \multicolumn{2}{c}{BCQ} & \multicolumn{2}{c}{CQL} & \multicolumn{2}{c}{MOPO} & \multicolumn{2}{c}{MABE} \\
			\cmidrule(r){3-4}  \cmidrule(r){5-6} \cmidrule(r){7-8} \cmidrule(r){9-10} \cmidrule(r){11-12}  \cmidrule(r){13-14} 
			& Random & 2.7   & 2.6 \textcolor{blue}{$\downarrow$}  & 1.4   & \textbf{8.8}  \textcolor{red}{$\uparrow$}  & 3.0     & 5.2 \textcolor{red}{$\uparrow$}  & {6.7}   & 6.4  \textcolor{blue}{$\downarrow$}  & -0.4  & -0.2  \textcolor{red}{$\uparrow$} & 5.0     & -0.2 \\
			& Medium & 0.5   & 0.1  \textcolor{blue}{$\downarrow$} & 55.8  & 72.9 \textcolor{red}{$\uparrow$} & 45.1  & 52.7 \textcolor{red}{$\uparrow$} & 76.6  & \textbf{81.2} \textcolor{red}{$\uparrow$} & 7.0     & 12.2 \textcolor{red}{$\uparrow$} & 49.4  & 48.7 \\
			& Medium-R & 3.2   & 10.4 \textcolor{red}{$\uparrow$} & 12.2  & \textbf{34.8} \textcolor{red}{$\uparrow$} & 13.5  & 14.6 \textcolor{red}{$\uparrow$} & -0.4  & 1.8 \textcolor{red}{$\uparrow$}  & 1.9   & 16.4 \textcolor{red}{$\uparrow$} & 4.5   & 1.6 \\
			& Medium-E & -0.4  & 0.6 \textcolor{red}{$\uparrow$}  & 71.7  & 74.3 \textcolor{red}{$\uparrow$} & 44.8  & 57.2 \textcolor{red}{$\uparrow$} & 104   & \textbf{116.5} \textcolor{red}{$\uparrow$} & 11.3  & 26.3  \textcolor{red}{$\uparrow$} & 84.7  & 82.6 \\
			\bottomrule
		\end{tabular}%
	\end{adjustbox}
	\label{appendix:walker46finetune}%
\end{table}

\subsubsection{Additional results on the quadruped robot}
\label{appendix:sec:additional-results-on-dog}

In this offline sim2real setting, we collect the source offline data in the simulator ($10^6$ or $2*10^6$ steps) and target offline data in the real world ($3*10^4$ steps). See Appendix~\ref{appendix:environments-and-dataset} for details. 
For testing, we directly deploy the learned policy in the real (flat or obstructive) environment and adopt the average distance covered in an episode (300 steps) as our evaluation metrics. 

\begin{figure}[H]
	\centering
	\includegraphics[scale=0.05]{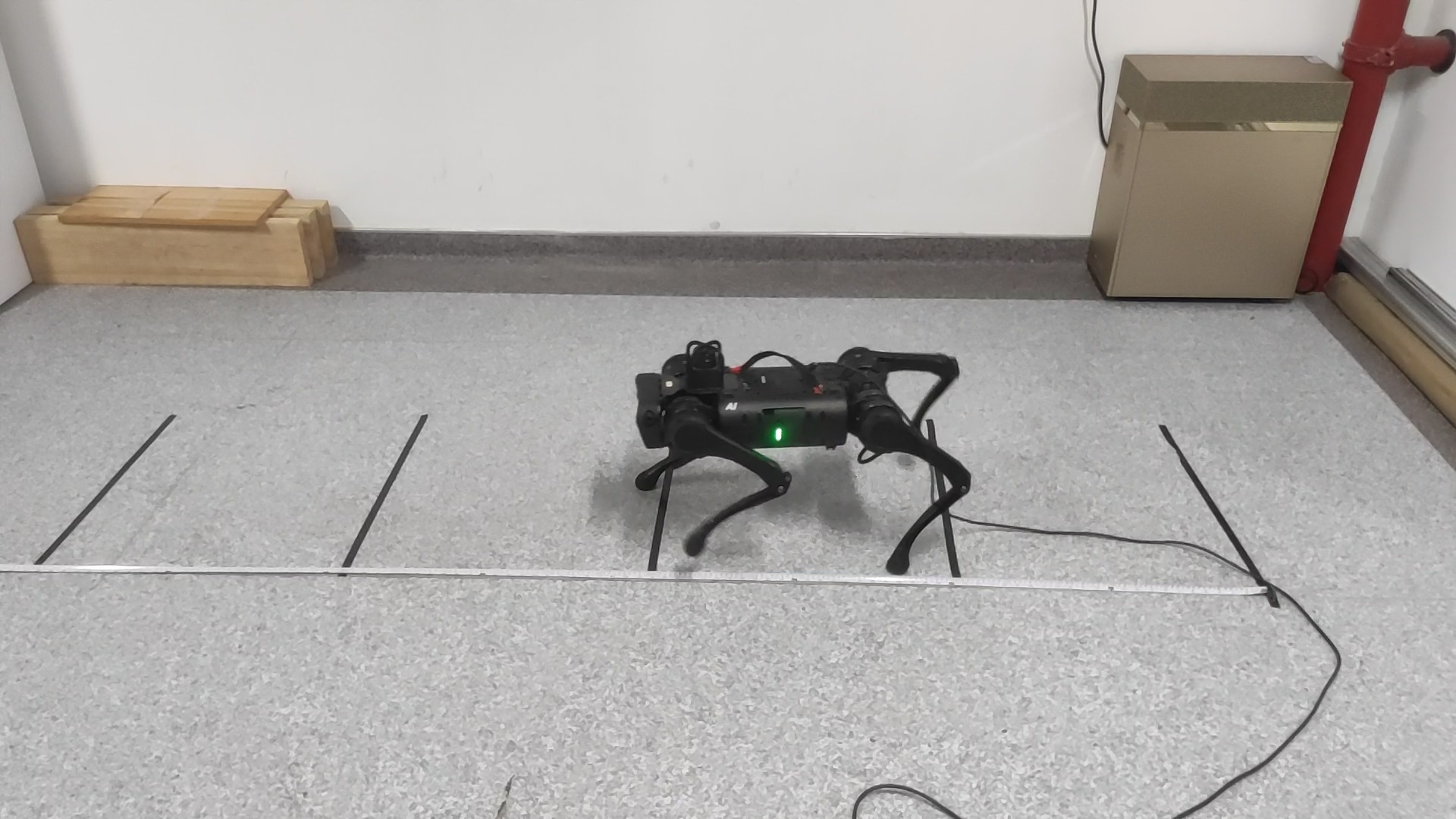} \ \ \ \ \ \ \ \quad \quad 
	\includegraphics[scale=0.05]{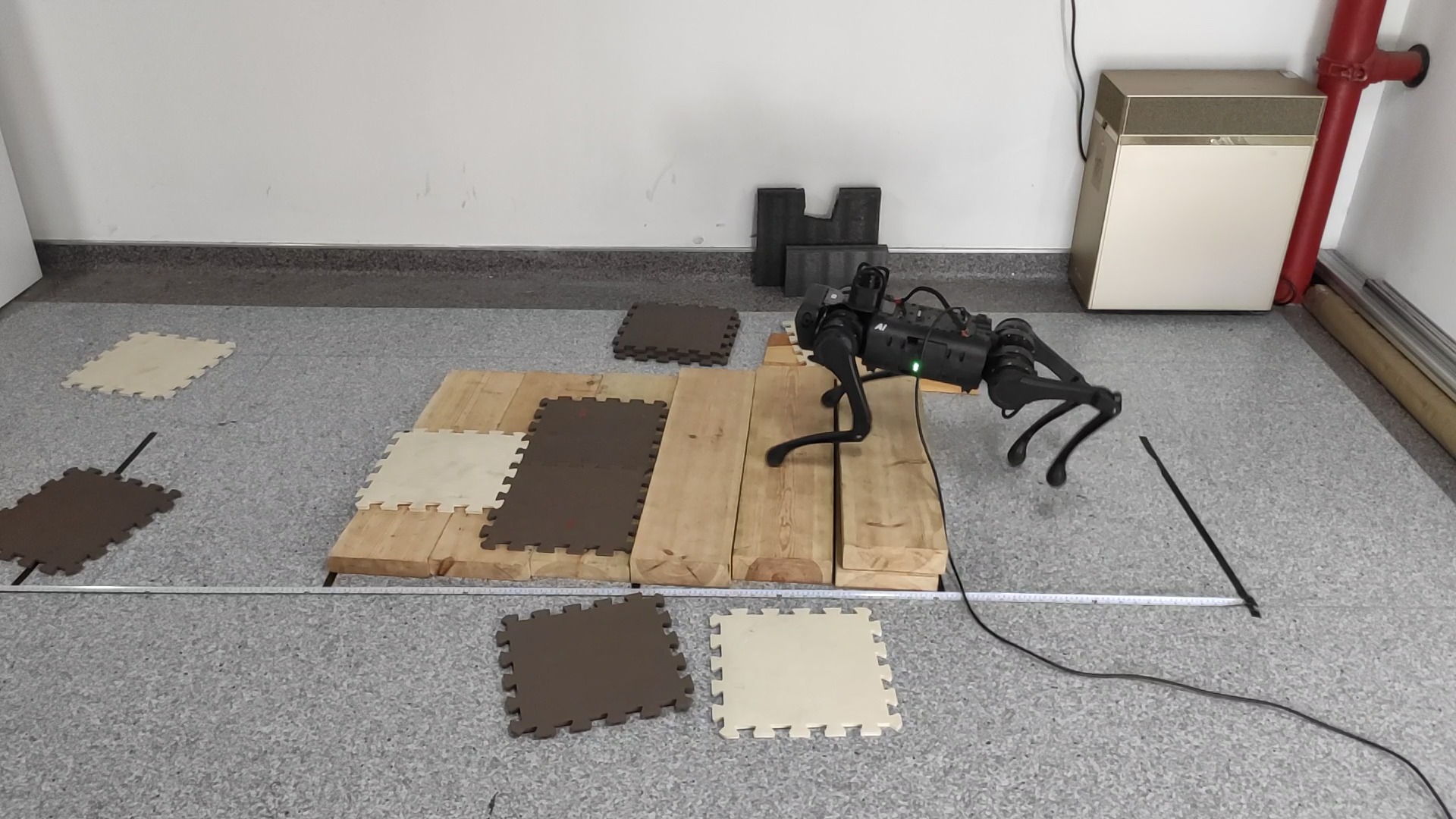}
	\centering
	\caption{Illustration of the real environment (for testing): \emph{(left)} the flat and static environment, \emph{(right)} the obstructive and dynamic environment.} 
	\label{appendix:environmnts-illustration-flat-or-obstructive}
\end{figure}

\begin{table}[H]
	\centering
	\caption{Average distance (m) covered in an episode (300 steps) in flat and static (real) environment.}
	\begin{adjustbox}{max width=1.0\textwidth}
		\begin{tabular}{clcccccc}
			\toprule
			\multicolumn{2}{l}{Sim2real (Flat and Static)} & \multicolumn{1}{l}{w/o Aug.} & \multicolumn{1}{l}{DARA} & \multicolumn{1}{l}{w/o Aug.} & \multicolumn{1}{l}{DARA} & \multicolumn{1}{l}{w/o Aug.} & \multicolumn{1}{l}{DARA} \\
			\midrule
			\multirow{5}[1]{*}{\rotatebox{0}{\colorbox{magiccolor}{Quadruped Robot}}} &       & \multicolumn{2}{c}{BCQ} & \multicolumn{2}{c}{CQL} & \multicolumn{2}{c}{MOPO} \\
			\cmidrule(r){3-4}  \cmidrule(r){5-6} \cmidrule(r){7-8}
			& Medium & 1.56  & {1.64} \textcolor{red}{$\uparrow$}  & 1.80 \quad & \textbf{1.82} \textcolor{red}{$\uparrow$} \quad   & 0.00   & 0.00   \\
			& Medium-R & 0.00     & 0.00 \ \  \    & {---} \quad   & {---}  \ \  \    & {---}  \quad    & {---} \quad \\
			& Medium-E & 2.16  & \textbf{2.47} \textcolor{red}{$\uparrow$}  & 2.03  & 2.02 \textcolor{blue}{$\downarrow$}  & 0.00      & 0.00  \\
			& Medium-R-E & 1.69  & \textbf{2.28} \textcolor{red}{$\uparrow$}  & 0.00     & 0.00 \ \  \     & {---}   \quad   & {---}  \quad  \\
			\midrule
			\multicolumn{2}{l}{Average performance improvement} & & 13.6\% & & 0.2\% & & 0.0\%  \\
			\bottomrule 
		\end{tabular}%
	\end{adjustbox}
	\label{appendix:tab:dog-results-on-real-dog-flat}%
\end{table}%

\textbf{(Flat and static environment)} 
We first deploy our learned policy in the flat and static environment. 
The results (distance covered in an episode) are provided in Table~\ref{appendix:tab:dog-results-on-real-dog-flat}. 

\begin{figure}[H]
	\centering
	\includegraphics[width=0.95\textwidth]{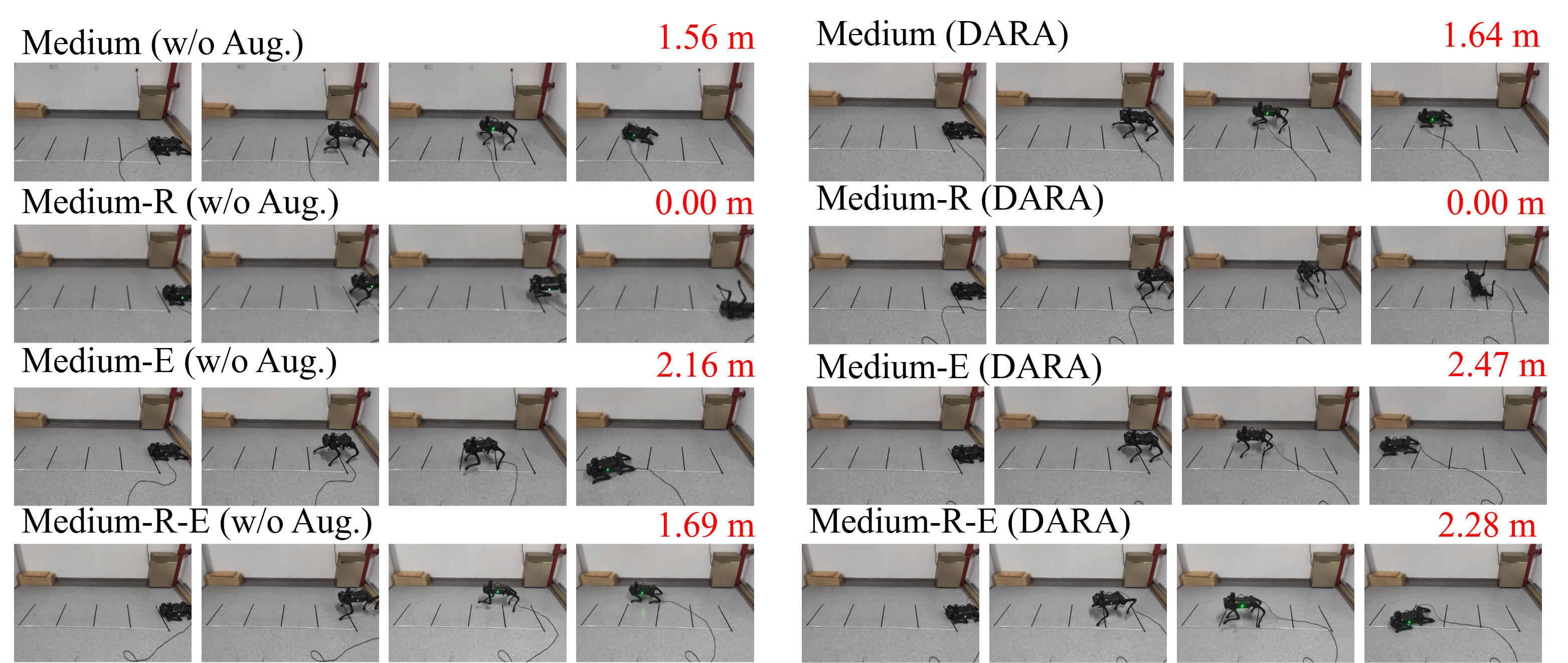}
	\centering
	\caption{Deployment on the flat and static environment of BCQ.}
	\label{appendix:fig:dog-loco-flat-bcq}
\end{figure}

1) BCQ (Figure~\ref{appendix:fig:dog-loco-flat-bcq}): We find that with Medium-R offline data,  \emph{w/o~Aug. BCQ}  and \emph{DARA BCQ}  both could not acquire the locomotion skills, which we think is caused by the lack of high-quality offline data. 
With more "expert" data (Medium-R $\to$ Medium $\to$ Medium-E, or Medium-R $\to$ Medium-R-E), \emph{w/o-Aug. BCQ} allows for progressive performance ($0.00 \to 1.56 \to 2.16$, or $0.00 \to 1.69$ in BCQ), but with our reward augmentation, such performance can be further improved (with average improvement $13.6\%$).  

\begin{figure}[H]
	\centering
	\includegraphics[width=0.95\textwidth]{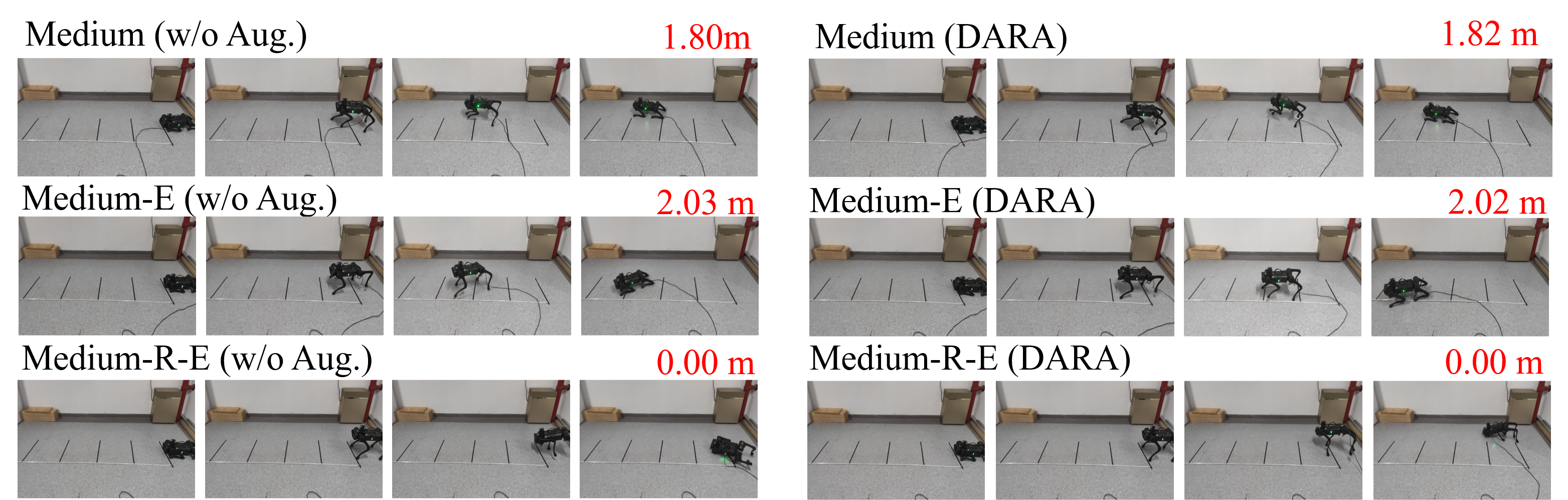}
	\centering
	\caption{Deployment on the flat and static environment of CQL.}
	\label{appendix:fig:dog-loco-flat-cql}
\end{figure}

2) CQL (Figure~\ref{appendix:fig:dog-loco-flat-cql}): We find that with Medium-R or Medium-R-E offline data, \emph{w/o~Aug. CQL}  and \emph{DARA CQL}  both could not learn the locomotion skills, which we think is caused by the low-quality "Replay" offline data. With Medium or Medium-E offline data, \emph{w/o~Aug. CQL} and \emph{DARA CQL} acquire similar performance on this flat and static environment. 

3) MOPO: We find that the model-based MOPO (both \emph{w/o Aug.} and \emph{DARA}) could hardly learn the locomotion skill under the provided offline data.

\begin{table}[H]
	\centering
	\caption{Average distance (m) covered in an episode (300 steps) in the obstructive and dynamic (real) environment.}
	\begin{adjustbox}{max width=1.0\textwidth}
		\begin{tabular}{clcccccc}
			\toprule
			\multicolumn{2}{l}{Sim2real (Obstructive and Dynamic)} & \multicolumn{1}{l}{w/o Aug.} & \multicolumn{1}{l}{DARA} & \multicolumn{1}{l}{w/o Aug.} & \multicolumn{1}{l}{DARA} & \multicolumn{1}{l}{w/o Aug.} & \multicolumn{1}{l}{DARA} \\
			\midrule
			\multirow{5}[1]{*}{\rotatebox{0}{\colorbox{magiccolor}{Quadruped Robot}}} &       & \multicolumn{2}{c}{BCQ} & \multicolumn{2}{c}{CQL} & \multicolumn{2}{c}{MOPO} \\
			\cmidrule(r){3-4}  \cmidrule(r){5-6} \cmidrule(r){7-8}
			& Medium & 0.85  & {1.35} \textcolor{red}{$\uparrow$}  & 0.92 \quad & \textbf{1.40} \textcolor{red}{$\uparrow$} \quad   & {---}  \quad  & {---}  \quad\\
			& Medium-R & {---}  \quad    & {---}  \quad  & {---}  \quad  & {---}  \quad   & {---}  \quad    & {---} \quad \\
			& Medium-E & 1.15  & \textbf{1.41} \textcolor{red}{$\uparrow$}  & 0.77  & 1.32 \textcolor{red}{$\uparrow$}  &{---}  \quad    & {---}  \quad \\
			& Medium-R-E & 1.27  & \textbf{1.55} \textcolor{red}{$\uparrow$}  & {---}  \quad  & {---}  \quad  & {---}   \quad   & {---}  \quad  \\
			\midrule
			\multicolumn{2}{l}{Average performance improvement} & & 25.9\% & & 30.9\% & & --- \quad \\
			\bottomrule 
		\end{tabular}%
	\end{adjustbox}
	\label{appendix:tab:dog-results-on-real-dog-wood}%
\end{table}%

\textbf{(Obstructive and dynamic environment)} 
We then deploy our learned policy in the obstructive and dynamic environment. 
The results (distance covered in an episode) are provided in Table~\ref{appendix:tab:dog-results-on-real-dog-wood}.

\begin{figure}[H]
	\centering
	\includegraphics[width=0.95\textwidth]{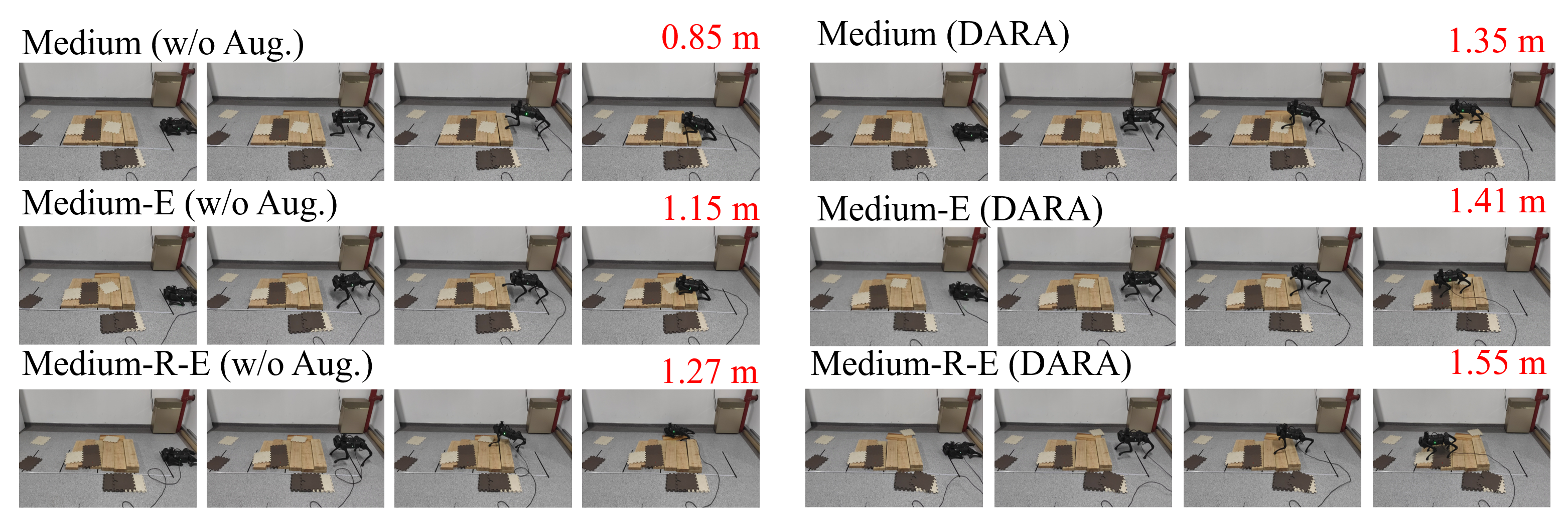}
	\centering
	\caption{Deployment on the obstructive and dynamic environment of BCQ.}
	\label{appendix:fig:dog-loco-wood-bcq}
\end{figure}

1) BCQ (Figure~\ref{appendix:fig:dog-loco-wood-bcq}): In this obstructive environment, we can obtain similar results as in the flat environment.  With more "expert" data (Medium $\to$ Medium-E  $\to$ Medium-R-E), \emph{w/o Aug. BCQ} allows for progressive performance ($0.85 \to 1.15 \to  1.27$), and with our reward augmentation, such performance can be further improved (with average improvement $25.9\%$). At the same time, we can also find that due to the presence of environmental obstacles, the performance of both \emph{w/o Aug. BCQ} and \emph{w/o Aug. DARA} is decreased compared to the deployment on the flat environment. However, we find that our DARA exhibits greater average performance improvement under this obstructive environment ($13.6\% \to \textbf{25.9\%}$) compared to that in the flat environment.  
\emph{These results demonstrate that our DARA can learn an adaptive policy for the target environment and thus show a greater advantage over \emph{w/o-Aug.} in more complex environments.} 

\begin{figure}[H]
	\centering
	\includegraphics[width=0.95\textwidth]{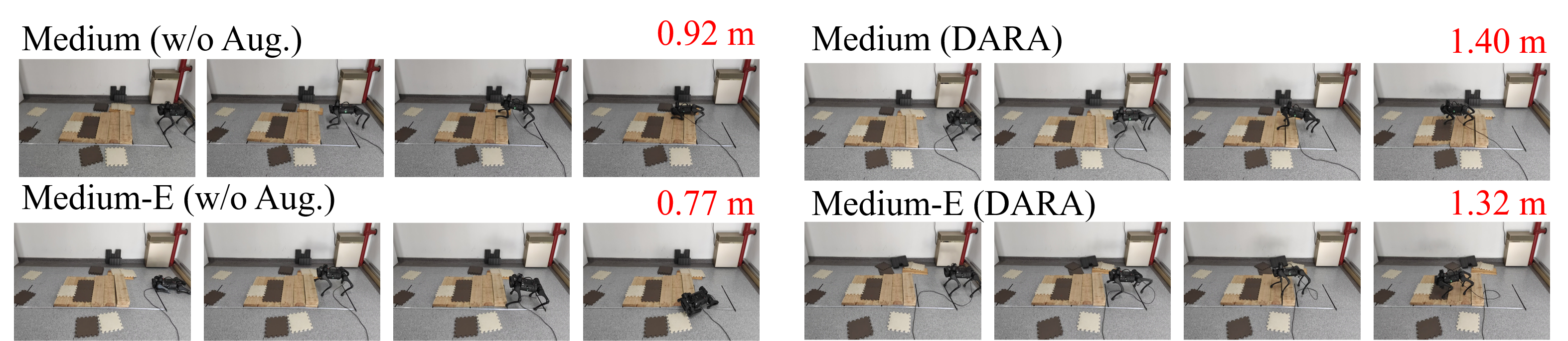}
	\centering
	\caption{Deployment on the obstructive and dynamic environment of CQL.}
	\label{appendix:fig:dog-loco-wood-cql}
\end{figure}

2) CQL (Figure~\ref{appendix:fig:dog-loco-wood-cql}): Similar to BCQ, our \emph{DARA CQL} exhibits a greater performance improvement over baseline in the obstructive and dynamic environment ($0.2\% \to \textbf{30.9\%}$) compared to that in the flat and static environment. 

\textbf{In summary}, The results in the quadruped robot tasks support our conclusion in
the main text regarding the dynamics shift problem in offline RL --- with only modest amounts of target offline data ($3*10^4$ steps), DARA-based methods can acquire an adaptive policy for the   (both flat and obstructive) target environment and exhibit better performance compared to baselines under the dynamics adaptation setting.

\subsubsection{Ablation study with respect to the amount of target offline data}
\label{app:sec:ablation-study-on-target-data}

To see whether the amount of target offline data can be further reduced, we  show the results of the ablation study with respect to the amount of target offline data in Tables~\ref{app:tab:ablation-amount-hopper41} and \ref{app:tab:ablation-amount-walker41}.


\begin{table}[H]
	\centering
	\caption{Ablation study with respect to the amount of target Hopper data (body mass shift tasks). 10\%, 5\% and 1\% denote training with 10\%, 5\% and 1\% of target offline data, respectively,  and additional 100\% source offline data.}
	\begin{adjustbox}{max width=\textwidth}
	\begin{tabular}{clrrrrrrrrrrrr}
		\toprule
		\multicolumn{2}{c}{Body Mass Shift} & 10\%  & 5\%   & 1\%   & 10\%  & 5\%   & 1\%   & 10\%  & 5\%   & 1\%   & 10\%  & 5\%   & 1\% \\
		\midrule
		\multirow{3}[1]{*}{\rotatebox{0}{\colorbox{magiccolor}{\quad Hopper \quad}}} &       & \multicolumn{3}{c}{BEAR} & \multicolumn{3}{c}{BRAC-p} & \multicolumn{3}{c}{BCQ} & \multicolumn{3}{c}{CQL} \\
		\cmidrule(r){3-5}  \cmidrule(r){6-8}  \cmidrule(r){9-11}  \cmidrule(r){12-14}  
		& Medium-R & 34.1  & 10.7  & 6.4   & 30.8  & 27.7  & 20.0  & 32.8  & 20.5  & 16.3  & 3.7   & 3.2   & 2.3 \\
		& Medium-E & 1.2   & 0.6   & 0.6   & 34.7  & 25.1  & 20.6  & 84.2  & 65.1  & 55.6  & 99.7  & 52.3  & 38.5 \\
		\bottomrule
	\end{tabular}%
\end{adjustbox}
\label{app:tab:ablation-amount-hopper41}%
\end{table}%

\begin{table}[H]
	\centering
	\caption{Ablation study with respect to the amount of target Walker2d data (body mass shift tasks). 10\%, 5\% and 1\% denote training with 10\%, 5\% and 1\% of target offline data, respectively,  and additional 100\% source offline data.}
		\begin{adjustbox}{max width=\textwidth}
	\begin{tabular}{clrrrrrrrrrrrr}
		\toprule 
		\multicolumn{2}{c}{Body Mass Shift} & 10\%  & 5\%   & 1\%   & 10\%  & 5\%   & 1\%   & 10\%  & 5\%   & 1\%   & 10\%  & 5\%   & 1\% \\
		\midrule
		\multirow{3}[1]{*}{\rotatebox{0}{\colorbox{magiccolor}{   Walker2d  }}} &       & \multicolumn{3}{c}{BEAR} & \multicolumn{3}{c}{BRAC-p} & \multicolumn{3}{c}{BCQ} & \multicolumn{3}{c}{CQL} \\
		\cmidrule(r){3-5}  \cmidrule(r){6-8}  \cmidrule(r){9-11}  \cmidrule(r){12-14}  
		& Medium-R & 7.3   & 5.9   & 1.3   & 18.6  & 21.6  & 15.8  & 15.1  & 12.7  & 9.7   & 2.0   & 1.3   & 0.5 \\
		& Medium-E & 2.3   & -0.2  & -0.3  & 77.5  & 2.0   & -0.2  & 57.2  & 29.7  & 20.6  & 93.3  & 0.1   & -0.3 \\
		\bottomrule
	\end{tabular}%
		\end{adjustbox}
	\label{app:tab:ablation-amount-walker41}%
\end{table}%

\subsubsection{Illustration of whether the learned policy is limited to the source offline data}

\begin{figure}[htbp]
	\vspace{-3pt}
	\centering
	\includegraphics[scale=0.35]{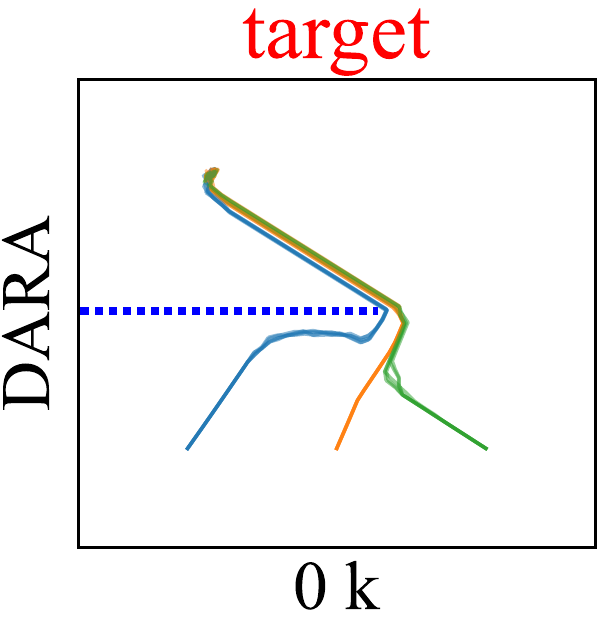} \ \ \ 
	\includegraphics[scale=0.35]{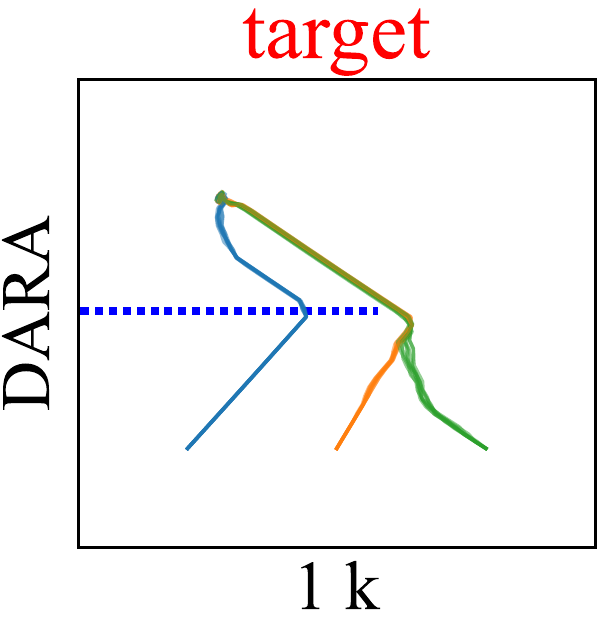} \ \ \ 
	\includegraphics[scale=0.35]{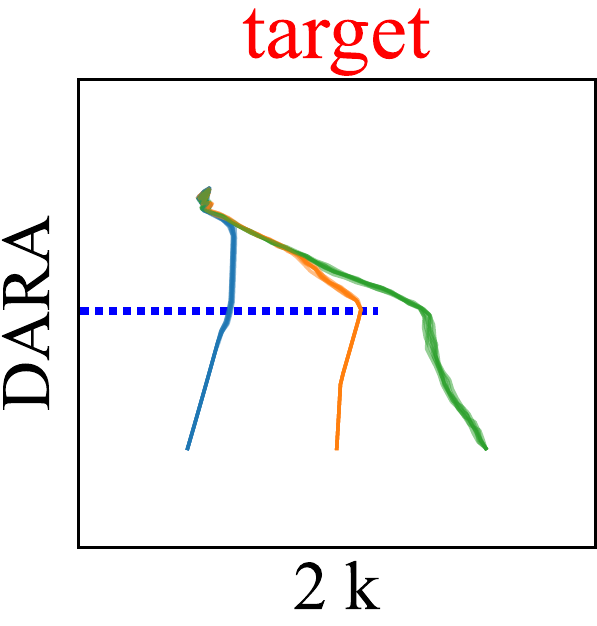} \ \ \ 
	\includegraphics[scale=0.35]{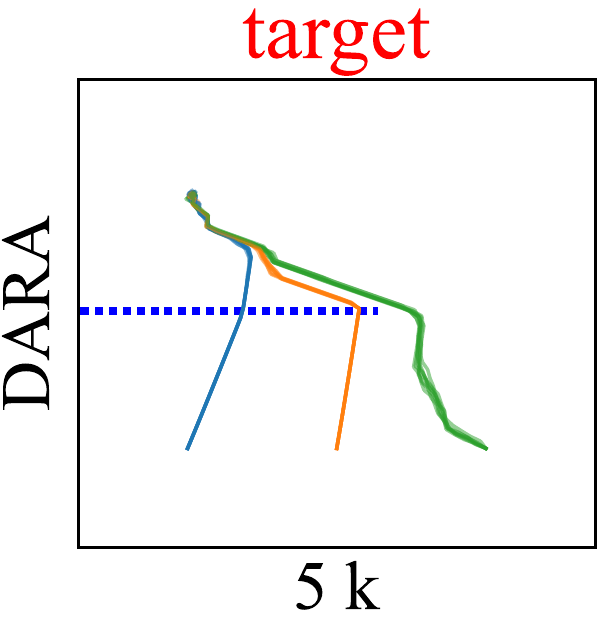} \ \ \ 
	\includegraphics[scale=0.35]{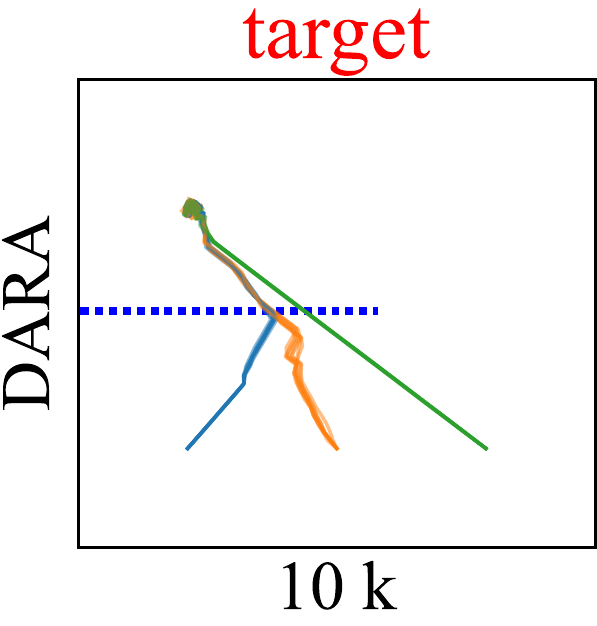} 
	\centering
	\vspace{-3pt}
	\caption{We exchange the source environment and the target environment in Figure 2 (in the main text) so that the source environment has an obstacle and the target environment has no obstacles. In the source domain, we collect 100k of random transitions. In the target domain, we collect 0k, 1k, 2k, 5k, and 10k random transitions respectively. We set $\eta = 0.1$. We can find that if we perform DARA with only source offline data $\DS$ (\ie 0k target data), we indeed can not acquire the optimal trajectory (eg. the short path without the obstacle). However, even there is no transition of passing through obstacles in the source data, performing DARA with $\{ \DS \cup \DT \}$ enables us to acquire the behavior of moving through obstacles. As we increase the number of target offline data $\DT$, training with $\{\DS \cup \DT  \}$ can gradually acquire optimal trajectories.  }
	\label{appendix:inverse:dynamics}
\end{figure}

If we directly perform DARA with only the source offline data $\DS$, the learned behaviors will be restricted to the source offline data. For example, in the Map task, collecting source dataset with the obstacle and collecting target dataset without the obstacle. In this case, it can be harder for DARA (with only the source $\DS$) to capture the change in the transition dynamics, thus harder for the agent to figure out the new optimal policy (the shorter path without the obstacle). However, as stated in Algorithm~\ref{alg-dara}, we perform offline RL algorithms with both target offline data and source offline data $\{\DS \cup \DT \}$. Thus, to some extent, such limitation can be overcome as long as offline RL algorithm captures the information (eg. the short path without the obstacle) contained in the (limited) target $\DT$, see Figure~\ref{appendix:inverse:dynamics} for the illustration.

\subsubsection{Comparison between DARA and importance sampling (IS) based dynamics correction}

In Table~\ref{tab:is_vs_dara}, we report the experimental comparison between DARA and importance sampling based dynamics adaption. We can find that in most of the tasks, our DARA performs better than the IS-based approaches.

\begin{table}[htbp]
	\centering
	\caption{Comparison between DARA and importance sampling (IS) based dynamics correction.}
	\begin{tabular}{clllllll}
		\toprule
		\multicolumn{2}{c}{Body Mass Shift} & \multicolumn{1}{c}{IS}    & \multicolumn{1}{c}{DARA}  & \multicolumn{1}{c}{IS}    & \multicolumn{1}{c}{DARA}  & \multicolumn{1}{c}{IS}    & \multicolumn{1}{c}{DARA} \\
		\midrule
		\multirow{5}[2]{*}{\rotatebox{90}{\colorbox{magiccolor}{\quad Hopper \quad}}} &       & \multicolumn{2}{c}{BEAR} & \multicolumn{2}{c}{BRAC-p} & \multicolumn{2}{c}{AWR} \\
		\cmidrule(r){3-4} \cmidrule(r){5-6} \cmidrule(r){7-8}
		& Random & 4.6 $\pm$ 2.8 & \textbf{8.4} $\pm$ 1.2 & 10.8 $\pm$ 0.5 & \textbf{11} $\pm$ 0.6 & \textbf{10.2} $\pm$ 0.3 & 4.5 $\pm$ 0.9 \\
		& Medium & 1 $\pm$ 0.4 & \textbf{1.6} $\pm$ 1 & 17.4 $\pm$ 10.6 & \textbf{32.9} $\pm$ 7.5 & 24.8 $\pm$ 7.7 & \textbf{28.9} $\pm$ 5.5 \\
		& Medium-R & 17.3 $\pm$ 4.7 & \textbf{34.1} $\pm$ 5.8 & 21.6 $\pm$ 8.3 & \textbf{30.8} $\pm$ 4.9 & \textbf{14} $\pm$ 2.2 & 4.2 $\pm$ 3.5 \\
		& Medium-E & 0.8 $\pm$ 0.2 & \textbf{1.2} $\pm$ 0.5 & \textbf{36} $\pm$ 13.5 & 34.7 $\pm$ 8.5 & \textbf{29.3} $\pm$ 2.6 & 26.6 $\pm$ 2 \\
		\midrule
		\multirow{5}[2]{*}{\rotatebox{90}{\colorbox{magiccolor}{\quad Hopper \quad}}} &       & \multicolumn{2}{c}{BCQ} & \multicolumn{2}{c}{CQL} & \multicolumn{2}{c}{MOPO} \\
		\cmidrule(r){3-4} \cmidrule(r){5-6} \cmidrule(r){7-8}
		& Random & 9.2 $\pm$ 1.1 & \textbf{9.7} $\pm$ 0.2 & 10.3 $\pm$ 0.4 & \textbf{10.4} $\pm$ 0.4 & \textbf{2.8} $\pm$ 3 & 2.1 $\pm$ 1.7 \\
		& Medium & 28.2 $\pm$ 8.8 & \textbf{38.4} $\pm$ 1.8 & 43.3 $\pm$ 10 & \textbf{59.3} $\pm$ 12.2 & 7.6 $\pm$ 7.2 & \textbf{10.7} $\pm$ 5.1 \\
		& Medium-R & 14.2 $\pm$ 1.3 & \textbf{32.8} $\pm$ 0.9 & 2.2 $\pm$ 0.3 & \textbf{3.7} $\pm$ 1.4 & 4.9 $\pm$ 3.8 & \textbf{8.4} $\pm$ 3.5 \\
		& Medium-E & 83.4 $\pm$ 23.7 & \textbf{84.2} $\pm$ 9.8 & 87.8 $\pm$ 16.9 & \textbf{99.7} $\pm$ 16.4 & 4.6 $\pm$ 2.9 & \textbf{5.8} $\pm$ 2.3 \\
		\bottomrule
	\end{tabular}%
	\label{tab:is_vs_dara}%
\end{table}%

\subsubsection{The sensitivity of the coefficient of the reward modification}

In Table~\ref{tab:hyper_eta}, We check the sensitivity of hyper-parameter $\eta$, \ie the coefficient of the reward modification in $r (\bs, \ba) - \eta \DR (\bs, \ba, \bs')$.

\begin{table}[htbp]
	\centering
	\caption{We show the normalized scores for the Hopper tasks with body mass shift, by varying $\eta \in \{ 0, 0.05, 0.1, 0.2, 0.5 \}$ over BEAR, BRAC-p, AWR, BCQ, CQL, and MOPO.}
	\begin{tabular}{cllllll}
		\toprule
		\multicolumn{2}{c}{\multirow{2}[1]{*}{Body Mass Shift}} & \multicolumn{5}{c}{Hyper-parameter $\eta $} \\
		\cmidrule(r){3-7}
		\multicolumn{2}{c}{} & \multicolumn{1}{c}{0} & \multicolumn{1}{c}{0.05} & \multicolumn{1}{c}{0.1} & \multicolumn{1}{c}{0.2} & \multicolumn{1}{c}{0.5} \\
		\midrule
		\multirow{5}[2]{*}{\rotatebox{0}{\colorbox{magiccolor}{\quad Hopper \quad}}} &       & \multicolumn{5}{c}{BEAR} \\
		\cmidrule(r){3-7}
		& Random & 4.6 $\pm$ 3.4 & 7.7 $\pm$ 0.9 & \textbf{8.4} $\pm$ 1.2 & 7 $\pm$ 1.2 & 4.2 $\pm$ 1.1 \\
		& Medium & 0.9 $\pm$ 0.3 & 1.1 $\pm$ 0.6 & \textbf{1.6} $\pm$ 1 & 0.9 $\pm$ 0.2 & 0.7 $\pm$ 0.1 \\
		& Medium-R & 18.2 $\pm$ 5 & 28.5 $\pm$ 5.9 & \textbf{34.1} $\pm$ 5.8 & 29.1 $\pm$ 4.4 & 18.1 $\pm$ 4.3 \\
		& Medium-E & 0.6 $\pm$ 0 & 0.8 $\pm$ 0.1 & \textbf{1.2} $\pm$ 0.5 & \textbf{1.2} $\pm$ 0.6 & 0.7 $\pm$ 0.1 \\
		\midrule
		\multirow{5}[2]{*}{\rotatebox{0}{\colorbox{magiccolor}{\quad Hopper \quad}}} &       & \multicolumn{5}{c}{BRAC-p} \\
		\cmidrule(r){3-7}
		& Random & 9.6 $\pm$ 3.3 & \textbf{11.2} $\pm$ 0.8 & 11 $\pm$ 0.6 & 10.6 $\pm$ 2.4 & 5.3 $\pm$ 1.2 \\
		& Medium & 29.2 $\pm$ 2.1 & 26.5 $\pm$ 1.8 & \textbf{32.9} $\pm$ 7.5 & 16.1 $\pm$ 0.9 & 16.7 $\pm$ 1.7 \\
		& Medium-R & 20.1 $\pm$ 4.8 & 17.8 $\pm$ 3.2 & \textbf{30.8} $\pm$ 4.9 & 13.9 $\pm$ 1.7 & 10.4 $\pm$ 2.4 \\
		& Medium-E & 32.3 $\pm$ 7.8 & \textbf{40.4} $\pm$ 4.4 & 34.7 $\pm$ 8.5 & 29.4 $\pm$ 6.5 & 25.2 $\pm$ 4.1 \\
		\midrule
		\multirow{5}[2]{*}{\rotatebox{0}{\colorbox{magiccolor}{\quad Hopper \quad}}} &       & \multicolumn{5}{c}{AWR} \\
		\cmidrule(r){3-7}
		& Random & 3.4 $\pm$ 0.7 & 4.1 $\pm$ 1 & \textbf{4.5} $\pm$ 0.9 & 3.4 $\pm$ 0.7 & 2.5 $\pm$ 0.1 \\
		& Medium & 20.8 $\pm$ 6.3 & 31.8 $\pm$ 2.9 & \textbf{28.9} $\pm$ 5.5 & 26.6 $\pm$ 3.2 & 17.4 $\pm$ 1.5 \\
		& Medium-R & 4.1 $\pm$ 1.7 & 3 $\pm$ 0.5 & 4.2 $\pm$ 3.5 & 2.6 $\pm$ 0.6 & \textbf{4.3} $\pm$ 1.3 \\
		& Medium-E & 26.8 $\pm$ 0.4 & \textbf{27} $\pm$ 0 & 26.6 $\pm$ 2 & 17.8 $\pm$ 5.6 & 24.2 $\pm$ 3.9 \\
		\midrule
		\multirow{5}[2]{*}{\rotatebox{0}{\colorbox{magiccolor}{\quad Hopper \quad}}} &       & \multicolumn{5}{c}{BCQ} \\
		\cmidrule(r){3-7}
		& Random & 8.3 $\pm$ 0.3 & 9.6 $\pm$ 0.3 & \textbf{9.7} $\pm$ 0.2 & 7.4 $\pm$ 0.1 & 7.6 $\pm$ 0.3 \\
		& Medium & 25.7 $\pm$ 5.5 & 24.1 $\pm$ 0.8 & \textbf{38.4} $\pm$ 1.8 & 27.1 $\pm$ 1.7 & 26.7 $\pm$ 0.8 \\
		& Medium-R & 28.7 $\pm$ 1.9 & 29.5 $\pm$ 3 & \textbf{32.8} $\pm$ 0.9 & 25.9 $\pm$ 6 & 21 $\pm$ 2.2 \\
		& Medium-E & 75.4 $\pm$ 7.8 & 70.4 $\pm$ 5.4 & \textbf{84.2} $\pm$ 9.8 & 67.9 $\pm$ 8 & 61.9 $\pm$ 4.3 \\
		\midrule
		\multirow{5}[2]{*}{\rotatebox{0}{\colorbox{magiccolor}{\quad Hopper \quad}}} &       & \multicolumn{5}{c}{CQL} \\
		\cmidrule(r){3-7}
		& Random & 10.2 $\pm$ 0.3 & 10 $\pm$ 0 & \textbf{10.4} $\pm$ 0.4 & 10 $\pm$ 0 & 10 $\pm$ 0 \\
		& Medium & 44.9 $\pm$ 2.7 & \textbf{59.8} $\pm$ 6 & 59.3 $\pm$ 12.2 & 44.2 $\pm$ 1 & 37.1 $\pm$ 2.6 \\
		& Medium-R & 1.4 $\pm$ 0.3 & 2.1 $\pm$ 0.2 & 3.7 $\pm$ 1.4 & \textbf{3.9} $\pm$ 1.7 & 3.4 $\pm$ 1 \\
		& Medium-E & 53.6 $\pm$ 21.2 & 65.3 $\pm$ 15.4 & \textbf{99.7} $\pm$ 16.4 & 60.5 $\pm$ 16 & 75.9 $\pm$ 30 \\
		\midrule
		\multirow{5}[2]{*}{\rotatebox{0}{\colorbox{magiccolor}{\quad Hopper \quad}}} &       & \multicolumn{5}{c}{MOPO} \\
		\cmidrule(r){3-7}
		& Random & 2 $\pm$ 2.1 & 1.8 $\pm$ 0 & \textbf{2.1} $\pm$ 1.7 & 1.2 $\pm$ 0.4 & 0.8 $\pm$ 0 \\
		& Medium & 5 $\pm$ 5.3 & 6.5 $\pm$ 1 & \textbf{10.7} $\pm$ 5.1 & 5.3 $\pm$ 1.6 & 2.8 $\pm$ 0.7 \\
		& Medium-R & 5.5 $\pm$ 4.6 & 7.5 $\pm$ 0.8 & \textbf{8.4} $\pm$ 3.5 & 5.7 $\pm$ 3.5 & 1.9 $\pm$ 0.6 \\
		& Medium-E & 4.8 $\pm$ 2.9 & \textbf{8.1} $\pm$ 1 & 5.8 $\pm$ 2.3 & 4.7 $\pm$ 0.7 & 2.1 $\pm$ 0.2 \\
		\bottomrule
	\end{tabular}%
	\label{tab:hyper_eta}%
\end{table}%

\subsection{Environments and dataset}
\label{appendix:environments-and-dataset}

\begin{figure}[H]
	\centering
	\includegraphics[scale=0.250]{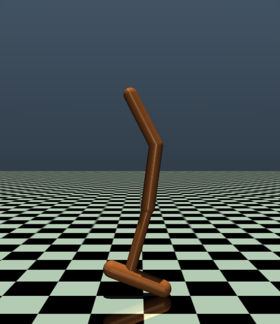}
	\includegraphics[scale=0.250]{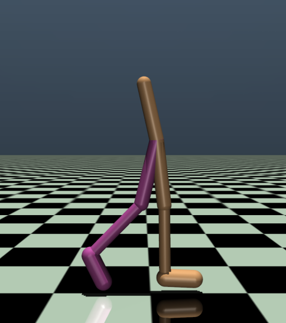}
	\includegraphics[scale=0.250]{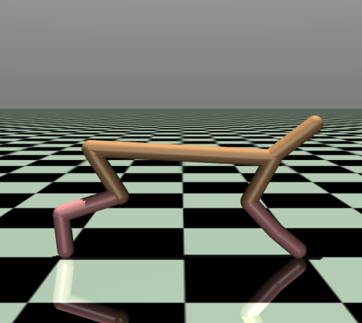}
	\includegraphics[scale=0.250]{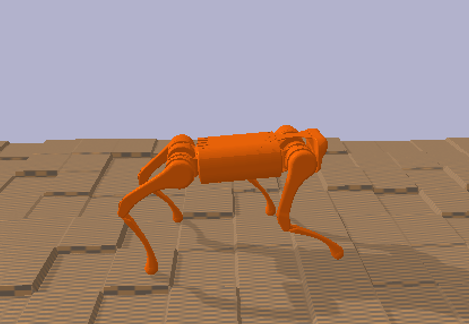}
	\includegraphics[scale=0.250]{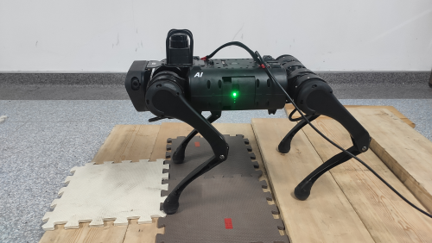}
	\centering
	\caption{Illustration of the suite of tasks considered in this work: (from left to right) Hopper, Walker2d, Halfcheetah, simulated and real-world quadruped robots. These tasks require the RL agent to learn locomotion gaits for the illustrated characters.}
	\label{appendix:environmnts-illustration}
\end{figure}

In this work, the tasks include Hopper, Walker2d, HalfCheetah, simulated (see the dynamics parameters in  \cite{zhang2021terrain}) and real-world quadruped robot, which are illustrated in Figure~\ref{appendix:environmnts-illustration}.

\begin{table}[h]
	\centering
	\caption{Dynamics shift for Hopper, Walker2d, and Halfcheetah tasks. For the body mass shift, we change the mass of the body in the source MDP $\MS$. For the joint noise shift, we add a noise (randomly sampling in $[-0.05, +0.05]$) to the actions when we collect the source offline data, \ie $\DS := \{(\bs,\ba,r,\bs')\} \sim d_\DS(\bs)\pibs(\ba|\bs)r(\bs,\ba)\TS(\bs'|\bs,\ba+\text{noise})$.}
	\begin{adjustbox}{max width=1.0\textwidth}
	\begin{tabular}{lcccccc}
		\toprule
		& \multicolumn{2}{c}{Hopper} & \multicolumn{2}{c}{Walker2d} & \multicolumn{2}{c}{HalfCheetah} \\
		\cmidrule(r){2-3}  \cmidrule(r){4-5} \cmidrule(r){6-7}
		& \multicolumn{1}{l}{Body Mass Shfit} & \multicolumn{1}{l}{Joint Noise Shift} & \multicolumn{1}{l}{Body Mass Shfit} & \multicolumn{1}{l}{Joint Noise Shift} & \multicolumn{1}{l}{Body Mass Shfit} & \multicolumn{1}{l}{Joint Noise Shift} \\
		Source &  mass[-1]=2.5     &  action[-1]+noise     &  mass[-1]=1.47     &   action[-1]+noise    &   mass[4]=0.5    & action[-1]+noise \\
		Target &  mass[-1]=5.0     &  action[-1]+0    &  mass[-1]=2.94     &   action[-1]+0    &   mass[4]=1.0    & action[-1]+0 \\
		\bottomrule 
	\end{tabular}
	\label{appendix:tab:dynamics-shift-mass-joint}
	\end{adjustbox}
\end{table}

In the Hopper, Walker2d and HalfCheetah dynamics adaptation setting, we set the D4RL~\citep{d4rl2020Fu} {dataset} as our target {domain}. 
For the source dynamics, we change the body mass (body mass shift) or add noises to joints (joint noise shift) of the agents (see Table~\ref{appendix:tab:dynamics-shift-mass-joint} for the details) and then collect the source offline dataset in the changed environment. 
Following \citet{d4rl2020Fu}, on the changed {source} environment, we collect the 1) "Random" offline data, generated by unrolling a randomly initialized policy, 2) "Medium" offline data,  generated by a trained policy with the “medium” level of performance {in the source environment}, 3) "Medium-Replay" (Medium-R) offline data, consisting of recording all samples in the replay buffer observed during training until the policy reaches the “medium” level of performance, 4) "Medium-Expert" (Medium-E)  offline data, mixing equal amounts of expert demonstrations and "medium" data  {in the source environment}.

\begin{wrapfigure}{r}{0.345\textwidth}
	\vspace{-11pt}
	\begin{center}
		\includegraphics[scale=0.2345]{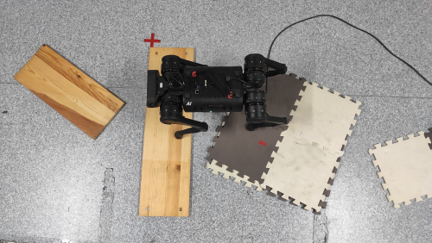}
	\end{center}
	\vspace{-5pt}
	\caption{Real-world terrains (for collecting the target offline data).}
	\vspace{-5pt}
	\label{appendix:environments-illustration-real-world-terrains}
\end{wrapfigure}
In the sim2real setting (for the quadruped robot), we use the A1 dog from Unitree~\citep{unitree}. We collect the target offline data  using five target behavior policies in the real-world with changing terrains, as shown in Figure~\ref{appendix:environments-illustration-real-world-terrains}, and collect the "Medium", "Medium-Replay" (Medium-R), "Medium-Expert" (Medium-E), "Medium-Replay-Expert" (Medium-R-E) source offline data  in the simulator, where "Medium-Replay-Expert" denotes mixing equal amounts of "Medium-Replay" data and expert demonstrations  {in the simulator}. In Section~\ref{appendix:sec:training-a1-dog}, we provide the details of how to obtain the target  and source behavior policy, so as to collect our target and source offline data. 

\begin{table}[t]
	\centering
	\caption{Statistics for each task in our adaptation setting. }
	\begin{adjustbox}{max width=\textwidth}
		\begin{tabular}{lllll}
			\toprule 
			\textbf{Environment} & \multicolumn{1}{l}{\textbf{Dynamics Shift}} & \textbf{Task Name} & \textbf{Target} (1T) & \textbf{Source} (10S)\\
			\midrule
			\multicolumn{1}{l}{\multirow{8}[4]{*}{Hopper}} & \multirow{4}[2]{*}{Body Mass Shfit} & Random & $10^5$ (D4RL) & $10^6$ \\
			&       & Medium & $10^5$ (D4RL) & $10^6$ \\
			&       & Medium-Replay & $20092$ (D4RL) & $10^6$ \\
			&       & Medium-Expert & $2*10^5$ (D4RL) & $2*10^6$ \\
			\cmidrule{2-5}          & \multirow{4}[2]{*}{Joint Noise Shift} & Random & $10^5$ (D4RL) & $10^6$ \\
			&       & Medium & $10^5$ (D4RL) & $10^6$ \\
			&       & Medium-Replay & $20092$ (D4RL) & $10^6$ \\
			&       & Medium-Expert & $2*10^5$ (D4RL) & $2*10^6$ \\
			\midrule
			\multicolumn{1}{l}{\multirow{8}[4]{*}{Walker2d}} & \multirow{4}[2]{*}{Body Mass Shfit} & Random & $10^5$ (D4RL) & $10^6$ \\
			&       & Medium & $10^5$ (D4RL) & $10^6$ \\
			&       & Medium-Replay & $10093$ (D4RL) & $10^6$ \\
			&       & Medium-Expert & $2*10^5$ (D4RL) & $2*10^6$ \\
			\cmidrule{2-5}          & \multirow{4}[2]{*}{Joint Noise Shift} & Random & $10^5$ (D4RL) & $10^6$ \\
			&       & Medium & $10^5$ (D4RL) & $10^6$ \\
			&       & Medium-Replay & $10093$ (D4RL) & $10^6$ \\
			&       & Medium-Expert & $2*10^5$ (D4RL) & $2*10^6$ \\
			\midrule
			\multicolumn{1}{l}{\multirow{8}[4]{*}{HalfCheetah}} & \multirow{4}[2]{*}{Body Mass Shfit} & Random & $10^5$ (D4RL) & $10^6$ \\
			&       & Medium & $10^5$ (D4RL) & $10^6$ \\
			&       & Medium-Replay & $10100$ (D4RL) & $10^6$ \\
			&       & Medium-Expert & $2*10^5$ (D4RL) & $2*10^6$ \\
			\cmidrule{2-5}          & \multirow{4}[2]{*}{Joint Noise Shift} & Random & $10^5$ (D4RL) & $10^6$ \\
			&       & Medium & $10^5$ (D4RL) & $10^6$ \\
			&       & Medium-Replay & $10100$ (D4RL) & $10^6$ \\
			&       & Medium-Expert & $2*10^5$ (D4RL) & $2*10^6$ \\
			\midrule
			\multirow{4}[1]{*}{A1 robot (Unitree)} & \multirow{4}[1]{*}{Sim2Real} & Medium & $3*10^4$ (real-world) & $10^6$ (simulator) \\
			&       & Medium-Replay & $3*10^4$ (real-world) & $10^6$ (simulator) \\
			&       & Medium-Expert & $3*10^4$ (real-world) & $2*10^6$ (simulator) \\
			&       & Medium-Replay-Expert  & $3*10^4$ (real-world) & $2*10^6$ (simulator) \\
			\bottomrule 
		\end{tabular}%
		\label{appendix:environment-statistics-samples}
	\end{adjustbox}
\end{table}

We list our tasks properties in Table~\ref{appendix:environment-statistics-samples} and provide our collected dataset in supplementary material. {In implementation, we set} $\eta = 0.1$ {for all simulated tasks and set} $\eta = 0.01$ {for the sim2real task. In} Table~\ref{tab:hyper_eta}, {we also report the sensitivity of DARA on the hyper-parameters} $\eta$.

\subsection{Training the (target and source) behavior policy for the quadruped robot}
\label{appendix:sec:training-a1-dog}

To obtain a behavior policy that can be deployed in simulator (for collecting the source offline data) or real-world (for collecting the target offline data), we introduce the prior knowledge~\citep{priorknow2018iscen} and domain randomization~\citep{2017Domain, sim2real2018peng}.

\textbf{Prior Knowledge:} 
To reduce the impact of the foot at the moment of touching the ground during the robot locomotion, we designed a compound cycloid trajectory \citep{foottraj1990Sakakibara} as prior knowledge. 
In our implementation for the foot trajectory, four aspects are mainly considered: 1) The robot walks stably without obvious shaking; 
2) The joint impact of the robot during the locomotion is small; 
3) The joint speed and acceleration of the robot during the locomotion are continuous and smooth; 
4) The feet of the robot will not slide when they are in contact with the ground. 
Similar to ~\citet{eth2020}, we define a periodic phase variable $\phi_i \in [0.0,0.6), i={1,2,3,4}$ for each leg, which represents swing phase if $\phi_i \in [0.0,0.3)$ and contact phase if $\phi_i \in [0.3,0.6)$. 
At every time step $t$, $\phi_i = (t*f_0 + \phi_0[i] + \phi_{\text{offset}}[{i}]) (\text{mod\ } 2T_m) $ where $T_m=0.3$, and $f_0=1.1$ is the base frequency, and $\phi_0=[0,0.3,0.3,0]$ is the initial phase. 
$\phi_{\text{offset}}$ is part of the output of the controller. 
The trajectory of the swing leg is:
\begin{equation}
\left\{\begin{array}{lc}
x_i= S\left[\frac{t}{T_{m}}-\frac{1}{2 \pi} \sin \left(\frac{2 \pi t}{T_{m}}\right)\right] + S_{0},i=1,2 \\
x_i= S\left[\frac{t}{T_{m}}-\frac{1}{2 \pi} \sin \left(\frac{2 \pi t}{T_{m}}\right)\right]-S+S_{0},i=3,4 \\
y = Y_{0} \\
z= H\left[\operatorname{sgn}\left(\frac{T_{m}}{2}-t\right)\left(2 f_{E}(t)-1\right)+1\right]+Z_{0}
\end{array},\right. \nonumber 
\end{equation}
where
\begin{equation}
f_{E}(t)=\frac{t}{T_{m}}-\frac{1}{4 \pi} \sin \left(\frac{4 \pi t}{T_{m}}\right), \nonumber 
\end{equation}
and 
\begin{equation}
\operatorname{sgn}\left(\frac{T_{m}}{2}-t\right)=\left\{\begin{array}{rl}
1 & 0 \le t<\frac{T_{m}}{2} \\
-1 & \frac{T_{m}}{2} \le t<T_{m}
\end{array}.\right.  \nonumber
\end{equation}
The trajectory  of the standing leg is: 
\begin{equation}
\left\{\begin{array}{l}
x_i=S\left(\frac{2 T_{m}-t}{T_{m}}+\frac{1}{2 \pi} \sin \left(\frac{2 \pi t}{T_{m}}\right)\right)+S_0,i=1,2 \\
x_i=S\left(\frac{2 T_{m}-t}{T_{m}}+\frac{1}{2 \pi} \sin \left(\frac{2 \pi t}{T_{m}}\right)\right)-S+S_0,i=3,4 \\
y=Y_0 \\
z=Z_0 
\end{array}.\right.  \nonumber
\end{equation} 
where $S=0.14m, H=0.18m$ are the maximum foot length and height.
$
S_0=[0.17,0.17,-0.2,-0.2], Y_0=[-0.13,0.13,-0.13,0.13],  Z_0=[-0.32,-0.32,-0.32,-0.32]
$
are the default target foot position in body frame. 

\textbf{Domain Randomization:}  To encourage the policy to be robust to variations in the dynamics, we incorporate the domain randomization. In Table~\ref{appendix:tab:dog-domain-randomization}, we provide the dynamics parameters and their respective range of values. 
\begin{table}[h]
	\centering
	\caption{Dynamic parameters and their respective range of values utilized during training.
	}\label{appendix:tab:dog-domain-randomization}
	\begin{tabular}{ l l }
		\toprule
		\textbf{Parameter} & \textbf{Range}      \\ \midrule 
		Mass             & {[0.95, 1.1]} $\times \text{default value}$  \\  
		Inertia          & {[0.80, 1.2]}  $\times \text{default value}$  \\  
		Motor Strength   & {[0.80, 1.2]}  $\times \text{default value}$  \\  
		Latency          & {[}0, 0.04{]} \sl{\it{s}}          \\  
		Lateral Friction & {[}0.5, 1.25{]} \sl{\it{Ns/m}}            \\  
		Joint Friction & {[}0, 0.05{]} \sl{\it{Nm}}       \\  
		\bottomrule 
	\end{tabular}
\end{table}

\textbf{State Space, Action Space and Reward Function:}  
The action is a 16-dimensional vector consisting of leg phase and target foot position residuals in the body frame. 
The design of state space and reward function mainly follows the prior work \citet{eth2020}. 
In Table \ref{tab.obs}, we provide the state representation. 
\begin{table}[h]
	\centering
	\caption{
		State representation for the behavior policy.
	}
	\scalebox{1.}{
		\begin{tabular}{lc}
			\toprule
			\textbf{Data} & \textbf{Dimension} \\ \midrule
			Desired direction$\left(\left(_{I B}^{B}{ \hat{v}_{d}}\right)_{x y}\right)$            & 2         \\
			Euler angle$\left(rpy\right)$                                                           & 3         \\
			Base angular velocity$\left(_{I B}^{B}{\omega}\right)$                                 & 3         \\
			Base linear velocity$\left(_{I B}^{B}{v}\right)$                                       & 3         \\
			Joint position/velocity$\left(\theta_{i}, \dot{\theta}_{i}\right)$                           & 24        \\
			FTG phases$\left(\sin \left(\phi_{i}\right), \cos \left(\phi_{i}\right)\right)$              & 8         \\
			FTG frequencies$\left(f_{i}\right)$                                                 & 4         \\
			Base frequency$\left(f_{o}\right)$                                                           & 1         \\
			Joint position error history                                                                 & 24        \\
			Joint velocity history                                                                       & 24        \\
			Foot target history$\left(\left(r_{f, d}\right)_{t-1, t-2}\right)$                           & 24        \\ 
			\bottomrule 
		\end{tabular}
	}
	\label{tab.obs}
\end{table}

The reward function is defined as $$0.1r_{lv}+0.05r_{y}+0.05r_{rp}+0.005r_{b}+0.02r_{bc}+0.025r_{s}+2\cdot10^{-5}r_{\tau}.$$
The individual terms are defined as follows. 

1) Linear velocity reward $r_{lv}$ : 
\begin{equation}
r_{lv}:=
\left\{
\begin{array}{lcl}
exp(-30|v_{pr} - 0.2|) &  v_{pr}<0.2 \\
1 & v_{pr} \ge 0.2
\end{array}
\right., \nonumber 
\end{equation}
where $v_{pr} = v_{xy}\cdot \hat{v}_{xy}$ is the base linear velocity projected onto the command direction.

2) Yaw angle reward  $r_{y}$ : 
\begin{equation}
r_{y}:= exp(-(y - \hat{y})^2),
\end{equation}
where $y$ and $\hat{y}$ is the yaw and desired yaw angle.

3) Roll and pitch reward $r_{rp}$ : 
\begin{equation}
r_{rp}:= exp(-1.5\sum(\phi - [0,\arccos(\frac{<P_{xz},(0,0,1)^T>}{\parallel P_{xz} \parallel})-\pi/2])^2),
\end{equation}
where $\phi$ are the roll and pitch angle.
$P_{xz}=P_1-P_4$ or $P_{xz}=P_2-P_3$,$P_i, i \in[1,4]$ are the foot position in world frame.
The advantage of designing the target pitch angle in this way is to ensure that the body of the robot is parallel to the supporting surface of the stand legs, thereby ensuring that the robot can smoothly over challenge terrain, such as upward stairs. 

4) Base motion reward $r_b$ : 
\begin{equation}
r_{b}:= exp(-1.5\sum(v_{xy}-v_{pr}*\hat{v}_{xy})^2)+ exp(-1.5\sum(\omega_{xy})^2),
\end{equation}
where $\omega_{xy}$ are the roll and pitch rates.

5) Body collision reward $r_{bc}$ : 
\begin{equation}
r_{bc}:= -|I_{body}/I_{foot}|,
\end{equation}
where $I_{body}$ and $I_{foot}$ are the contact numbers of robot's body parts and foot with the terrain, respectively. 

6) Target smooth reward $r_s$ : 
\begin{equation}
r_{s}:= -||f_{d,t}-2f_{d,t-1}+f_{d,t-2}||,
\end{equation}
where $f_{d,i} (i=t, t-1,t-2)$ are the target foot positions in the time-step $t$, $t-1$ and $t-2$. 

7) Torqure reward $r_\tau$: 
\begin{equation}
r_{\tau}:= -\sum_{i}|\tau_i|,
\end{equation}
where $\tau_i$ is the joint torques. 

\textbf{Training Details:}  
Both the behavior policy and value networks are Multilayer Perceptron (MLP) with 3 hidden layers, which have 256, 128 and 64 nodes. 
The activation function is the \textit{Tanh} function, and the optimizer is \textit{Adam}. 
With the above prior knowledge, domain randomization and reward function, we train our behavior policy with SAC~\citep{SAC2018harrnoja} in  PyBullet~\citep{coumans2021}.

\subsection{Additional Results}

Here we provide additional results regarding the error bars (Tables~\ref{tab:error_bar:body_mass_shift}~and~\ref{tab:error_bar:joint_noise_shift}).

\begin{sidewaystable}[htbp]
	\centering
	\caption{Normalized scores for the D4RL tasks (with body mass shift). We take the baseline results (for 10T) of MOPO from their original papers and that of the other model-free methods (BEAR, BRAC-p, AWR, BCQ and CQL) from the D4RL paper~\citep{d4rl2020Fu}. }
	\begin{tabular}{clrlllrlllrlll}
		\toprule
		\multicolumn{2}{c}{\multirow{2}[1]{*}{\textcolor{black}{Body Mass Shift}}} & \multicolumn{1}{c}{\multirow{2}[1]{*}{10T}} & \multicolumn{1}{c}{\multirow{2}[1]{*}{1T}} & 1T+10S & 1T+10S & \multicolumn{1}{c}{\multirow{2}[1]{*}{10T}} & \multicolumn{1}{c}{\multirow{2}[1]{*}{1T}} & 1T+10S & 1T+10S & \multicolumn{1}{c}{\multirow{2}[1]{*}{10T}} & \multicolumn{1}{c}{\multirow{2}[1]{*}{1T}} & 1T+10S & 1T+10S \\
		\multicolumn{2}{c}{} &       &       & w/o Aug. & (DARA) &       &       & w/o Aug. & (DARA) &       &       & w/o Aug. & (DARA) \\
		\cmidrule{1-14}    \multirow{5}[4]{*}{\rotatebox{90}{\colorbox{magiccolor}{\quad Hopper \quad}}} &       & \multicolumn{4}{c}{BEAR}      & \multicolumn{4}{c}{BRAC-p}    & \multicolumn{4}{c}{AWR} \\
		\cmidrule(r){3-6} \cmidrule(r){7-10} \cmidrule(r){11-14}          & Random & 11.4  & 1 $\pm$ 0.5 & 4.6 $\pm$ 3.4 & 8.4 $\pm$ 1.2 & 11    & 10.9 $\pm$ 0.1 & 9.6 $\pm$ 3.3 & 11 $\pm$ 0.6 & 10.2  & 10.3 $\pm$ 0.3 & 3.4 $\pm$ 0.7 & 4.5 $\pm$ 0.9 \\
		& Medium & 52.1  & 0.8 $\pm$ 0 & 0.9 $\pm$ 0.3 & 1.6 $\pm$ 1 & 32.7  & 29 $\pm$ 6.2 & 29.2 $\pm$ 2.1 & 32.9 $\pm$ 7.5 & 35.9  & 30.9 $\pm$ 0.4 & 20.8 $\pm$ 6.3 & 28.9 $\pm$ 5.5 \\
		& Medium-R & 33.7  & 1.3 $\pm$ 1.5 & 18.2 $\pm$ 5 & 34.1 $\pm$ 5.8 & 0.6   & 5.4 $\pm$ 3.3 & 20.1 $\pm$ 4.8 & 30.8 $\pm$ 4.9 & 28.4  & 8.8 $\pm$ 4.9 & 4.1 $\pm$ 1.7 & 4.2 $\pm$ 3.5 \\
		& Medium-E & 96.3  & 0.8 $\pm$ 0.1 & 0.6 $\pm$ 0 & 1.2 $\pm$ 0.5 & 1.9   & 34.5 $\pm$ 14.7 & 32.3 $\pm$ 7.8 & 34.7 $\pm$ 8.5 & 27.1  & 27 $\pm$ 1.3 & 26.8 $\pm$ 0.4 & 26.6 $\pm$ 2 \\
		\cmidrule{1-14}    \multirow{5}[4]{*}{\rotatebox{90}{\colorbox{magiccolor}{\quad Hopper \quad}}} &       & \multicolumn{4}{c}{BCQ}       & \multicolumn{4}{c}{CQL}       & \multicolumn{4}{c}{MOPO} \\
		\cmidrule(r){3-6} \cmidrule(r){7-10} \cmidrule(r){11-14}          & Random & 10.6  & 10.6 $\pm$ 0.1 & 8.3 $\pm$ 0.3 & 9.7 $\pm$ 0.2 & 10.8  & 10.6 $\pm$ 0.1 & 10.2 $\pm$ 0.3 & 10.4 $\pm$ 0.4 & 11.7  & 4.8 $\pm$ 2.4 & 2 $\pm$ 2.1 & 2.1 $\pm$ 1.7 \\
		& Medium & 54.5  & 37.1 $\pm$ 6.3 & 25.7 $\pm$ 5.5 & 38.4 $\pm$ 1.8 & 58    & 43 $\pm$ 9.2 & 44.9 $\pm$ 2.7 & 59.3 $\pm$ 12.2 & 28    & 4.1 $\pm$ 2 & 5 $\pm$ 5.3 & 10.7 $\pm$ 5.1 \\
		& Medium-R & 33.1  & 9.3 $\pm$ 4.4 & 28.7 $\pm$ 1.9 & 32.8 $\pm$ 0.9 & 48.6  & 9.6 $\pm$ 5.2 & 1.4 $\pm$ 0.3 & 3.7 $\pm$ 1.4 & 67.5  & 1 $\pm$ 0.6 & 5.5 $\pm$ 4.6 & 8.4 $\pm$ 3.5 \\
		& Medium-E & 110.9 & 58 $\pm$ 16.2 & 75.4 $\pm$ 7.8 & 84.2 $\pm$ 9.8 & 98.7  & 59.7 $\pm$ 34.5 & 53.6 $\pm$ 21.2 & 99.7 $\pm$ 16.4 & 23.7  & 1.6 $\pm$ 0.6 & 4.8 $\pm$ 2.9 & 5.8 $\pm$ 2.3 \\
		\cmidrule{1-14}    \multirow{5}[4]{*}{\rotatebox{90}{\colorbox{magiccolor}{ Walker2d \quad}}} &       & \multicolumn{4}{c}{BEAR}      & \multicolumn{4}{c}{BRAC-p}    & \multicolumn{4}{c}{AWR} \\
		\cmidrule(r){3-6} \cmidrule(r){7-10} \cmidrule(r){11-14}          & Random & 7.3   & 1.5 $\pm$ 0.9 & 3.1 $\pm$ 0.9 & 3.2 $\pm$ 0.4 & -0.2  & 0 $\pm$ 0.2 & 1.3 $\pm$ 0.7 & 3.2 $\pm$ 2.5 & 1.5   & 1.3 $\pm$ 0.4 & 2 $\pm$ 1 & 2.4 $\pm$ 0.8 \\
		& Medium & 59.1  & -0.5 $\pm$ 0.3 & 0.6 $\pm$ 0.5 & 0.3 $\pm$ 0.7 & 77.5  & 6.4 $\pm$ 9.9 & 70 $\pm$ 10.1 & 78 $\pm$ 3.1 & 17.4  & 14.8 $\pm$ 2.8 & 17.1 $\pm$ 0.2 & 17.2 $\pm$ 0.1 \\
		& Medium-R & 19.2  & 0.7 $\pm$ 0.6 & 6.5 $\pm$ 5.1 & 7.3 $\pm$ 1.3 & -0.3  & 8.5 $\pm$ 2.2 & 9.9 $\pm$ 2 & 18.6 $\pm$ 6.5 & 15.5  & 7.4 $\pm$ 2.1 & 1.6 $\pm$ 0.4 & 1.5 $\pm$ 0.3 \\
		& Medium-E & 40.1  & -0.1 $\pm$ 0.1 & 1.5 $\pm$ 2.5 & 2.3 $\pm$ 2.2 & 76.9  & 20.6 $\pm$ 16.8 & 64.1 $\pm$ 10.8 & 77.5 $\pm$ 3.1 & 53.8  & 35.5 $\pm$ 10.4 & 52.5 $\pm$ 1.2 & 53.3 $\pm$ 0.3 \\
		\cmidrule{1-14}    \multirow{5}[3]{*}{\rotatebox{90}{\colorbox{magiccolor}{ Walker2d \quad}}} &       & \multicolumn{4}{c}{BCQ}       & \multicolumn{4}{c}{CQL}       & \multicolumn{4}{c}{MOPO} \\
		\cmidrule(r){3-6} \cmidrule(r){7-10} \cmidrule(r){11-14}         & Random & 4.9   & 1.8 $\pm$ 0.9 & 4.5 $\pm$ 0.5 & 4.8 $\pm$ 0.3 & 7     & 1.7 $\pm$ 1.3 & 3.2 $\pm$ 1.4 & 3.4 $\pm$ 1.9 & 13.6  & -0.2 $\pm$ 0.2 & -0.1 $\pm$ 0.1 & -0.1 $\pm$ 0.2 \\
		& Medium & 53.1  & 32.8 $\pm$ 8.2 & 50.9 $\pm$ 4.3 & 52.3 $\pm$ 1.4 & 79.2  & 42.9 $\pm$ 24.2 & 80 $\pm$ 1.2 & 81.7 $\pm$ 3.1 & 17.8  & 7 $\pm$ 3.6 & 5.7 $\pm$ 4.7 & 11 $\pm$ 4.3 \\
		& Medium-R & 15    & 6.9 $\pm$ 0.6 & 14.9 $\pm$ 0.2 & 15.1 $\pm$ 0.2 & 26.7  & 4.6 $\pm$ 3.9 & 0.8 $\pm$ 0.5 & 2 $\pm$ 1.5 & 39    & 5.1 $\pm$ 5.7 & 3.1 $\pm$ 2.4 & 14.2 $\pm$ 4.5 \\
		& Medium-E & 57.5  & 32.5 $\pm$ 9.1 & 55.2 $\pm$ 3.8 & 57.2 $\pm$ 0.2 & 111   & 49.5 $\pm$ 26.7 & 63.5 $\pm$ 22.5 & 93.3 $\pm$ 8.8 & 44.6  & 5.3 $\pm$ 3.9 & 5.5 $\pm$ 3.5 & 17.2 $\pm$ 8.7 \\
		\bottomrule
	\end{tabular}%
	\label{tab:error_bar:body_mass_shift}%
\end{sidewaystable}

\begin{sidewaystable}[htbp]
	\centering
	\caption{Normalized scores for the D4RL tasks (with joint noise shift). We take the baseline results (for 10T) of MOPO from their original papers and that of the other model-free methods (BEAR, BRAC-p, AWR, BCQ and CQL) from the D4RL paper~\citep{d4rl2020Fu}. }
	\begin{tabular}{clrlllrlllrlll}
		\toprule
		\multicolumn{2}{c}{\multirow{2}[1]{*}{\textcolor{black}{Joint Noise Shift}}} & \multicolumn{1}{c}{\multirow{2}[1]{*}{10T}} & \multicolumn{1}{c}{\multirow{2}[1]{*}{1T}} & 1T+10S & 1T+10S & \multicolumn{1}{c}{\multirow{2}[1]{*}{10T}} & \multicolumn{1}{c}{\multirow{2}[1]{*}{1T}} & 1T+10S & 1T+10S & \multicolumn{1}{c}{\multirow{2}[1]{*}{10T}} & \multicolumn{1}{c}{\multirow{2}[1]{*}{1T}} & 1T+10S & 1T+10S \\
		\multicolumn{2}{c}{} &       &       & w/o Aug. & (DARA) &       &       & w/o Aug. & (DARA) &       &       & w/o Aug. & (DARA) \\
		\cmidrule{1-14}    \multirow{5}[4]{*}{\rotatebox{90}{\colorbox{magiccolor}{\quad Hopper \quad}}} &       & \multicolumn{4}{c}{BEAR}      & \multicolumn{4}{c}{BRAC-p}    & \multicolumn{4}{c}{AWR} \\
		\cmidrule(r){3-6} \cmidrule(r){7-10} \cmidrule(r){11-14}          & Random & 11.4  & 0.6 $\pm$ 0 & 7.4 $\pm$ 0.5 & 4.2 $\pm$ 3.6 & 11 & 10.8 $\pm$ 0.2 & 10 $\pm$ 0.8 & 10.8 $\pm$ 0 & 10.2 & 10.1 $\pm$ 0 & 3.6 $\pm$ 0 & 4 $\pm$ 0.4 \\
		& Medium & 52.1  & 0.8 $\pm$ 0 & 2 $\pm$ 1 & 2 $\pm$ 0.1 & 32.7 & 26.6 $\pm$ 4.8 & 27.6 $\pm$ 2.8 & 37.6 $\pm$ 7 & 35.9 & 30.3 $\pm$ 0.1 & 38.8 $\pm$ 3.9 & 41.3 $\pm$ 5.4 \\
		& Medium-R & 33.7  & 2.7 $\pm$ 1.6 & 3.6 $\pm$ 0.4 & 9.9 $\pm$ 6.3 & 0.6 & 13.4 $\pm$ 6.4 & 89.9 $\pm$ 7.8 & 101.4 $\pm$ 0.2 & 28.4 & 12.4 $\pm$ 6 & 6.7 $\pm$ 4.2 & 7.2 $\pm$ 0.2 \\
		& Medium-E & 96.3  & 0.8 $\pm$ 0.2 & 0.8 $\pm$ 0 & 1.4 $\pm$ 0.6 & 1.9 & 19.8 $\pm$ 14 & 57.6 $\pm$ 23.4 & 87.8 $\pm$ 13.3 & 27.1 & 25.5 $\pm$ 1.2 & 27 $\pm$ 0 & 27 $\pm$ 0.1 \\
		\cmidrule{1-14}    \multirow{5}[4]{*}{\rotatebox{90}{\colorbox{magiccolor}{\quad Hopper \quad}}} &       & \multicolumn{4}{c}{BCQ}       & \multicolumn{4}{c}{CQL}       & \multicolumn{4}{c}{MOPO} \\
		\cmidrule(r){3-6} \cmidrule(r){7-10} \cmidrule(r){11-14}          & Random & 10.6  & 10.5 $\pm$ 0.1 & 7 $\pm$ 0 & 9.6 $\pm$ 0 & 10.8 & 10.4 $\pm$ 0.1 & 10.4 $\pm$ 0.4 & 10.8 $\pm$ 0 & 11.7 & 1.5 $\pm$ 0.8 & 1.3 $\pm$ 0.5 & 2.9 $\pm$ 1.5 \\
		& Medium & 54.5  & 45.8 $\pm$ 2.2 & 49 $\pm$ 1.7 & 54.4 $\pm$ 0.1 & 58 & 46.2 $\pm$ 11.9 & 58 $\pm$ 0 & 58 $\pm$ 0 & 28 & 2.7 $\pm$ 2.1 & 9.2 $\pm$ 5.4 & 17.3 $\pm$ 3.4 \\
		& Medium-R & 33.1  & 13 $\pm$ 5 & 23.8 $\pm$ 3.2 & 32 $\pm$ 0.9 & 48.6 & 13.6 $\pm$ 6.4 & 2.6 $\pm$ 0.3 & 3.6 $\pm$ 0.6 & 67.5 & 0.8 $\pm$ 0.1 & 2.3 $\pm$ 1.7 & 6.4 $\pm$ 0.8 \\
		& Medium-E & 110.9 & 44.6 $\pm$ 18.6 & 96 $\pm$ 0.5 & 109 $\pm$ 0.2 & 98.7 & 50.7 $\pm$ 26.9 & 73.4 $\pm$ 1.5 & 108.9 $\pm$ 0.7 & 23.7 & 1 $\pm$ 0.2 & 6.1 $\pm$ 1.4 & 7.5 $\pm$ 0.6 \\
		\cmidrule{1-14}    \multirow{5}[4]{*}{\rotatebox{90}{\colorbox{magiccolor}{ Walker2d \quad}}} &       & \multicolumn{4}{c}{BEAR}      & \multicolumn{4}{c}{BRAC-p}    & \multicolumn{4}{c}{AWR} \\
		\cmidrule(r){3-6} \cmidrule(r){7-10} \cmidrule(r){11-14}          & Random & 7.3   & 2.2 $\pm$ 0.1 & 0.6 $\pm$ 0.1 & 2.6 $\pm$ 0.3 & -0.2 & 2.8 $\pm$ 2.8 & 3.3 $\pm$ 2.9 & 8.8 $\pm$ 8 & 1.5 & 0.9 $\pm$ 0.1 & 1.5 $\pm$ 0.6 & 1.5 $\pm$ 0.2 \\
		& Medium & 59.1  & -0.4 $\pm$ 0.1 & 0.6 $\pm$ 0 & 0.1 $\pm$ 0.3 & 77.5 & 28.8 $\pm$ 28.4 & 55.2 $\pm$ 15.8 & 72.9 $\pm$ 9.1 & 17.4 & 12.2 $\pm$ 0.3 & 17.2 $\pm$ 0.2 & 17.2 $\pm$ 0.2 \\
		& Medium-R & 19.2  & 0.4 $\pm$ 0.2 & 4 $\pm$ 0.2 & 10.4 $\pm$ 2.4 & -0.3 & 6.3 $\pm$ 1.2 & 32.1 $\pm$ 11.9 & 34.8 $\pm$ 10.5 & 15.5 & 6 $\pm$ 1 & 1.4 $\pm$ 0 & 2.1 $\pm$ 0.9 \\
		& Medium-E & 40.1  & -0.2 $\pm$ 0.2 & 0.8 $\pm$ 0.4 & 0.6 $\pm$ 0.6 & 76.9 & 21.8 $\pm$ 18.4 & 62.3 $\pm$ 13.1 & 74.3 $\pm$ 1.8 & 53.8 & 40.4 $\pm$ 12.6 & 53 $\pm$ 0.1 & 53.6 $\pm$ 0 \\
		\cmidrule{1-14}    \multirow{5}[3]{*}{\rotatebox{90}{\colorbox{magiccolor}{ Walker2d \quad}}} &       & \multicolumn{4}{c}{BCQ}       & \multicolumn{4}{c}{CQL}       & \multicolumn{4}{c}{MOPO} \\
		\cmidrule(r){3-6} \cmidrule(r){7-10} \cmidrule(r){11-14}         & Random & 4.9   & 3.7 $\pm$ 1.8 & 3.4 $\pm$ 0.4 & 5.2 $\pm$ 0.3 & 7 & 0.5 $\pm$ 1 & 2.7 $\pm$ 0.2 & 6.4 $\pm$ 0.6 & 13.6 & -0.3 $\pm$ 0.1 & -0.2 $\pm$ 0 & -0.2 $\pm$ 0.2 \\
		& Medium & 53.1  & 43 $\pm$ 8.3 & 44.9 $\pm$ 3.3 & 52.7 $\pm$ 0.3 & 79.2 & 43.9 $\pm$ 21.7 & 73.2 $\pm$ 0.8 & 81.2 $\pm$ 1.1 & 17.8 & 5.8 $\pm$ 5.9 & 7.8 $\pm$ 6.2 & 12.2 $\pm$ 5 \\
		& Medium-R & 15    & 5.7 $\pm$ 0.5 & 9.8 $\pm$ 5.2 & 14.6 $\pm$ 0.4 & 26.7 & 1.8 $\pm$ 1.2 & 1.4 $\pm$ 0.4 & 1.8 $\pm$ 0.4 & 39 & 0.8 $\pm$ 0.7 & 9.3 $\pm$ 5.2 & 16.4 $\pm$ 4.9 \\
		& Medium-E & 57.5  & 44.5 $\pm$ 3.6 & 40.6 $\pm$ 16.4 & 57.2 $\pm$ 0.2 & 111 & 46.8 $\pm$ 40 & 109.9 $\pm$ 4.5 & 116.5 $\pm$ 9.1 & 44.6 & 2.9 $\pm$ 3.1 & 15.2 $\pm$ 12.8 & 26.3 $\pm$ 18.4 \\
		\bottomrule
	\end{tabular}%
	\label{tab:error_bar:joint_noise_shift}%
\end{sidewaystable}

\end{document}